# Arc Gradient Descent: A Geometrically Motivated Gradient Descent-based Optimiser with Phase-Aware, User-Controlled Step Dynamics

# (proof-of-concept)

Nikhil Verma (nikhil.verma@vtt.fi), Joonas Linnosmaa(joonas.linnosmaa@vtt.fi),
Espinosa-Leal Leonardo (leonardo.espinosa-leal@vtt.fi), Napat Vajragupta (napat.vajragupta@vtt.fi)

VTT Technical Research Centre of Finland Ltd

## Abstract

This paper introduces Arc Gradient Descent (ArcGD), a geometrically motivated gradient descent-based method. ArcGD distinguishes itself from mainstream optimisers by enforcing explicit bounds on step sizes and dynamically adjusting update magnitudes. It offers fine-grained control over convergence behaviour and parameter evolution throughout different gradient phases, while robustly addressing issues related to exploding and vanishing gradients.

The paper presents the formulation, implementation, and evaluation of the ArcGD optimiser. The evaluation is conducted initially on a non-convex benchmark function and subsequently on a real-world ML dataset. The initial comparative study using the Adam optimiser is conducted on a stochastic variant of the highly non-convex and notoriously challenging Rosenbrock function, renowned for its narrow, curved valley, across dimensions ranging from 2D to 1000D and an extreme case of 50,000D. Two configurations were evaluated to eliminate learning-rate bias: (i) both using ArcGD's effective learning rate and (ii) both using Adam's default learning rate. ArcGD consistently outperformed Adam under the first setting and, although slower under the second, achieved superior final solutions in most cases. In the second evaluation, ArcGD is evaluated against state-of-the-art optimizers (Adam, AdamW, Lion, SGD) on the CIFAR-10 image classification dataset across 8 diverse MLP architectures ranging from 1 to 5 hidden layers. ArcGD achieved the highest average test accuracy (50.7%) at 20,000 iterations, outperforming AdamW (46.6%), Adam (46.8%), SGD (49.6%), and Lion (43.4%), winning or tying on 6 of 8 architectures. Notably, while Adam and AdamW showed strong early convergence at 5,000 iterations, but regressed with extended training, whereas ArcGD continued improving, demonstrating generalization and resistance to overfitting without requiring early stopping tuning. Strong performance on geometric stress tests and standard deep-learning benchmarks indicates broad applicability, highlighting the need for further exploration. Moreover, it is also shown that both a limiting variant of ArcGD and a momentum-augmented ArcGD, recover sign-based momentum updates, revealing a clear conceptual link between ArcGD's phase structure and the core mechanism of the Lion optimiser.

Aimed at both researchers and practitioners, this work offers a concise, implementation-oriented perspective on a controllable alternative to conventional optimisers. While emphasising practical applicability over exhaustive theoretical exposition, the paper provides motivation for the formulation of update rule for the ArcGD mechanism. Formal convergence proofs, extended theoretical analysis, and extensive empirical evaluation are reserved for future work.

---

# 1. Introduction

Modern machine learning techniques now permeate nearly every domain of knowledge. A significant part of their success stems from optimisation algorithms, which are fundamental to training models particularly artificial neural networks. Training typically involves minimising (or maximising) an objective function with respect to its parameters [Bian & Priyadarshi, 2024]. In traditional machine learning, these objective functions are usually differentiable, making gradient descent (GD) an efficient choice for minimisation. However, modern neural network-based algorithms introduce a challenge: the optimisation function is inherently stochastic. Consequently, stochastic gradient descent (SGD) has become the primary method for minimisation. SGD was instrumental in driving the resurgence of machine learning and deep learning during the early 21st century. In recent years, more adaptive and specialised optimisers have emerged. Nevertheless, Adam and its variants namely Adamax [Kingma & Ba, 2015], AdamW [Loshchilov & Hutter, 2019], and Adafactor [Shazeer & Stern, 2018] remain the de facto choices for training state-of-the-art neural networks, especially large language models (LLMs).

More recently, other optimisation methods have been introduced using a more formal differentiation approach via mathematical approximation of the objective function instead of relying on numerical partitioning. For instance, Refined Lion (RLion) uses a continuous monotonic arctan function to construct the update rule [Rong et al., 2024], and Zheng et al. [Zheng et al., 2024] propose optimisers enhanced with sigmoid and hyperbolic tangent functions. Our paper presents Arc Gradient Descent (ArcGD), a simple yet flexible iterative optimisation framework that provides users with explicit control over step size and update dynamics through step regulation and dynamic update scaling. The paper takes a strongly practical approach, presenting a detailed formulation of ArcGD mechanism, and implementation guidance. It includes a stress test and a comparative evaluation against the Adam optimiser on the challenging non-convex Rosenbrock function (with stochasticity) across dimensions from low to ultra-high. Subsequently, ArcGD is benchmarked against leading optimisers (Adam, AdamW, Lion, SGD) on the CIFAR-10 image classification task using eight varied MLP architectures with one to five hidden layers. While the present study prioritises usability and immediate applicability, a detailed convergence analysis, broader theoretical extensions, and extensive empirical evaluation of ArcGD are reserved for future work.

## 2. Motivation for ArcGD

Gradient-based optimization methods frequently encounter two critical challenges: (i) unstable updates in steep regions caused by exploding gradients, and (ii) stagnation in flat regions due to vanishing gradients. Existing adaptive techniques mitigate these issues by automatically adjusting step sizes; however, they typically obscure user control, preventing direct dynamic regulation of updates for different magnitudes of gradients from exploding gradients to vanishing gradients.

The ArcGD algorithm addresses such limitation by explicitly constraining parameter updates within a user-defined ceiling and floor:

- **Ceiling constant**: caps the maximum allowable step size, thereby preventing instability in the presence of large or exploding gradients.
- **Floor constant**: ensures a minimum update magnitude, maintaining progress even in regions where gradients nearly vanish or have completely vanished for a reasonable update. Moreover, the floor constraint can also be made self-adjusting, enabling responsive adaptation that supports convergence while mitigating vanishing gradients.
- **Smooth transition**: between the ceiling and floor, updates evolve continuously, with an optional regulation mechanism that allows the user to accelerate the progress in the mid-range, facilitating a controlled transition between extremes.

Through such explicit bounding of updates augmented by optional regulation, ArcGD provides predictable, stable, and user-controllable convergence across steep, flat, and noisy optimization landscapes. Unlike conventional adaptive methods, ArcGD grants practitioners direct authority over both stability and speed, bridging the gap between rigid automatic schemes and flexible, interpretable optimization.

## 3. Formulation of ArcGD (Core part)

The ArcGD algorithm employs an element-wise update scheme in which all parameters are updated simultaneously, each according to its own partial derivative, analogous to standard gradient descent. The update rule is geometrically motivated by the arc-length formulation in the one-dimensional case from calculus, providing a principled mechanism for step-size control. When generalized to an $n$-dimensional setting, the arc-length formulation is applied per coordinate, treating each dimension independently as a two-dimensional plane spanned by the parameter axis and its gradient direction.

To develop the formulation, one-dimensional objective function $y = f(x)$ that is to be minimized is considered. The differential arc length along the curve is given by

$$ds = \sqrt{dx^2 + dy^2} = \sqrt{1 + (f'(x))^2}\, dx \tag{1}$$

so that a step of fixed arc length $0 < \alpha \ll 1$ corresponds to

$$\Delta x = \pm \frac{\alpha}{\sqrt{1 + (f'(x))^2}} \tag{2}$$

$$\Delta y = \pm \frac{\alpha f'(x)}{\sqrt{1 + (f'(x))^2}} \tag{3}$$

To ensure descent[1], the update must move opposite to the gradient direction. Accordingly, the horizontal update is chosen to satisfy

$$\Delta x = -\frac{\alpha . sign(f'(x))}{\sqrt{1 + (f'(x))^2}} \tag{4}$$

which yields the corresponding vertical displacement

$$\Delta y = f'(x)\, \Delta x = -\frac{\alpha f'(x)}{\sqrt{1 + (f'(x))^2}} \tag{5}$$

Such formulations allow us to analyze the behavior of strict arc-length–based updates in two extreme slope regimes.

Flat regions ($|f'(x)| \to 0$)

- Horizontal step: $\Delta x \approx -\alpha$
- Vertical step: $\Delta y \approx 0$

[1] Such descent (Equation (4)) can be termed as Arc-length Descent (ALD) which is different from ArcGD (Arc Gradient Descent) covered in the paper.

In flat regions, most of the movement occurs along the parameter axis, with negligible reduction in the objective value. Although arc length along the curve is preserved, such updates may overshoot or fail to effectively follow the slope in shallow regions.

Steep regions ($|f'(x)| \rightarrow \infty$)
- Horizontal step: $\Delta x \approx 0$
- Vertical step: $\Delta y \approx -\alpha$

In steep regions, the update becomes almost purely vertical, resulting in negligible progress in parameter space despite substantial movement along the curve. From an optimization perspective, it leads to slow convergence.

Neither extreme is desirable for practical parameter-space optimization. Instead, an update rule is needed that allows meaningful progress in both horizontal and vertical directions, avoiding excessive overshooting in flat regions and stagnation in steep regions.

To address such imbalance, ArcGD assumes a fixed arc-length step $\alpha$ along the curve and evaluates the resulting decrease in the $f(x)$. The measured drop in $f(x)$ (if descent happens) is used for effective progress along the parameter direction and is used to guide the update toward the minimum. In contrast to standard gradient descent, which follows the classical slope interpretation (rise over run resulting in *run proportional to slope*), ArcGD interprets progress through the horizontal run induced by the measured drop in $f(x)$. With that, the decrease in $f(x)$ value determines how far the update moves along the parameter axis (*run with measured drop*), rather than relying directly on raw slope magnitude. Thus, ArcGD decouples parameter-space motion from raw gradient magnitude and instead aligns movement with the observed reduction in the function value. As the optimization approaches the final convergence region, such drop naturally becomes smaller, leading to progressively slower updates of the parameters as they approach the minimum:

$$\Delta x = -\frac{\alpha f'(x)}{\sqrt{1 + (f'(x))^2}} \tag{6}$$

Such modification admits a clear geometric interpretation. The update scales the horizontal motion as if the local slope were 1, while the actual direction of the step still follows the gradient. Only the magnitude of the step is steered toward the 45° case, providing a balanced, (average) intermediate behaviour between the extremes of very flat (0°) and very steep regions (90°). Importantly, the update does not constrain the trajectory to lie along a 45° line. Rather, it adjusts the step magnitude to approximate what would occur if the slope of the curve were moderate (i.e., horizontal and vertical changes are equal), ensuring stable and well-scaled movement in parameter space.

- In shallow regions, the horizontal step is reduced relative to the strict arc-length update, preventing overshooting.
- In steep regions, the horizontal step is increased relative to the strict arc-length update, enabling meaningful progress in parameter space.
- At a slope of 45°($|f'(x)| = 1$), the ArcGD update coincides exactly with the true arc-length step, representing a balanced midpoint between the two extremes.

By explicitly incorporating the descent direction and biasing updates toward a balanced regime that approximates a 45°configuration in $(x, f)$ space, ArcGD preserves the downhill direction while avoiding the extreme horizontal or vertical displacements produced by strict arc-length parametrization in very flat or very steep regions. Consequently, Equation (6) constitutes the core update rule of ArcGD.

By defining $T_x$ (referred to as $T$-transformation for ArcGD in general) as

$$T_x = \frac{g_x}{\sqrt{1 + g_x^2}} \tag{7}$$

it can be noted that $T_x \in (-1, 1)$ for all $g_x \in \mathbb{R}$ where $g_x$ is $f'(x)$. Thus, the update rule simplifies to

$$\Delta x = -\alpha T_x \tag{8}$$

which ensures that

$$\Delta x \in (-\alpha, \alpha) \tag{9}$$

Consequently, every parameter update is inherently constrained within the range $(-\alpha, \alpha)$ which helps reduce the risk of divergence and promotes more stable updates during training. Although such bounding approach shares some similarities with RLion [Rong et al., 2024] and the AlphaGrad update rule [Sane, 2025], ArcGD is formulated based on the arc-length principle, providing a distinct theoretical foundation and leading to a formulation that is fundamentally different in motivation and structure.

## 3.1 Analysis of Core Part

Under the analysis of the core component of the ArcGD, the update formulation is examined for extreme gradient conditions in gradient descent, specifically the cases of exploding and vanishing gradients, as well as its transitional behaviour between these regimes.

Starting from the definition of $T_x$,

$$T_x = \frac{g_x}{\sqrt{1 + g_x^2}} \tag{7}$$

it can be equivalently expressed as

$$T_x = \frac{1}{\sqrt{1 + \frac{1}{g_x^2}}} \tag{10}$$

For extremely large gradients ($g_x \to \pm\infty$),

$$\lim_{g_x \to \pm\infty} T_x = \lim_{g_x \to \pm\infty} \frac{\pm 1}{\sqrt{1 + \frac{1}{g_x^2}}} = \pm 1$$

Hence, the update $\Delta x \to \pm\alpha$, indicating that exploding gradients do not destabilize the update process based on $\alpha$. The update magnitude approaches the learning rate $\alpha$, thereby maintaining stability even under high gradient magnitudes.

For extremely small gradients ($g_x \to 0$),

$$\lim_{g_x \to \pm 0} T_x = \lim_{g_x \to \pm 0} \frac{g_x}{\sqrt{1 + g_x^2}} = \lim_{g_x \to \pm 0} \frac{0}{\sqrt{1 + 0}} = 0$$

Thus, $T_x \to 0$ and consequently $\Delta x \to 0$, implying that the present ArcGD formulation does not mitigate the vanishing gradient problem. In this regime, it reduces to standard gradient descent, leading to stagnation in learning when gradients are very small.

Thus, ArcGD effectively stabilizes the update rule under exploding gradients, but it remains susceptible to the vanishing gradient issue. In the subsequent section, such limitation is addressed through a modified ArcGD formulation. Before introducing the modification, the numerical behavior of $(T_x)$ with respect to $(g_x)$ is examined. The characterization of the $(T_x)$ $vs$ $(g_x)$ relationship delineates distinct operational regions and provides deeper insight into how the core update component behaves. Such understanding is essential to identify potential points where formulative improvements can enhance the update dynamics.

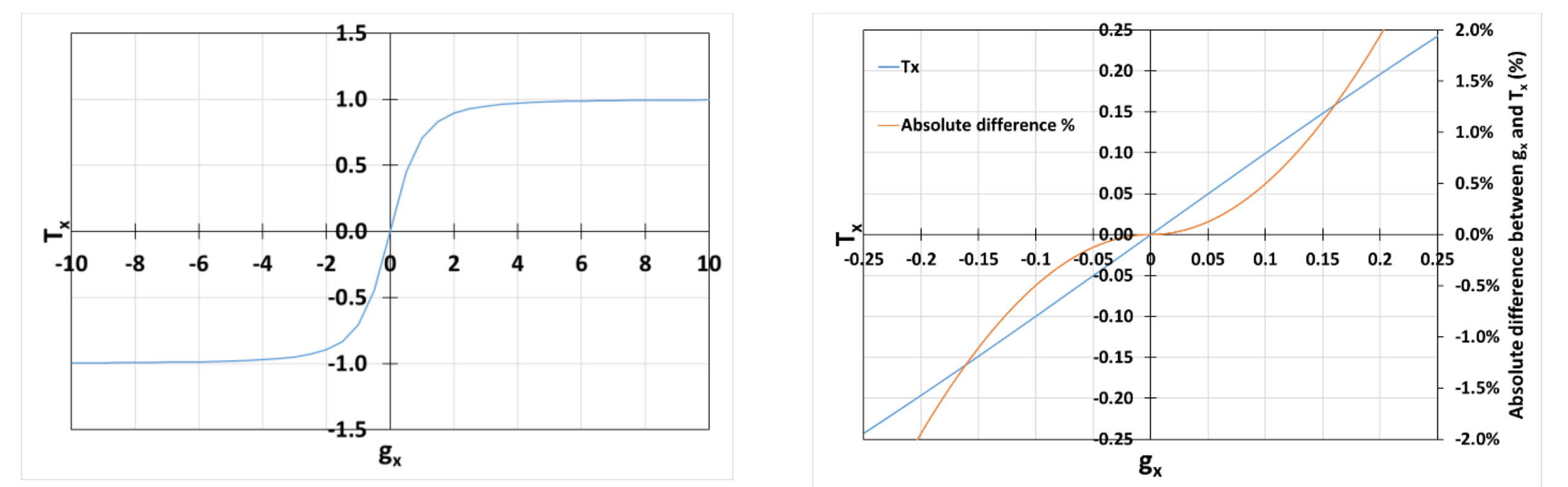

*Figure 1. $T_x$ vs $g_x$*

Figure 1 (left graph) plots $T_x$ versus $g_x$. The curve saturates rapidly: for $|g_x| > 10,\ T_x = sign(g_x).1$ (two-decimal accuracy). Such phase is termed as **exploration phase** for gradients. Saturation effectively and smoothly clips large gradients, mitigating exploding-gradient effects for $|g_x| > 10$.

For $0.01 < |g_x| \leq 10$, the gradients are in a **transition phase.** In the transition phase for $0.2 < |g_x| < 10$, change in $T_x$ is nonlinear in $g_x$. For $0.01 \leq |g_x| \leq 0.2$, change in $T_x$ is approximately linear in $g_x$ and $T_x \approx g_x$ (two-decimal accuracy) (Figure 1, right graph) In such region, ArcGD behaves approximately like standard gradient descent. However, the linear trend continues below 0.01 and gradient is identified to enter **vanishing phase** at $|g_x| = 0.01$. The choice of 0.01 is explained in the next section (section 4).

Thus, ArcGD in its present form provides safe saturated exploration phase for large gradient which can destabilize the update. In addition, the transition phase with linear trend avoids over-damping small, informative gradients, helping fine-grained convergence. The nonlinear transition smoothly bridges these extremes, avoiding hard discontinuities in the update rule. However, as $|g_x| \to 0$ (hence $T_x \to 0$), ArcGD reflects the usual vanishing-gradient behaviour which has been dealt in the next section.

## 4. Modification of ArcGD

The update rule of ArcGD is again reproduced here:

$$\Delta x = -\alpha T_x \tag{8}$$

Let

$$\alpha = eta_{high} + eta_{middle} + eta_{low} \tag{11}$$

where $eta_{low} \ll 1$, $eta_{middle} \ll 1$, $eta_{high} \ll 1$, and overall $\alpha \ll 1$. Substituting Equation (11) in Equation (8) yields:

$$\Delta x = -eta_{high}T_x - eta_{middle}T_x - eta_{low}T_x \tag{12}$$

The term $eta_{high}T_x$ represents the core component of ArcGD. The term $eta_{middle}$ is introduced for the transition region such that it only activates in the transition phase and has diminishing contribution to the updates in other phases. Similarly, the term $eta_{low}$ is introduced for the vanishing phase such that it only significantly contributes to the vanishing phase and has diminishing contribution to the updates in other phases. Clearly in the present form, neither $eta_{low}$ nor $eta_{middle}$ meets the set objective of phase specific significant contribution. To meet the objective, firstly $eta_{low}$ is modified and secondly $eta_{middle}$ is modified.

As mentioned above, in its current form the update rule with $eta_{low}$ does not overcome vanishing gradient problem, since the contribution of $eta_{low}$ is nullified by the small magnitude of $T_x$ in the vanishing phase of gradient. To address such problem, $eta_{low}$ should be formulated in such a way that it

- Activates fully in vanishing phase of gradients,
- Addresses for the diminishing $T_x$ in the product, i.e., in $eta_{low}T_x$.

A natural choice for $eta_{low}$ is as follows:

$$eta_{low} = \frac{c}{|T_x|}(1 - |T_x|) \tag{13}$$

where $c \ll 1$ and $c + eta_{middle} + eta_{high} \ll 1$, which leads to:

$$eta_{low}(T_x) = \frac{c}{|T_x|}(1 - |T_x|)(T_x) \tag{14}$$

$$eta_{low}(T_x) = c.sign(T_x)(1 - |T_x|) \tag{15}$$

which remains finite and non-zero when $T_x \to 0$. Such formulation eliminates the vanishing product problem by "factoring out" $T_x$ from $eta_{low}$, ensuring the new $eta_{low}$ term:

- Activates only when the gradient is very small,
- Smoothly decays to zero as the gradient increases,
- Does not disturb the stable, convergent behaviour in the other regions.

Moreover, when $eta_{low}(T_x)$ gains 99% of the value of $c$ through $(1 - |T_x|)$, i.e., when $1 - |T_x| \geq 0.99$, the vanishing phase for gradients is identified to start. Thus, $|T_x| = 0.01$ or $|g_x| = 0.01$ (two-

decimal accuracy) marks the start of vanishing phase. Furthermore, it is important here to note that $eta_{low}(T_x)$ will never gain the 100% of $c$.

For convenience, let $eta_{high}$ be denoted by $a$. The updated ArcGD rule becomes:

$$\Delta x = -a.T_x - eta_{middle}T_x - c.sign(T_x)(1 - |T_x|) \quad (16)$$

As $g_x \rightarrow 0$, leading to $T_x \rightarrow 0$, we obtain:

$$\Delta x = - c.sign(T_x) \quad (17)$$

Such formulation ensures a guaranteed update of magnitude $\approx c$ even in flat regions, preventing the optimiser from stalling and ensuring continuous progress until a stopping criterion is met (users can also use different learning rate decay rules for $c$). Moreover, in practical implementations, sign $(T_x)$ is defined as zero at $T_x = 0$, ensuring that the floor update is active only for small but non-zero gradients and vanishes exactly at stationary points. Overall, ArcGD maintains a bounded update step within a user-defined range (c, a), and for practical purpose $0 < c \leq |\Delta x| \leq a$. Here, $a$ and $c$ serve as the ceiling and floor constants, respectively, limiting the update magnitude. Moreover, Appendix A elaborates upon a variant of ArcGD (based on Equation (17)) to be a special case of the Lion optimiser [Chen et al., 2023] along with other connection with Lion optimiser. Furthermore, the recently introduced optimizer ThermoLion is driven by the same underlying motivation as ArcGD, but it is inspired by a different conceptual framework and uses a distinct mathematical formulation [Nebli, 2025].

Similarly, to restrict the major contribution of $eta_{middle}$ to the transition phase, $eta_{middle}$ should be formulated in such a way that it activates fully in transition phase of gradients with diminishing effects in other phases. A natural choice for $eta_{middle}$ is as follows:

$$eta_{middle} = b(1 - |T_x|) \quad (18)$$

where $b \ll 1$ and $c + b + a \ll 1$, which leads to:

$$eta_{middle}(T_x) = b(1 - |T_x|)T_x \quad (19)$$

The function $eta_{middle}(T_x)$ peaks at $T = 0.5$, and $eta_{middle}(T_x) \rightarrow 0$ when $|T_x| \rightarrow 1$ or $|T_x| \rightarrow 0$ thus only majorly contributing in the transition phase. Thus $eta_{low}$

- Activates (dominantly) only when the gradient is in transition phase,
- Smoothly decays to zero as the gradient either enter saturation phase or vanishing phase,
- Does not disturb the stable, convergent behaviour in the saturation phase based on $b$,
- For $b \ll 1$, $eta_{middle} \ll 1$ as $(1 - |T_x|)$ is always less than 1.

Thus, the final ArcGD update rule is:

$$\Delta x = -a.T_x - b.T_x(1 - |T_x|) - c.sign(T_x)(1 - |T_x|) \quad (20)$$

where $a$ is the **ceiling constant**, $b$ is the **transition constant**, and $c$ is the **floor constant**. With constant $c$, the update will not settle at a single point but instead oscillates within a small, well-defined region around a minima. In such state, the random fluctuations of the gradients are expected to roughly cancel each other out, preventing the parameters from drifting too far, while the curvature of the loss function is expected to act like a restoring force that keeps them confined. Such behaviour is desirable in deep learning because it biases models toward flatter minima, improves generalization, and stabilizes training even when update magnitudes do not shrink. Optimizers like Lion and signSGD [Bernstein et al., 2018] are likely to exhibit such behavior, particularly near minima where gradients are small and noisy. While they don't always maintain a perfectly stable (stochastic) equilibrium, their design makes them prone to hovering near minima instead of strictly converging to a single point. Notably also, as $T_x$ approaches 1, the update rule converges to the behavior of a sign-based optimizer.

Users may omit the transition-phase terms entirely if desired for reduced computational cost (or choose a very small value for $b$ compared to $a$), as the remaining terms with $a$ and $c$ can still handle updates across all regions. On simplification of Equation (20), it can be noted that magnitude of coefficient for $T$ is $a + b - c$ which will be termed as *effective learning rate*[2] (effective approximate with $b < a$, $b \ll 1$) for ArcGD. An optional gradient-dependent decay of $c$ is derived in Section 4.1 and may be employed in settings where a fixed floor constant is impractical due to the nature of the optimization problem.

Moreover, it is important to note that ArcGD does not explicitly define or enforce gradient phases. Instead, phase-like behavior emerges naturally from the smooth, bounded functional dependence of the update on the $T_x$. References to exploration, transition, and vanishing phases are purely descriptive and are used only to interpret the behavior of the update rule. In particular, values such as $|T_x| \approx 0.5$ or $|T_x| \ll 1$ are not enforced during optimization but arise naturally from the extrema and limiting behavior of the smooth functions $T_x$, $T_x(1 - |T_x|)$, and $(1 - |T_x|)$. Accordingly, these phases are not intended as design choices; rather, they serve as analytical descriptions of the update rule's behavior as a continuous function of the gradient.

## 4.1 Analysis of $eta_{low}$

The formulation of $eta_{low}$ is reproduced here as follows:

$$eta_{low} = \frac{c}{|T_x|}(1 - |T_x|) \tag{13}$$

As $|T_x| \to 0$

$$\lim_{|T_x| \to 0} eta_{low} = \frac{c}{\lim_{|T_x| \to 0} |T_x|} = \infty$$

Theoretically, such divergence or blow up of $eta_{low}$ will destabilise the descent process. Moreover, by definition of $\alpha$, of which $eta_{low}$ is a component, $eta_{low} \ll 1$ leading to the condition:

$$eta_{low} = \frac{c}{|T_x|}(1 - |T_x|) \ll 1 \tag{21}$$

[2] Effective learning rate is defined only to compare the performance of ArcGD with other optimisers.

To satisfy such constraint $eta_{low}$ can be selected as 0.1 (fast late phase convergence but could be risky), 0.01 (default) or 0.001 (conservative). For any chosen value of $eta_{low}$, $b + a \ll 1 - eta_{low}$ must hold (for instance, with default $eta_{low} = 0.01$, the requirement becomes $b + a \ll 0.99$, which is, for example, satisfied by $a = 0.01$ and $b = 0.001$). Hence, $c$ can be expressed as

$$c \leq \frac{eta_{low}|T_x|}{1 - |T_x|} \tag{22}$$

For computational purposes, $c$ can be defined adaptively as $c_{adapt}$:

$$c_{adapt} = minimum\left(c, \frac{eta_{low}|T_x|}{1 - |T_x|}\right) \tag{23}$$

Effectively, in the vanishing gradient phase (refer to equation 23), $1 - |T_x| \rightarrow 1$ leading to $c_{adapt}$ primarily being governed by $eta_{low}|T_x|$. For example, if $c = 0.0001$ and $eta_{low} = 0.01$, then approximately up to $|T_x| = \frac{c}{eta_{low}} = \frac{0.0001}{0.01} = 0.01$, $c$ will contribute to the floor update and after that $c_{adapt}$ will take over. Therefore, the magnitude of $eta_{low}$ effectively regulates and limits the influence of constant $c$ during the vanishing phase of gradients.

With adaptive $c$, ArcGD update rule becomes:

$$\Delta x = -a.T_x - b.T_x(1 - |T_x|) - c_{adapt}.sign(T_x)(1 - |T_x|) \tag{24}$$

It can be noted from Equation (23) that the transition from constant $c$ to $\frac{eta_{low}|T_x|}{1-|T_x|}$ takes place when

$$c = \frac{eta_{low}|T_x|}{1 - |T_x|} \tag{25}$$

which leads to

$$|T_x| = \frac{c}{eta_{low} + c} \tag{26}$$

Thus, based on the value of $c$ and $eta_{low}$, user can decide what minimum threshold $|T_x|$ should reliably be reached before transition from constant $c$ to $\frac{eta_{low}|T_x|}{1-|T_x|}$ takes place. Moreover, after such transition, ArcGD starts to behave like standard gradient descent approximately. So, it is important to conclude here that $c_{adapt}$ provides a mechanism at exact $T_x$ where the transition from Lion optimiser *type* behaviour (constant $c$) to standard gradient descent takes place thus unifying both optimiser through smooth transition. Here (for connection with Lion), $T_x$ is assumed to represent a momentum-based term (e.g., an exponential moving average of gradients) which is explained in section 5.3 and 5.4.

Appendix F provides some heuristic suggestions for the update rule of ArcGD.

# 5. Generalisation to n-Dimensional Setting

In n-dimensional space, let $f: \mathbb{R}^n \to \mathbb{R}$ be the loss function and $x = (x_1, \ldots, x_n)$ the parameters. ArcGD constructs the update by examining each coordinate direction independently via a 2D slice of the loss surface. For each coordinate $x_i$, the slice is the plane defined by $x_i$and $f(x)$, holding all other coordinates fixed. Along such slice, the arc-length differential is

$$ds_i = \sqrt{1 + (\partial f / \partial x_i)^2}\, dx_i \tag{27}$$

where the partial derivative $g_i = \partial f / \partial x_i$ enters directly. Consequently, $T_x$ from Equation (7) is applied for each coordinate direction as:

$$T_i = \frac{g_i}{\sqrt{1 + g_i^2}} \tag{28}$$

Repeating such process for all coordinates produces a vector-valued update (fully vectorised) where each component reflects the local geometry along its 2D slice. Although each slice is considered independently, all components are applied simultaneously in the same iteration, so the method is not sequential coordinate descent. Such construction ensures that in n-D, each parameter direction preserves the geometric property defined by its 2D slice, providing a well-defined foundation for elementwise updates in ArcGD.

## 5.1 ArcGD with non-adaptive $c$

ArcGD will update parameters $\boldsymbol{x_t} \in \mathbb{R}^n$ using gradients at iteration $t$, i.e., $\boldsymbol{g_t} = \nabla f(x_t)$, as follows:

$$T_i = \frac{g_{t,i}}{\sqrt{1 + g_{t,i}^2}} for\ i = 1, 2, 3 \ldots \ldots .., n$$

$$\Delta x_i = -(a.T_i + b.T_i(1 - |T_i|) + c.sign(T_i)(1 - |T_i|))$$

$$\boldsymbol{x_{t+1}} = \boldsymbol{x_t} + \Delta \boldsymbol{x}$$

where,

- $g_{t,i}$ is the gradient of the $i$-th parameter at iteration t.
- $\Delta x = [\Delta x_1, \Delta x_2, \ldots \ldots \ldots \ldots \ldots ., \Delta x_n]$
- $a$, $b$, and $c$ are scalar hyperparameters controlling step shape and scaling.
- Default values for $a$, $b$, and $c$ are 0.01, 0.001, and 0.0001 respectively, where $c + b + a \ll 1$.

## 5.2 ArcGD with adaptive $c$

ArcGD will update parameters $\boldsymbol{x_t} \in \mathbb{R}^n$ using gradients at iteration $t$, i.e., $\boldsymbol{g_t} = \nabla f(x_t)$, as follows:

$$T_i = \frac{g_{t,i}}{\sqrt{1 + g_{t,i}^2}} for\ i = 1, 2, 3 \ldots \ldots .., n$$

$$c_{adapt_i} = minimum\left(c, \quad \frac{eta_{low}|T_i|}{1-|T_i|}\right)$$

$$\Delta x_i = -(a.T_i + b.T_i(1-|T_i|) + c_{adapt_i}.sign(T_i)(1-|T_i|))$$

$$\boldsymbol{x_{t+1}} = \boldsymbol{x_t} + \Delta \boldsymbol{x}$$

where,

- $g_{t,i}$ is the gradient of the $i$-th parameter at iteration t.
- $\Delta x = [\Delta x_1, \Delta x_2, \dots \dots \dots \dots \dots, \Delta x_n]$
- $a$, $b$, $c$ and $eta_{low}$ are scalar hyperparameters controlling step shape and scaling.
- Default values for $a$, $b$, and $c$ are 0.01, 0.001, and 0.0001 respectively, where $c + b + a \ll 1$.
- Default value for $eta_{low}$ is 0.01.

## 5.3 ArcGD with non-adaptive $c$ for noisy landscape

ArcGD will update parameters $\boldsymbol{x_t} \in \mathbb{R}^n$ using moving average of gradients as low pass filter at iteration $t$, i.e., $\boldsymbol{g_t} = \nabla f(x_t)$, as follows:

$$\boldsymbol{m_t} = \beta \boldsymbol{m_{t-1}} + (1-\beta)\boldsymbol{g_t}$$

$$T_i = \frac{m_{t,i}}{\sqrt{1+m_{t,i}^2}} for\ i = 1,2,3 \dots \dots .., n$$

$$\Delta x_i = -(a.T_i + b.T_i(1-|T_i|) + c.sign(T_i)(1-|T_i|))$$

$$\boldsymbol{x_{t+1}} = \boldsymbol{x_t} + \Delta \boldsymbol{x}$$

where,

- $g_{t,i}$ is the gradient of the $i$-th parameter at iteration t.
- $m_t$ is the moving average of gradients where $\beta = 0.9$ and $\boldsymbol{m_0} = \boldsymbol{g_t}$ (first evaluated gradient).
- $\Delta \boldsymbol{x} = [\Delta x_1, \Delta x_2, \dots \dots \dots \dots \dots, \Delta x_n]$.
- $a$, $b$, and $c$ are scalar hyperparameters controlling step shape and scaling.
- Default values for $a$, $b$, and $c$ are 0.01, 0.001, and 0.0001 respectively, where $c + b + a \ll 1$.

## 5.4 ArcGD with adaptive $c$ for noisy landscape

ArcGD will update parameters $\boldsymbol{x_t} \in \mathbb{R}^n$ using moving average of gradients as low pass filter at iteration $t$, i.e., $\boldsymbol{g_t} = \nabla f(x_t)$, as follows:

$$\boldsymbol{m_t} = \beta \boldsymbol{m_{t-1}} + (1-\beta)\boldsymbol{g_t}$$

$$T_i = \frac{m_{t,i}}{\sqrt{1+m_{t,i}^2}} for\ i = 1,2,3 \dots \dots .., n$$

$$c_{adapt_i} = minimum\left(c, \quad \frac{eta_{low}|T_i|}{1 - |T_i|}\right)$$

$$\Delta x_i = -(a.T_i + b.T_i(1 - |T_i|) + c_{adapt_i}.sign(T_i)(1 - |T_i|))$$

$$\boldsymbol{x_{t+1}} = \boldsymbol{x_t} + \Delta\boldsymbol{x}$$

where,

- $g_{t,i}$ is the gradient of the $i$-th parameter at iteration t.
- $m_t$ is the moving average of gradients where $\beta = 0.9$ and $\boldsymbol{m_0} = \boldsymbol{g_t}$ (first evaluated gradient).
- $\Delta\boldsymbol{x} = [\Delta x_1, \Delta x_2, \dots \dots \dots \dots \dots, \Delta x_n]$.
- $a$, $b$, $c$, and $eta_{low}$ are scalar hyperparameters controlling step shape and scaling.
- Default values for $a$, $b$, and $c$ are 0.01, 0.001, and 0.0001 respectively, where $c + b + a \ll 1$.
- Default value for $eta_{low}$ is 0.01.

# 6. Tests

The optimiser's performance is assessed using the stochastic Rosenbrock function over dimensions ranging from low to ultra-high, as well as on the CIFAR-10 image classification dataset. The stochastic Rosenbrock function, especially high-dimensional, provides a rigorous geometric benchmark that exposes an optimizer's intrinsic stability. Its narrow, curved valley and strong cross-coordinate coupling make progress extremely sensitive to step-size control, directional consistency, and curvature handling. Demonstrating reliable convergence in tens of thousands of dimensions can show that the method can sustain coherent descent in stiff, ill-conditioned landscapes without relying on architectural structure, implicit regularization, or second-moment smoothing.

CIFAR-10 complements the test on the stochastic Rosenbrock function by evaluating the optimizer under realistic deep-learning conditions. Training different levels of deep networks introduces substantial gradient noise, rapidly shifting curvature, and heterogeneous layer-wise scales. Robust performance across such spectrum can indicate that the optimizer adapts effectively to complex, nonstationary loss surfaces and maintains competitive convergence in practical workloads.

Taken together, such benchmarks, one defined by pure geometric difficulty and the other by real-world stochastic training, will provide a comprehensive assessment of optimizer behaviour. Consistently strong results on both would suggest that the method's advantages arise from the structure of the update rule itself rather than from task-specific coincidences or architecture-dependent effects.

## 6.1 Evaluation 1: Rosenbrock Function with stochasticity

Tests were conducted to evaluate the performance of ArcGD and Adam on the Rosenbrock function widely used benchmark in optimisation research. The function for a vector $\boldsymbol{x} = (x_1, x_2, \dots\dots, x_n)$ is

$$f(\boldsymbol{x}) = \sum_{i=1}^{n-1} [100(x_{i+1} - x_i^2) + (1 - x_i)^2]$$

with a global minimum at $x^* = (1, 1, \dots.,1)$. Its non-convex topology, characterized by a narrow, curved valley with steep curvature in the transverse direction and a flat slope along the longitudinal direction, presents a challenging test for gradient-based optimisation methods. The conditioning worsens rapidly with $n$ making the descent very hard towards minima. High dimensional tests will expose scalability, stability, and structure-handling weaknesses that small dimensional tests can easily hide. The gradient, employed for all the algorithms, is as follows:

for $1 < i < n$:

$$\frac{\partial f}{\partial x_i} = 200(x_i - x_{i-1}^2) - 400x_i(x_{i+1} - x_i^2) - 2(1 - x_i)$$

Boundary cases:

$$\frac{\partial f}{\partial x_1} = -400x_1(x_2 - x_1^2) - 2(1 - x_1)$$

$$\frac{\partial f}{\partial x_n} = 200(x_n - x_{n-1}^2)$$

providing precise first-order information for update steps.

The experimental configurations are summarized in Table 1 and Table 2, which together define two complementary test matrices designed to evaluate the performance and scalability of ArcGD relative to Adam. The test problems span dimensions from 2D to 1000D and an extreme case of 50,000D, enabling assessment across both low and ultra–high dimensional optimization landscapes. Table 1 (Configuration A) shows a configuration in which the Adam optimiser uses the same learning rate as the default effective learning rate of ArcGD $(0.0109)$. Table 2 (Configuration B) presents a similar configuration, where ArcGD's effective learning rate is $0.00099$, closely matching Adam's default learning rate of 0.001. These settings aim to compare the performance of the two optimisers without the influence of differing learning rates. ArcGD with adaptive c for noisy landscape is used for both configurations.

*Table 1. Configuration A*

| Test Set | Dimensions | Number of Runs |
| --- | --- | --- |
| A2 | 2 | 10 |
| A10 | 10 | 10 |
| A100 | 100 | 10 |
| A1000 | 1000 | 10 |
| A50000 | 50000 | 3 |
| Setting of Adam: learning rate = 0.0109, $\beta_1 = 0.9$. $\beta_2 = 0.999$, $\epsilon = 10^{-8}$ | | |

*Table 2. Configuration B*

| Test Set | Dimensions | Number of Runs |
| --- | --- | --- |
| B2 | 2 | 10 |
| B10 | 10 | 10 |
| B100 | 100 | 10 |
| B1000 | 1000 | 10 |
| B50000* | 50000 | 3 |
| Setting of Adam (all default): learning rate = 0.001, $\beta_1 = 0.9$. $\beta_2 = 0.999$, $\epsilon = 10^{-8}$<br>Setting of ArcGD: $a = 0.0009$, $b = 0.0001$, $c = 0.00001$, $\beta = 0.9$ | | |
| *These three tests were run on different computing resources from other test sets as ArcGD required more resources to complete because of smaller effective learning rate. | | |

The two-dimensional cases serve as computationally inexpensive, visually interpretable baselines that allow direct observation of parameter trajectories and oscillatory behaviours caused by momentum dynamics. Ten independent runs were conducted for each configuration (except the 50,000D case, which was limited to three runs due to computational cost). Such replication level balances statistical reliability with computational efficiency, providing representative estimates of performance consistency across dimensions.

The performance of both optimizers was evaluated using a stochastic variant of the n-dimensional Rosenbrock function, a highly non-convex landscape characterized by a narrow, curved valley and a challenging path to the global minimum at $x^* = [1, 1, \ldots\ldots, 1]$. Stochasticity was introduced via additive Gaussian noise ($\sigma = 1 \times 10^{-3}$) in the objective function, and $\sigma = 1 \times 10^{-4}$ in the gradient calculations, to emulate noisy real-world optimization conditions.

Performance was assessed using multiple criteria:

- Convergence speed, measured by the number of iterations required to reach a predefined loss tolerance.
- Stability, assessed by the consistency of convergence paths across multiple runs.

The loss was smoothed using an exponentially weighted moving average (90% previous value, 10% current loss) to suppress stochastic fluctuations. Convergence was declared when the smoothed loss showed no improvement greater than $1{\times}10^{-5}$ for more than 1000 consecutive iterations, indicating that progress had plateaued. A second condition was included to declare runs as False for convergence**:** if the smoothed loss remained above a fixed threshold of 0.1, the optimizer was treated as having failed to reach the valley of the Rosenbrock function. It prevents runs trapped in high-loss regions from being falsely counted as successes. The gradient norm was intentionally excluded from the stopping logic. In high-dimensional or noisy settings it becomes unreliable as its magnitude scales with √d, and noise prevents it from decaying even near good solutions. Including a gradient-based criterion would unfairly penalize optimizers like Adam, which may maintain relatively larger gradients due to adaptive scaling even when the loss has already stabilized. Using only the smoothed loss, together with the high-loss threshold, provides a consistent and fair measure of convergence across all dimensionalities, from 2-D up to 50,000-D, and ensures uniform comparison between the two optimisers.

All experiments used identical random initializations for reproducibility: each initial vector ($\boldsymbol{x_0}$) was sampled uniformly from the domain $[-3, 3]^n$ using a fixed random seed (numpy.random.seed(42)). Such setup ensured consistent evaluation conditions while capturing the inherent variability of stochastic optimization across different dimensional spaces.

## 6.2 Evaluation 2: CIFAR-10

ArcGD's empirical performance with adaptive $c$ (for noisy landscape) was further evaluated by comparing it to established optimizers (Adam, AdamW, Lion, SGD) on the CIFAR-10 image classification benchmark [Krizhevsky, 2009]. Eight MLP architectures of varying depth and width were tested (Table 3), ranging from shallow networks (1 hidden layer, ~98K parameters) to very deep networks (5 hidden layers, ~4.4M parameters).

*Table 3. MLP architectures tested. Total Parameters is an estimate on CIFAR-10 dataset (3072 input features, 10 output classes).*

| Name | Hidden Layers | Total Parameters |
|---|---|---|
| Tiny | [32] | ~98K |
| Shallow | [64] | ~197K |
| Medium | [512, 256] | ~2.0M |
| Deep | [1024, 512, 256, 128] | ~5.5M |
| Very deep | [512, 512, 512, 256, 256] | ~4.4M |
| Const_shallow | [256] | ~787K |
| Const_medium | [256, 256] | ~853K |
| Const_deep | [256, 256, 256] | ~919K |

Network configurations included ReLU activation functions for hidden layers, softmax for classification, cross-entropy loss, and He Normal weight initialization [He et al., 2015]. Commonly reported hyperparameters were used for Adam, AdamW, and Lion without architecture-specific tuning. SGD employed a learning rate of 0.005 (5× higher than default) to improve baseline performance. ArcGD parameters remained fixed as introduced above (default setting for adaptive c). Complete hyperparameter settings are provided in Table 4.

*Table 4. Optimiser hyperparameters.*

| Optimiser | Hyperparameters |
|---|---|
| ArcGD | a = 0.01, b = 0.001, c = 0.0001, beta = 0.9, eta_low = 0.01 |
| Adam | lr = 0.001, beta1 = 0.9, beta2 = 0.999, eps = 1e-8 |
| AdamW | lr = 0.001, weight_decay = 0.01, beta1 = 0.9, beta2 = 0.999, eps = 1e-8 |
| Lion | lr = 0.001, beta1 = 0.9, beta2 = 0.99, weight_decay = 0.01 |
| SGD | lr = 0.005 |

The MLPs were trained on a single NVIDIA Tesla V100 GPU for a maximum of 20000 iterations with a batch size of 128. Early stopping was implemented with a patience of 500 iterations and a minimum accuracy delta of 0.0001. A train-test split of 80:20 was used on the full CIFAR-10 dataset (50,000 training samples). Results are based on a single random seed without cross-validation, representing preliminary findings that warrant further statistical validation.

# 7. Results and Discussion

## 7.1 Evaluation 1: Rosenbrock Function with stochasticity

The summary of results from various Rosenbrock runs are shown in Table 5. It should be noted that negative final losses (in results) occur naturally because the stochastic Rosenbrock includes additive noise, which can push near-zero values slightly below zero. Moreover, the reported metrics in Table 5 which are computation time, final distance to the minima, final loss, and final gradient norm are evaluated as average values computed across all successful runs.

*Table 5. Summary of tests (Top: Configuration A; Bottom: Configuration B)*

| Run Summary | Optimizer | Total_Runs | Converged_Runs | Convergence_Rate | Avg_Iterations | Avg_Time | Avg_Distance | Avg_Final_Loss | Avg_Final_GradNorm |
|---|---|---|---|---|---|---|---|---|---|
| A2 (2D) | ADAM | 10 | 10 | 100 | 9440 | 0.47 | 2.27E-04 | -1.36E-05 | 3.58E-04 |
| | ArcGD | 10 | 10 | 100 | 2802 | 0.15 | 9.31E-06 | -7.37E-05 | 5.00E-04 |
| A10 (10D) | ADAM | 10 | 6 | 60 | 11840 | 0.85 | 2.47E-05 | 3.71E-04 | 2.45E-02 |
| | ArcGD | 10 | 8 | 80 | 2897 | 0.25 | 4.54E-06 | 3.74E-04 | 2.27E-03 |
| A100 (100D) | ADAM | 10 | 9 | 90 | 13432 | 1.29 | 1.99E-04 | 4.41E-03 | 3.20E+00 |
| | ArcGD | 10 | 9 | 90 | 4378 | 0.4 | 1.33E-06 | -2.86E-04 | 1.55E-02 |
| A1000 (1000D) | ADAM | 10 | 9 | 90 | 15658 | 1.99 | 2.71E-04 | 6.57E-02 | 1.50E+01 |
| | ArcGD | 10 | 10 | 100 | 9197 | 1.22 | 1.11E-06 | 6.91E-04 | 5.23E-02 |
| A50000 (50000D) | ADAM | 3 | 0 | 0 | N/A | N/A | N/A | N/A | N/A |
| | ArcGD | 3 | 3 | 100 | 22993 | 104.43 | 1.07072E-06 | -1.00437E-05 | 3.85E-01 |
| **Run Summary** | **Optimizer** | **Total_Runs** | **Converged_Runs** | **Convergence_Rate** | **Avg_Iterations** | **Avg_Time** | **Avg_Distance** | **Avg_Final_Loss** | **Avg_Final_GradNorm** |
| B2 (2D) | ADAM | 10 | 10 | 100 | 17443 | 0.81 | 2.08E-04 | -5.45E-04 | 3.57E-03 |
| | ArcGD | 10 | 10 | 100 | 10765 | 0.62 | 1.39E-03 | 5.60E-04 | 9.06E-04 |
| B10 (10D) | ADAM | 10 | 6 | 60 | 20126 | 0.98 | 3.90E-05 | 8.10E-04 | 3.04E-02 |
| | ArcGD | 10 | 7 | 70 | 12472 | 0.67 | 3.55E-04 | 2.58E-04 | 1.32E-03 |
| B100 (100D) | ADAM | 10 | 9 | 90 | 22994 | 1.11 | 1.40E-05 | 1.87E-05 | 2.28E-01 |
| | ArcGD | 10 | 9 | 90 | 27966 | 1.6 | 1.05E-05 | 3.60E-04 | 3.30E-03 |
| B1000 (1000D) | ADAM | 10 | 9 | 90 | 28290 | 2.22 | 2.33E-05 | 2.02E-04 | 1.30E+00 |
| | ArcGD | 10 | 10 | 100 | 75166 | 7.27 | 6.84E-06 | 4.80E-04 | 1.01E-02 |
| B50000 (50000D) | ADAM | 3 | 3 | 100 | 36731 | 92.53 | 2.66E-05 | 3.22E-02 | 1.07E+01 |
| | ArcGD | 3 | 3 | 100 | 211456 | 602.68 | 3.43E-07 | 1.67E-04 | 7.18E-02 |

*Avg means Average; Avg_Distance is final distance to minima.*

### 7.1.1 Configuration A

In the 2D problem (A2), both Adam and ArcGD achieved perfect convergence in all 10 runs. Adam required significantly more iterations (9,440) and slightly more time (0.47 s) than ArcGD (2,802 iterations, 0.15s). While ArcGD reached a slightly lower final loss (-7.37E-05 versus -1.36E-05), Adam achieved a marginally smaller gradient norm (3.58E-04 versus 5.00E-04). It indicates that both optimizers are highly effective in low dimensions, with ArcGD being faster and Adam slightly more precise in terms of gradient norm.

For the 10D problem (A10), ArcGD outperformed Adam in convergence rate and efficiency. Adam converged in only 6 out of 10 runs, whereas ArcGD succeeded in 8 out of 10. Adam required over 11,000 iterations and 0.85 seconds per run, while ArcGD converged with fewer iterations (2,897) and much less time (0.25 s). ArcGD also achieved smaller final distances to the minimum (4.54E-06 versus 2.47E-05) and lower gradient norms (2.27E-03 versus 2.45E-02), demonstrating more precise convergence in moderate dimensions.

In the 100D problem (A100), both optimizers had similar convergence rates (9/10 runs), but ArcGD was far more efficient, requiring only (4,378) iterations and (0.4) s versus Adam's 13,432 iterations and 1.29 s. ArcGD also reached much smaller final distances (1.33E-06 versus 1.99E-04) and gradient norms (1.55E-02 versus 3.20E+00), showing superior precision even as dimensionality increased.

For the 1,000D problem (A1000), Adam converged in 9 out of 10 runs, while ArcGD achieved perfect convergence. Adam required 15,658 iterations and 1.99 s per run, compared to ArcGD's 9,197 iterations and 1.22 s. ArcGD achieved far smaller distances to the minimum (1.11E-06 versus 2.71E-04) and gradient norms (5.23E-02 versus 1.50E+01), highlighting its robustness and accuracy in high dimensions.

In the extreme 50,000D problem (A50000), Adam failed to converge in any of the 3 runs (possibly due to high learning rate compared to its default), whereas ArcGD converged perfectly in all 3 runs, although at a higher computational cost (22,993 iterations and 104.43 seconds). ArcGD still achieved a very small distance to the minimum (1.07072E-06) and a moderate gradient norm (3.85E-01), demonstrating its ability to handle extremely high-dimensional optimization.

Overall, under configuration A tests, ArcGD consistently outperformed Adam in terms of efficiency, convergence reliability, and precision as dimensionality increased. ArcGD also remained robust even in 50,000 dimensions.

#### 7.1.2 Configuration B

In the 2D problem (A2), both Adam and ArcGD achieved perfect convergence in all 10 runs. Adam required more iterations (17,443) and slightly more time (0.81 s) compared to ArcGD (10,765 iterations, 0.62 s). In terms of solution quality, Adam reached a slightly higher gradient norm (3.57E-03) but also a more negative final loss (-5.45E-04), whereas ArcGD had a lower gradient norm (9.06E-04) and a positive final loss (5.60E-04). It suggests that while both methods converged reliably, Adam produced a marginally more precise solution in terms of loss.

For the 10D problem (A10), convergence rates decreased for both optimizers, with Adam succeeding in 6 of 10 runs and ArcGD in 7 of 10. Adam required more iterations (20,126) than ArcGD (12,472), and ArcGD was faster (0.67 s versus 0.98 s). ArcGD achieved a smaller gradient norm (1.32E-03) compared to Adam (3.04E-02) and a lower final loss (2.58E-04 versus 8.10E-04), indicating more accurate convergence in the 10-dimensional setting.

In the 100D problem (A100), both optimizers converged in 9 of 10 runs. Adam was slightly faster per run (1.11 s versus 1.6 s for ArcGD) and required fewer iterations, but ArcGD achieved a lower gradient norm (3.30E-03 versus 2.28E-01) while maintaining a small distance to the minimum. It shows that ArcGD provided more precise convergence despite requiring more iterations.

For the 1,000D problem (A1000), Adam converged in 9 of 10 runs, whereas ArcGD achieved perfect convergence in all 10 runs. Adam needed 28,290 iterations and 2.22 seconds per run, while ArcGD required 75,166 iterations and 7.27 seconds, significantly more computational effort. However, ArcGD achieved a far smaller final distance to the minimum (6.84E-06 versus 2.33E-05) and a much lower gradient norm (1.01E-02 versus 1.30E+00), demonstrating its superior precision in high dimensions.

In the extreme 50,000D problem (A50000), both optimizers converged in all 3 runs. Adam required 36,731 iterations and 92.53 seconds, while ArcGD required a massive 211,456 iterations and 602.68 seconds. Despite the heavy computational cost, ArcGD reached an extremely small distance to the minimum (3.43E-07) and a low gradient norm (7.18E-02), whereas Adam's gradient norm remained high (1.07E+01). Such results highlight ArcGD's ability to achieve highly precise solutions in very high-dimensional settings even with smaller effective learning rate, though at the expense of time.

Overall, under Configuration B, the results, in general, show that ArcGD is consistently reliable across all problem dimensions, achieving smaller distances to the minimum and lower gradient norms than Adam, particularly in higher-dimensional settings. While Adam generally converges more quickly in terms of iterations and runtime, ArcGD provides more precise solutions, and its advantage in solution quality becomes increasingly pronounced as dimensionality grows for Rosenbrock function with stochasticity.

Detail results for each of run of various test cases are provided Appendix B and Appendix C.

## 7.2 Evaluation 2: CIFAR-10

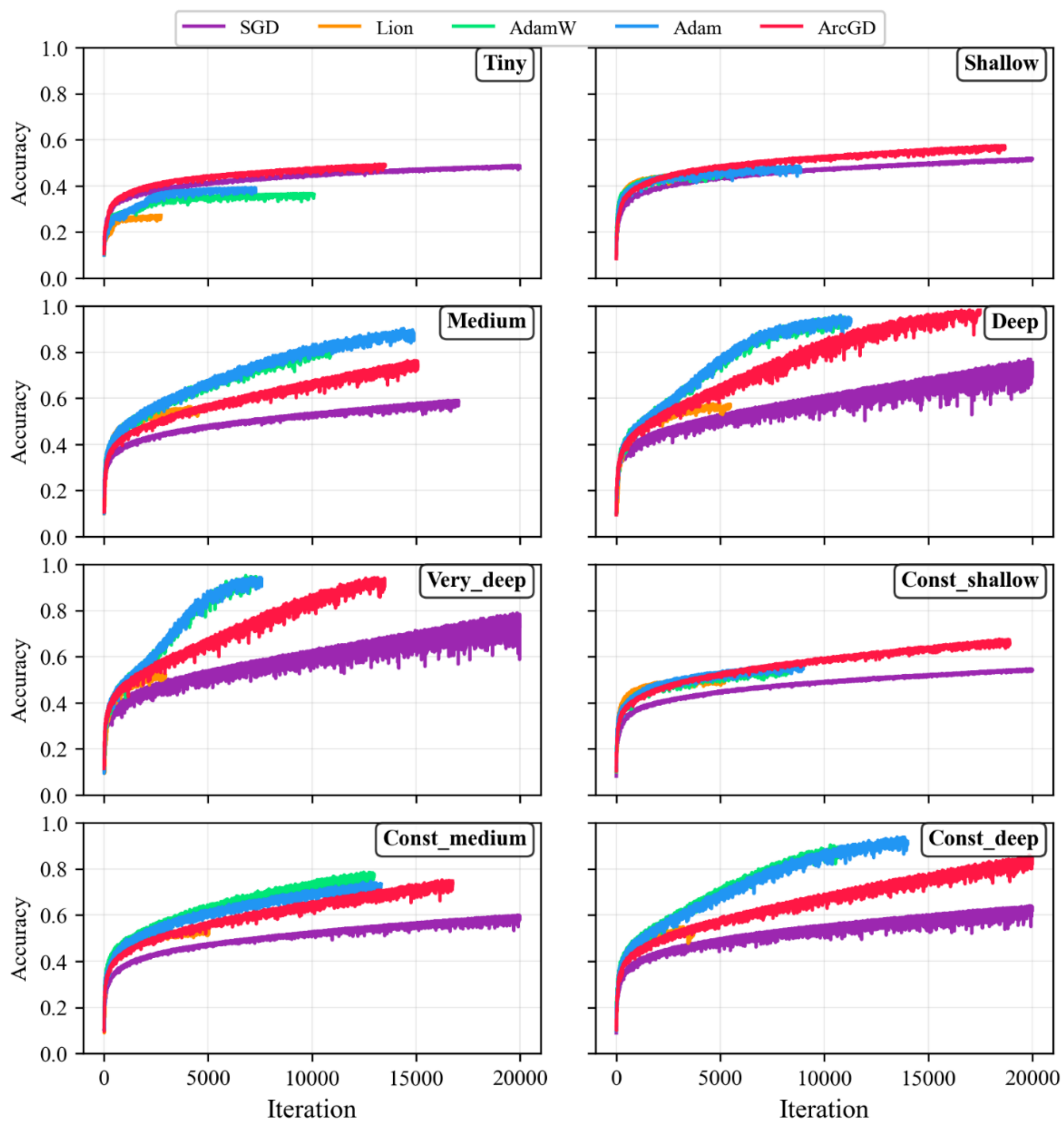

*Figure 2. Training accuracy on full CIFAR-10 dataset. ArcGD (blue), Adam (yellow), AdamW (green), Lion (red), SGD (purple).*

On training accuracy at 20,000 iterations (Figure 2), ArcGD demonstrated competitive performance on tiny and shallow depth MLPs, achieving accuracy comparable to Adam and AdamW. However, on deeper architectures (3+ hidden layers), ArcGD shows slower training convergence compared to AdamW and Lion.

As noted in Table 6, despite slower training convergence, ArcGD achieved the highest test accuracy at both 5,000 iterations (48.4% average) and 20,000 iterations (50.7% average), outperforming Adam (47.7% → 46.6%), AdamW (47.6% → 46.8%), Lion (42.7% → 43.3%), and SGD (44.1% → 49.6%). Notably, while Adam showed strong early performance at 5k iterations, it regressed with additional training (-1.1%), whereas ArcGD continued improving (+2.3%), demonstrating generalization and resistance to overfitting. An ablation study in Appendix D shows that increasing $eta_{low}$ accelerates training but degrades final accuracy.

At 20k iterations, ArcGD won on 6 of 8 architectures, with particularly strong performance on medium-width networks (medium: 53.8%, const_medium: 52.4%) and very deep networks (very_deep: 51.8%, deep: 53.1%). The train-test divergence, where slower training convergence yields superior final generalization, suggests ArcGD's accumulated gradient term provides implicit regularization that becomes increasingly beneficial with extended training.

*Table 6. Test accuracies on CIFAR-10 dataset at 5,000 and 20,000 training iterations.*

| Optimiser | ArcGD | Adam | AdamW | Lion | SGD |
|---|---|---|---|---|---|
| Tiny | 5K: **41.1 %**<br>20K: **44.8 %** | 5K: 37.2 %<br>20K: 37.4 % | 5K: 34.9 %<br>20K: 35.3 % | 5K: 25.9 %<br>20K: 25.9 % | 5K: 40.0 %<br>20K: 44.6 % |
| Shallow | 5K: **45.6 %**<br>20K: **48.6 %** | 5K: 43.7 %<br>20K: 43.0 % | 5K: 43.3 %<br>20K: 41.9 % | 5K: 41.2 %<br>20K: 41.1 % | 5K: 41.9 %<br>20K: 47.5 % |
| Medium | 5K: 50.6 %<br>20K: **53.8 %** | 5K: 51.2 %<br>20K: 49.8 % | 5K: **52.4 %**<br>20K: 50.7 % | 5K: 46.8 %<br>20K: 46.8 % | 5K: 45.0 %<br>20K: 51.4 % |
| Deep | 5K: **51.8 %**<br>20K: 53.1 % | 5K: 50.8 %<br>20K: 49.3 % | 5K: 51.7 %<br>20K: 49.7 % | 5K: 47.1 %<br>20K: 48.6 % | 5K: 47.2 %<br>20K: **53.2 %** |
| Very_deep | 5K: **50.6 %**<br>20K: **51.8 %** | 5K: 49.1 %<br>20K: 48.5 % | 5K: 49.1 %<br>20K: 49.3 % | 5K: 46.6 %<br>20K: 46.6 % | 5K: 47.0 %<br>20K: 49.9 % |
| Const_shallow | 5K: **48.2 %**<br>20K: **51.3 %** | 5K: 47.5 %<br>20K: 48.6 % | 5K: 47.5 %<br>20K: 47.2 % | 5K: 45.5 %<br>20K: 43.6 % | 5K: 42.3 %<br>20K: 49.3 % |
| Const_medium | 5K: 48.7 %<br>20K: **52.4 %** | 5K: 50.5 %<br>20K: 50.4 % | 5K: **51.8 %**<br>20K: 49.6 % | 5K: 46.2 %<br>20K: 47.0 % | 5K: 43.7 %<br>20K: 49.7 % |
| Const_deep | 5K: **51.0 %**<br>20K: 50.5 % | 5K: 50.8 %<br>20K: 47.7 % | 5K: 50.5 %<br>20K: 48.8 % | 5K: 42.3 %<br>20K: 47.1 % | 5K: 46.6 %<br>20K: **51.1 %** |
| **Average** | 5K: **48.4 %**<br>20K **50.7 %** | 5K: 47.7 %<br>20K: 46.6 % | 5K: 47.6 %<br>20K: 46.8 % | 5K: 42.7 %<br>20K: 43.3 % | 5K: 44.1 %<br>20K: 49.6 % |

# 8. Conclusion

Arc Gradient Descent (ArcGD) offers a mathematically grounded yet practical optimisation framework that combines stability, interpretability, and user control. By introducing explicit ceiling and floor constraints, ArcGD mitigates exploding and vanishing gradients, ensuring consistent progress across steep, flat, and noisy landscapes. Its dynamic update scaling and smooth transition mechanism enable fine-grained regulation of step sizes, predictable convergence, and a balance between stability and speed.

ArcGD stood out as a stable and effective optimisation method in both controlled experiments and a real-world scenario. It exceled in handling curvature, maintaining stable updates, and achieving precise convergence on high-dimensional stochastic Rosenbrock problems, even under severe anisotropy. Additionally, its robust step-size control across different gradient phases, combined with its ability to keep gradient norms small and deliver superior asymptotic accuracy under extreme dimensionality and curvature, makes it a strong candidate for high-dimensional optimisation. Such advantages appear to extend to deep-learning contexts, where ArcGD consistently demonstrated strong generalisation on CIFAR-10 and outperformed widely used optimisers in most runs, despite slower initial training on deeper networks.

The method's success seems rooted in the structure of its update rule rather than task-specific factors or model biases, underscoring its methodological soundness. Impressive results on both geometric stress tests and standard deep-learning benchmarks suggest broad applicability, warranting further investigation. Future research should confirm the findings on larger and more diverse benchmark suites, establish formal convergence guarantees, and evaluate robustness across an even wider range of real-world tasks.

# Acknowledgement

We acknowledge the financial support provided by VTT Technical Research Centre of Finland (grant 141083-1.14), enabling this study. The authors also wish to thank CSC – IT Center for Science, Finland, for the computational resources provided.

# Appendix A (ArcGD and Lion Optimiser)

**Lion Optimiser** [Chen et al., 2023]

Let $\theta_t \in \mathbb{R}^n$ denote the model parameters at iteration $t$, and $g_t = \nabla f(\theta_t)$ the corresponding gradient. The Lion optimizer maintains a momentum term $m_t$, initialized as $m_0 = 0$. The update rule (without weight decay) is:

$$c_t = \beta_1 m_{t-1} + (1 - \beta_1) g_t \tag{A1}$$

$$\theta_{t+1} = \theta_t - \eta \ \text{sign}(c_t) \tag{A2}$$

$$m_t = \beta_2 m_{t-1} + (1 - \beta_2) g_t \tag{A3}$$

Here, $\eta > 0$ is the learning rate, $\beta_1, \beta_2 \in [0,1)$ are momentum coefficients, $m_t$ is an exponential moving average of past gradients, and sign $(\cdot)$ is applied elementwise. The update depends only on the direction of a momentum-interpolated gradient, resulting in uniform-magnitude parameter updates across all coordinates.

**Limiting Case: ArcGD Reduces to a Lion-Type Sign-Momentum Update**

ArcGD update rule is:

$$\Delta x = -a. T_x - b. T_x(1 - |T_x|) - c. sign(T_x)(1 - |T_x|) \tag{20}$$

If the transition constant $b$ is set to zero, ArcGD update rule reduces to

$$\Delta x = -a. T_x - \ c. sign(T_x)(1 - |T_x|) \tag{A4}$$

For $\gamma \ll 1$, $a = c = \gamma$, and $a + c \ll 1$, Equation (A4) becomes

$$\Delta x = -\gamma. T_x - \ \gamma. sign(T_x)(1 - |T_x|) \tag{A5}$$

$$\Delta x = -\gamma. T_x - \ \gamma. sign(T_x) + \ \gamma. sign(T_x). |T_x| \tag{A6}$$

Where

$$sign(T_x) = \frac{T_x}{|T_x|} \tag{A7}$$

The derivation assumes $T_x \neq 0$; at $T_x = 0$, the update evaluates to zero since sign $(0) = 0$. Using Equation (A7) into Equation (A6) will lead to

$$\Delta x = -\gamma. T_x - \ \gamma. sign(T_x) + \ \gamma. \frac{T_x}{|T_x|}. |T_x| \tag{A8}$$

$$\Delta x = -\gamma. T_x - \ \gamma. sign(T_x) + \gamma. T_x \tag{A9}$$

$$\Delta x = -\gamma . sign(T_x) \tag{A10}$$

If $T_x$ represents a momentum-based term, the resulting update rule reduces to a pure sign-based momentum update. In such limiting case, ArcGD captures the core mechanism of the Lion optimizer, where parameter updates depend only on the direction of accumulated momentum rather than its magnitude.

Moreover, ArcGD with constant $c$ (corresponding to a noisy landscape setting where exponential moving average of gradients) recovers Lion-type behavior in both limiting phases: the saturation phase ($T_x \rightarrow 1$) and the vanishing phase ($T_x \rightarrow 0$). In contrast, with adaptive $c$, ArcGD retains Lion-type behavior only in the saturation phase ($T_x \rightarrow 1$), while in the vanishing phase ($T_x \rightarrow 0$) it instead recovers standard gradient descent behavior.

## Momentum-augmented ArcGD with Lion-Style Update Structure

A momentum-augmented version of ArcGD can be constructed by adopting Lion's two-moment update structure while replacing Lion's discontinuous sign(·) operator with ArcGD's smooth, bounded formulation for constant $c$.

The direction accumulator follows Lion's $\beta_1$-path,

$$c_t = \beta_1 m_{t-1} + (1 - \beta_1) g_t \tag{A11}$$

while the momentum state is updated via Lion's $\beta_2$ rule,

$$m_t = \beta_2 m_{t-1} + (1 - \beta_2) g_t \tag{A12}$$

Instead of applying $\mathrm{sign}(c_t)$ directly, $T$-transformation for ArcGD is computed as

$$T^{(c)} = \frac{c_t}{\sqrt{1 + c_t^2}} \tag{A13}$$

Using Equation (A13) in Equation (20) will lead to

$$\Delta x = -a . T_x^{(c)} - b . T_x^{(c)} \left(1 - \left|T_x^{(c)}\right|\right) - c . sign\left(T_x^{(c)}\right) \left(1 - \left|T_x^{(c)}\right|\right) \tag{A14}$$

The momentum-augmented ArcGD recovers Lion's behavior in both limiting regimes. In the steep-magnitude limit, where $T^{(c)} \rightarrow \mathrm{sign}(c_t)$ and the transition and floor terms vanish, giving

$$\Delta x \approx -a \, \mathrm{sign}(c_t)$$

matching Lion with learning rate approximately $a$. In the vanishing-magnitude limit, where $T^{(c)} \to 0$, and the floor term dominates, yielding

$$\Delta x \approx -c\,\mathrm{sign}(c_t),$$

i.e., a scaled version of Lion's sign-momentum update. Between these extremes, ArcGD provides a smooth mid-range response, shaped by the parameter $b$, which Lion lacks. As a result, the momentum-augmented ArcGD acts as a strictly smoother and more controllable analogue of Lion, retaining Lion's decisive sign-based behavior while offering continuous interpolation across gradient magnitudes and explicit hyperparameter control for different gradient-phase regimes.

Interestingly, it can also be noted that, in the conceptual limit as $\beta_1 \to 1^-$, the direction accumulator becomes dominated by the $\beta_2$-momentum state, i.e., $c_t \to m_{t-1}$. When Lion and momentum-augmented ArcGD share the same $\beta_2$ parameter, both methods therefore rely on the same underlying momentum dynamics. With such limit, ArcGD with constant $c$ continues to recover Lion in both the steep-magnitude phase (where $T^{(c)} \to \mathrm{sign}(c_t)$) and the vanishing-magnitude phase (where $T^{(c)} \approx c\,\mathrm{sign}(c_t)$), while still providing its characteristic smooth transition in the intermediate region. Thus, even as $\beta_1$ approaches 1, ArcGD retains Lion's sign-momentum behavior in the limiting phases.

# Appendix B (Evaluation 1: Configuration A)

This appendix presents the detailed results of each run for all test cases under evaluation 1 (configuration A). Individual run data are summarized in tabular form, while the evolution of the loss function and gradient norm is illustrated through graphs. Additionally, for the two-dimensional test case (A2), the iteration-wise changes in the vector components, $(x_1, x_2)$ are also provided to give visual insight into the optimization trajectory.

## 8.1 Test: A2

Table A2: Detailed results of individual runs for the two-dimensional test case (A2).

| Run 2D | Optimizer | EMA_Patience_Stopping & Low_Loss | Iterations | Final_Loss | Final_Gradient_Norm | Distance_to_Minima | Time(s) |
|---|---|---|---|---|---|---|---|
| 1 | ArcGD | TRUE | 3408 | 2.38E-03 | 1.08E-03 | 3.43E-05 | 0.19 |
| 1 | ADAM | TRUE | 9352 | 2.01E-04 | 3.87E-04 | 3.94E-04 | 0.49 |
| 2 | ArcGD | TRUE | 3791 | -4.91E-04 | 1.11E-04 | 5.58E-06 | 0.17 |
| 2 | ADAM | TRUE | 10352 | -2.38E-03 | 3.40E-04 | 1.27E-04 | 0.44 |
| 3 | ArcGD | TRUE | 2691 | 1.22E-03 | 1.46E-04 | 8.54E-06 | 0.13 |
| 3 | ADAM | TRUE | 9952 | 3.49E-04 | 3.73E-04 | 3.95E-04 | 0.42 |
| 4 | ArcGD | TRUE | 1969 | -1.11E-03 | 7.17E-04 | 9.93E-06 | 0.11 |
| 4 | ADAM | TRUE | 6551 | 8.50E-04 | 2.10E-04 | 4.22E-07 | 0.31 |
| 5 | ArcGD | TRUE | 2329 | 1.94E-04 | 8.75E-04 | 2.52E-06 | 0.14 |
| 5 | ADAM | TRUE | 12484 | -3.19E-04 | 1.35E-04 | 1.12E-05 | 0.64 |
| 6 | ArcGD | TRUE | 2610 | -1.23E-03 | 7.51E-04 | 1.08E-05 | 0.13 |
| 6 | ADAM | TRUE | 11090 | 1.03E-03 | 4.40E-04 | 1.20E-05 | 0.55 |
| 7 | ArcGD | TRUE | 3203 | -3.62E-04 | 2.81E-04 | 1.13E-05 | 0.16 |
| 7 | ADAM | TRUE | 6658 | 1.39E-03 | 5.55E-04 | 5.09E-04 | 0.27 |
| 8 | ArcGD | TRUE | 2320 | -2.32E-04 | 3.24E-04 | 6.11E-06 | 0.11 |
| 8 | ADAM | TRUE | 8851 | -8.40E-04 | 9.26E-05 | 1.84E-05 | 0.46 |
| 9 | ArcGD | TRUE | 2193 | -1.07E-03 | 5.16E-04 | 7.36E-07 | 0.11 |
| 9 | ADAM | TRUE | 9320 | -7.22E-04 | 5.10E-04 | 6.22E-04 | 0.49 |
| 10 | ArcGD | TRUE | 3510 | -3.36E-05 | 2.06E-04 | 3.22E-06 | 0.24 |
| 10 | ADAM | TRUE | 9793 | 3.13E-04 | 5.38E-04 | 1.79E-04 | 0.64 |

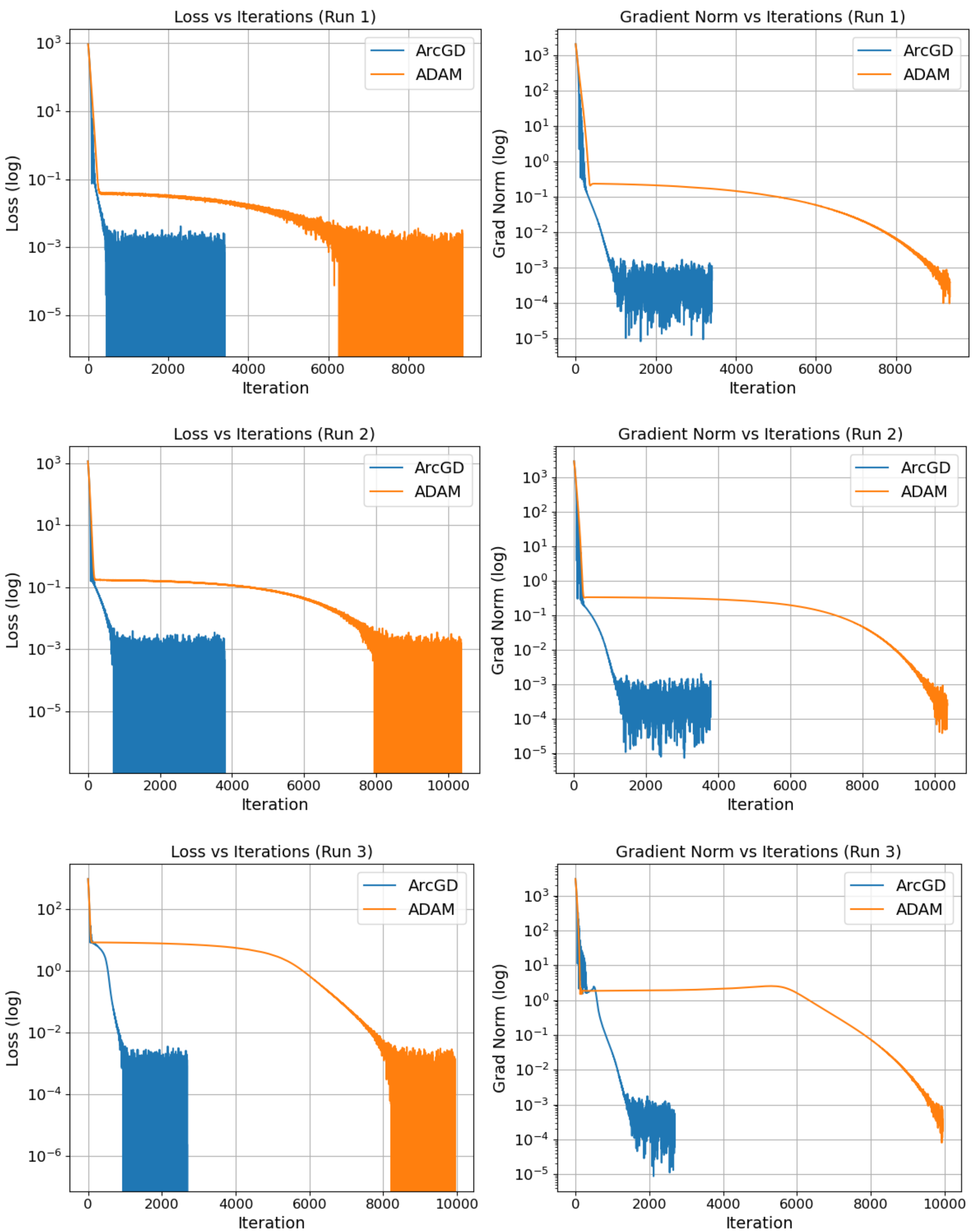

Figure A2.1: Loss and gradient norm for Run 1 to Run 3

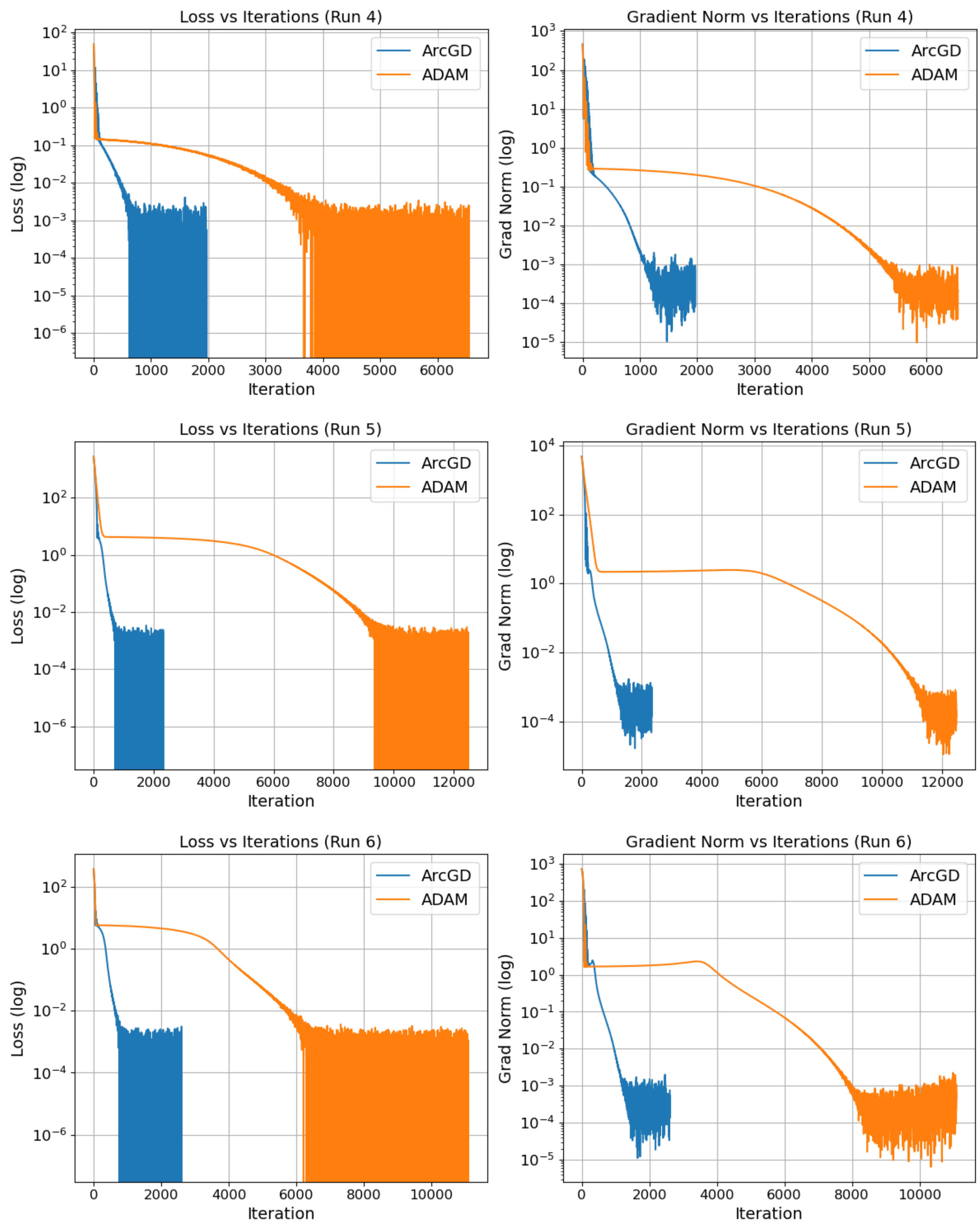

Figure A2.2: Loss and gradient norm for Run 4 to Run 6

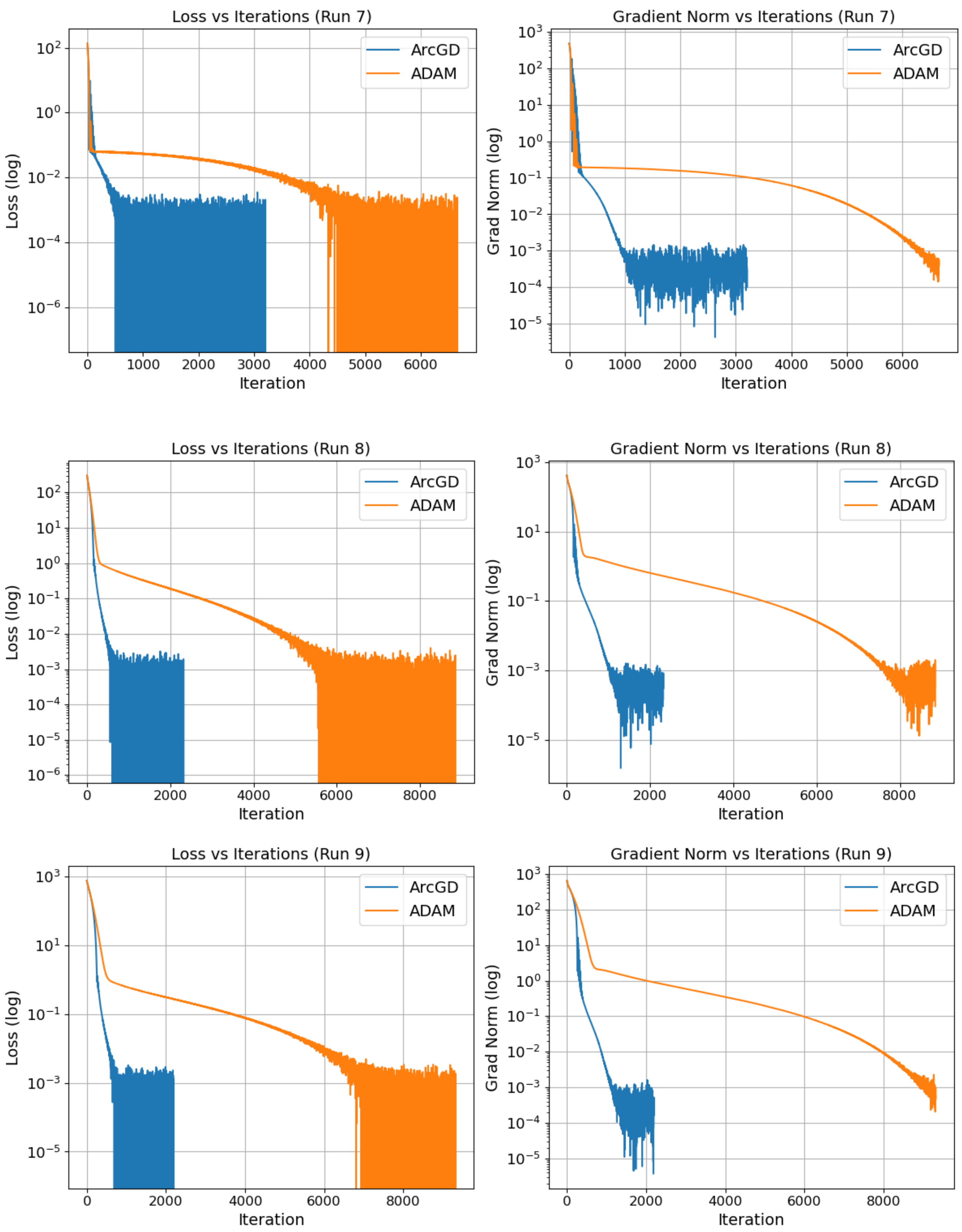

Figure A2.3: Loss and gradient norm for Run 7 to Run 9

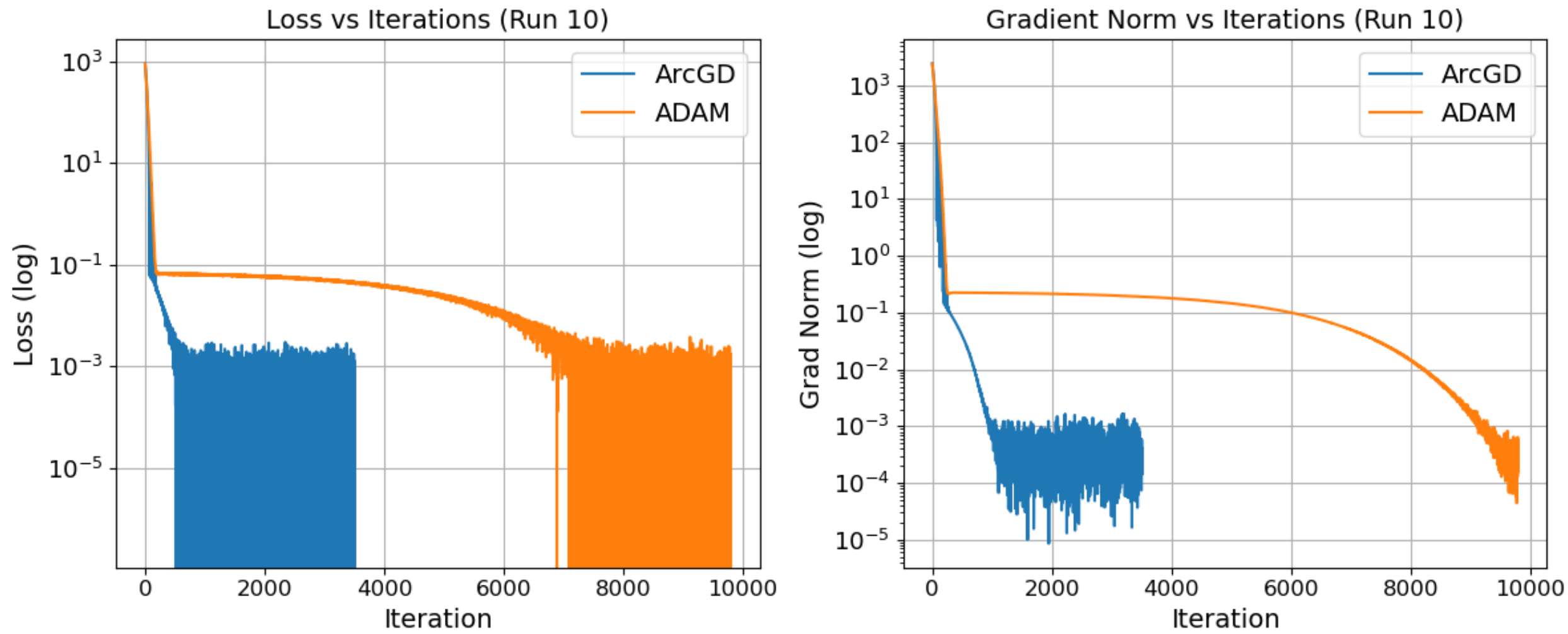

Figure A2.4: Loss and gradient norm for Run 10

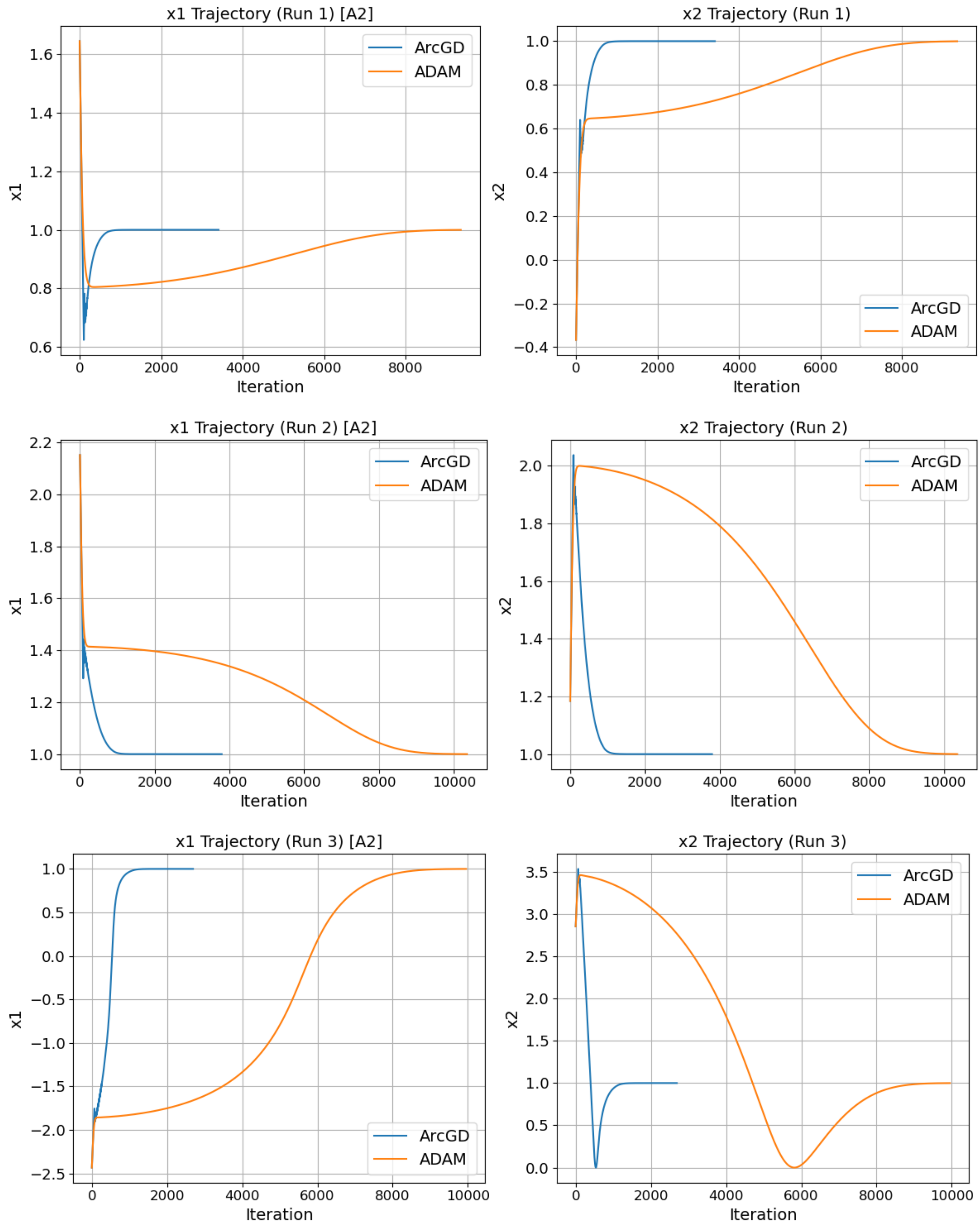

Figure A2.5: Change in $(x_1, x_2)$ for Run 1 to Run 3

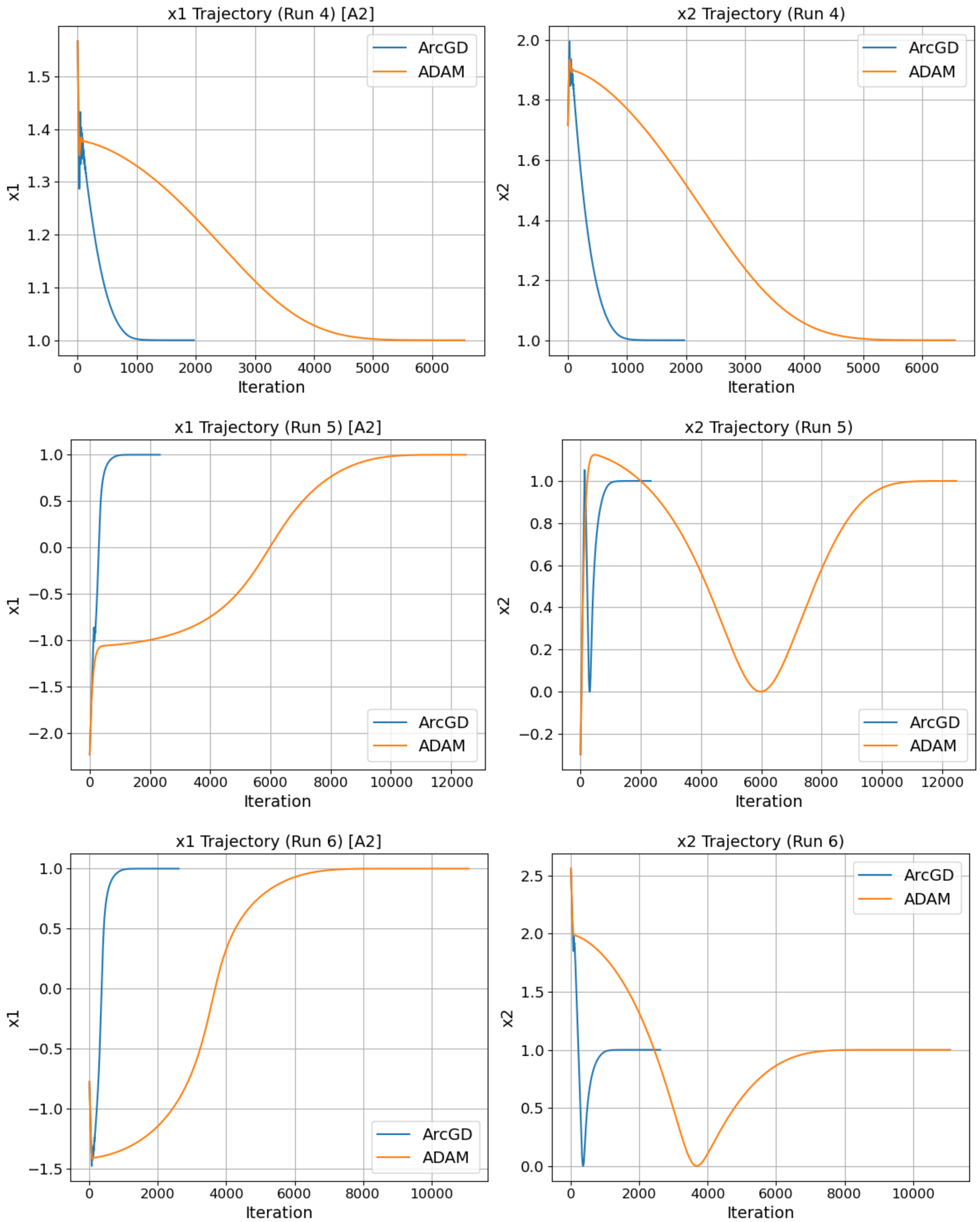

Figure A2.6: Change in $(x_1, x_2)$ for Run 4 to Run 6

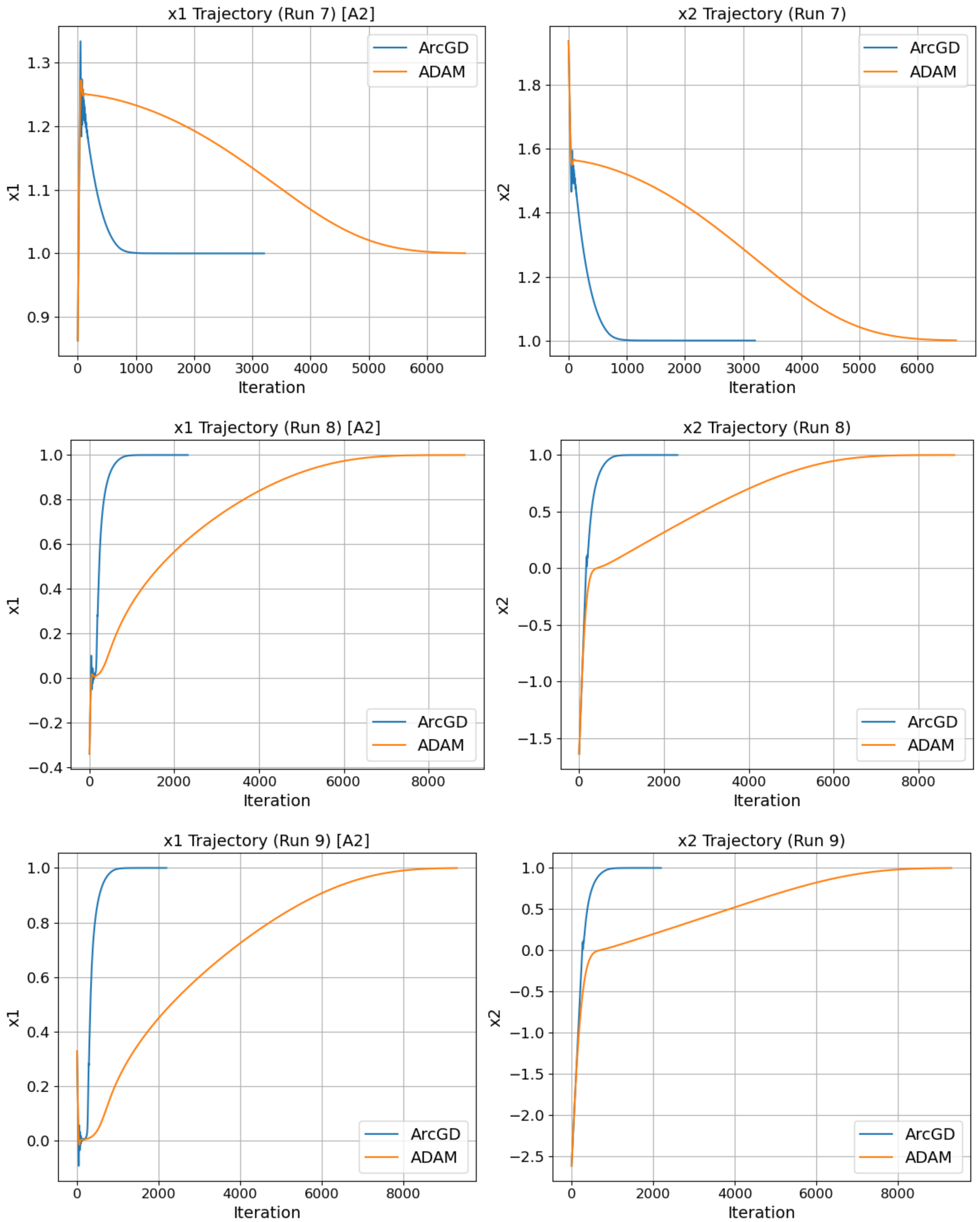

Figure A2.7: Change in $(x_1, x_2)$ for Run 7 to Run 9

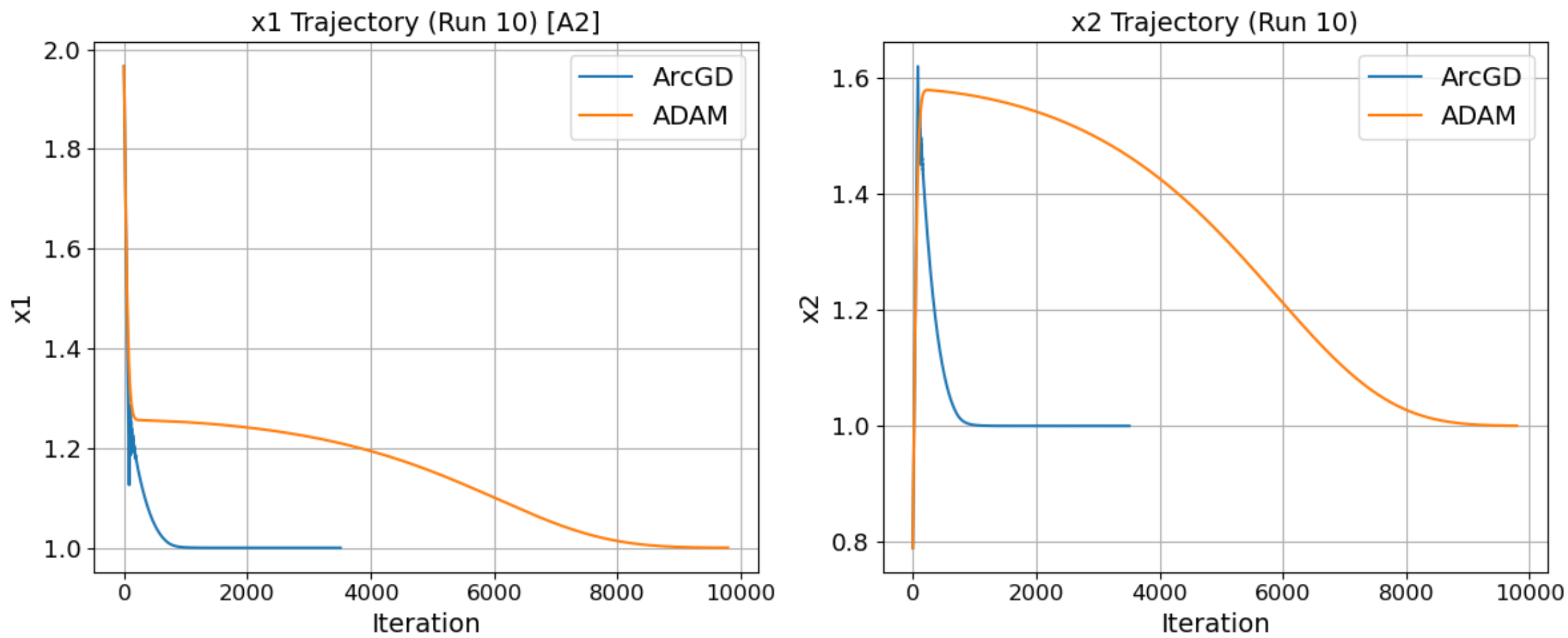

Figure A2.8 Change in $(x_1, x_2)$ for Run 10

## 8.2 Test: A10

Table A10: Detailed results of individual runs for the ten-dimensional test case (A10).

| Run 10D | Optimizer | EMA_Patience_Stopping & Low_Loss | Iterations | Final_Loss | Final_Gradient_Norm | Distance_to_Minima | Time(s) |
|---|---|---|---|---|---|---|---|
| 1 | ArcGD | TRUE | 2231 | 1.10E-04 | 4.24E-03 | 1.87E-06 | 0.17 |
| 1 | ADAM | TRUE | 11729 | 4.74E-05 | 3.53E-03 | 3.86E-05 | 0.68 |
| 2 | ArcGD | FALSE | 2532 | 3.99E+00 | 3.29E-03 | 6.30E-01 | 0.13 |
| 2 | ADAM | FALSE | 11290 | 3.99E+00 | 9.44E-01 | 6.30E-01 | 0.46 |
| 3 | ArcGD | TRUE | 3864 | 1.73E-03 | 1.26E-03 | 1.69E-06 | 0.2 |
| 3 | ADAM | TRUE | 11716 | 4.81E-04 | 1.25E-02 | 3.47E-05 | 0.55 |
| 4 | ArcGD | TRUE | 4003 | -2.30E-04 | 1.77E-03 | 3.95E-06 | 0.42 |
| 4 | ADAM | TRUE | 9980 | -1.41E-03 | 4.55E-02 | 1.04E-05 | 0.67 |
| 5 | ArcGD | TRUE | 2398 | -1.05E-03 | 3.01E-03 | 3.88E-06 | 0.12 |
| 5 | ADAM | FALSE | 9991 | 3.99E+00 | 3.33E-02 | 6.30E-01 | 0.42 |
| 6 | ArcGD | TRUE | 3685 | 2.49E-03 | 1.98E-03 | 7.12E-06 | 0.6 |
| 6 | ADAM | TRUE | 12488 | 9.67E-04 | 2.39E-03 | 2.90E-05 | 1.18 |
| 7 | ArcGD | TRUE | 2388 | -4.62E-04 | 1.40E-03 | 3.33E-06 | 0.2 |
| 7 | ADAM | TRUE | 13039 | 1.30E-03 | 4.73E-02 | 2.23E-05 | 1.26 |
| 8 | ArcGD | TRUE | 2243 | -3.71E-04 | 2.16E-03 | 2.06E-06 | 0.14 |
| 8 | ADAM | FALSE | 9778 | 3.99E+00 | 1.27E-02 | 6.30E-01 | 0.49 |
| 9 | ArcGD | TRUE | 2366 | 7.67E-04 | 2.38E-03 | 1.24E-05 | 0.12 |
| 9 | ADAM | TRUE | 12086 | 8.38E-04 | 3.56E-02 | 1.32E-05 | 0.74 |
| 10 | ArcGD | FALSE | 2713 | 3.99E+00 | 6.39E-03 | 6.30E-01 | 0.21 |
| 10 | ADAM | FALSE | 11754 | 3.99E+00 | 1.87E+00 | 6.30E-01 | 1.58 |

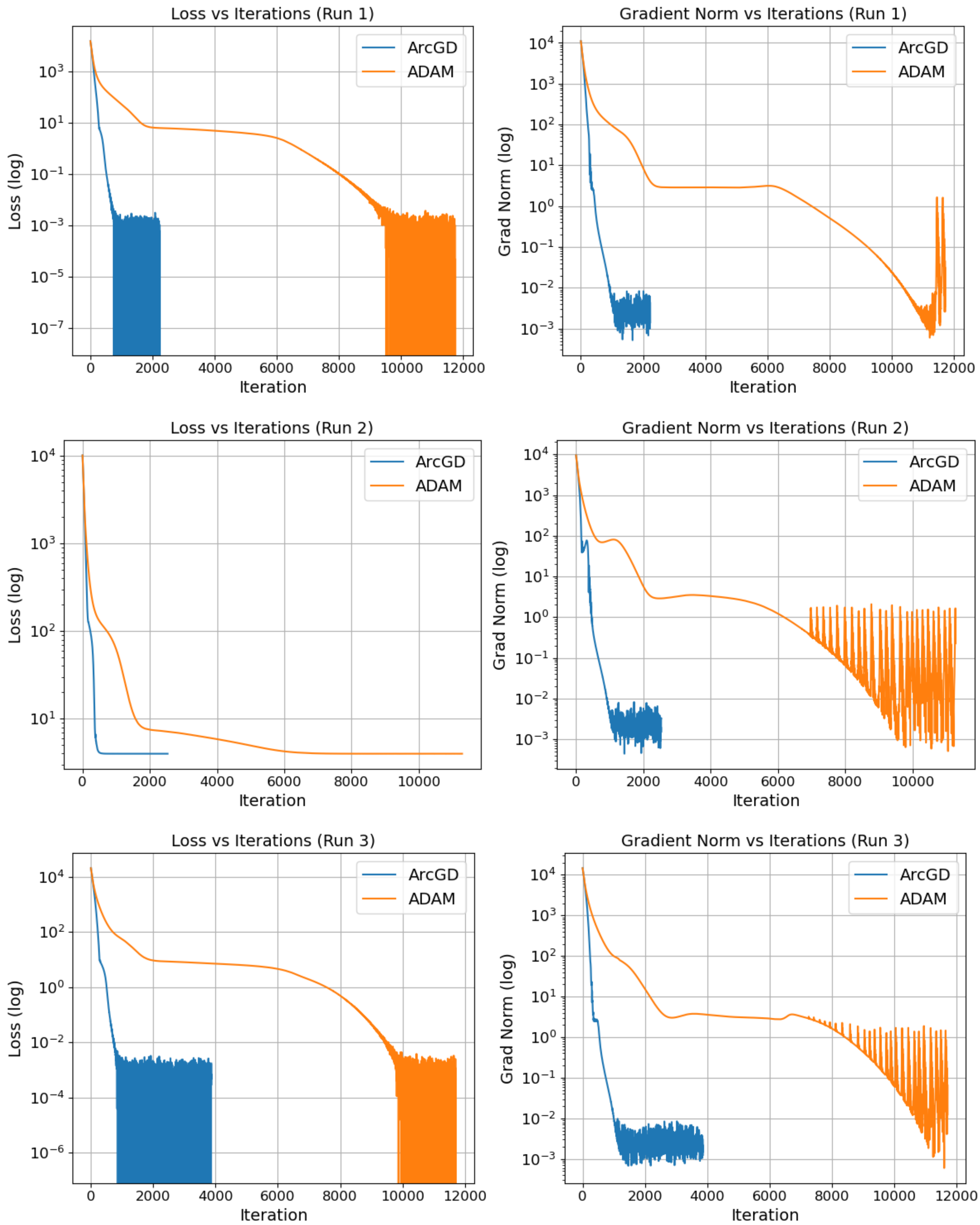

Figure A10.1: Loss and gradient norm for Run 1 to Run 3

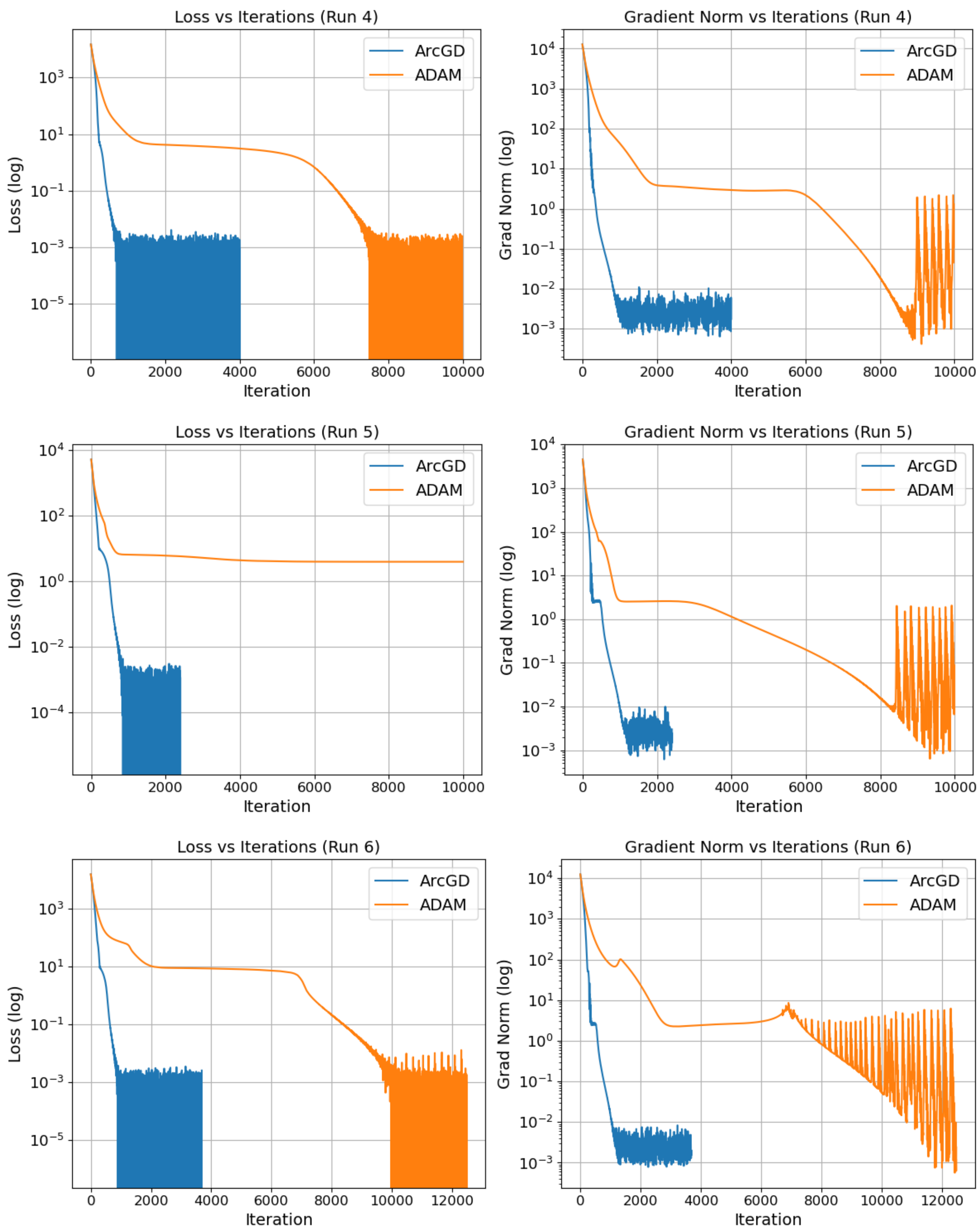

Figure A10.2: Loss and gradient norm for Run 4 to Run 6

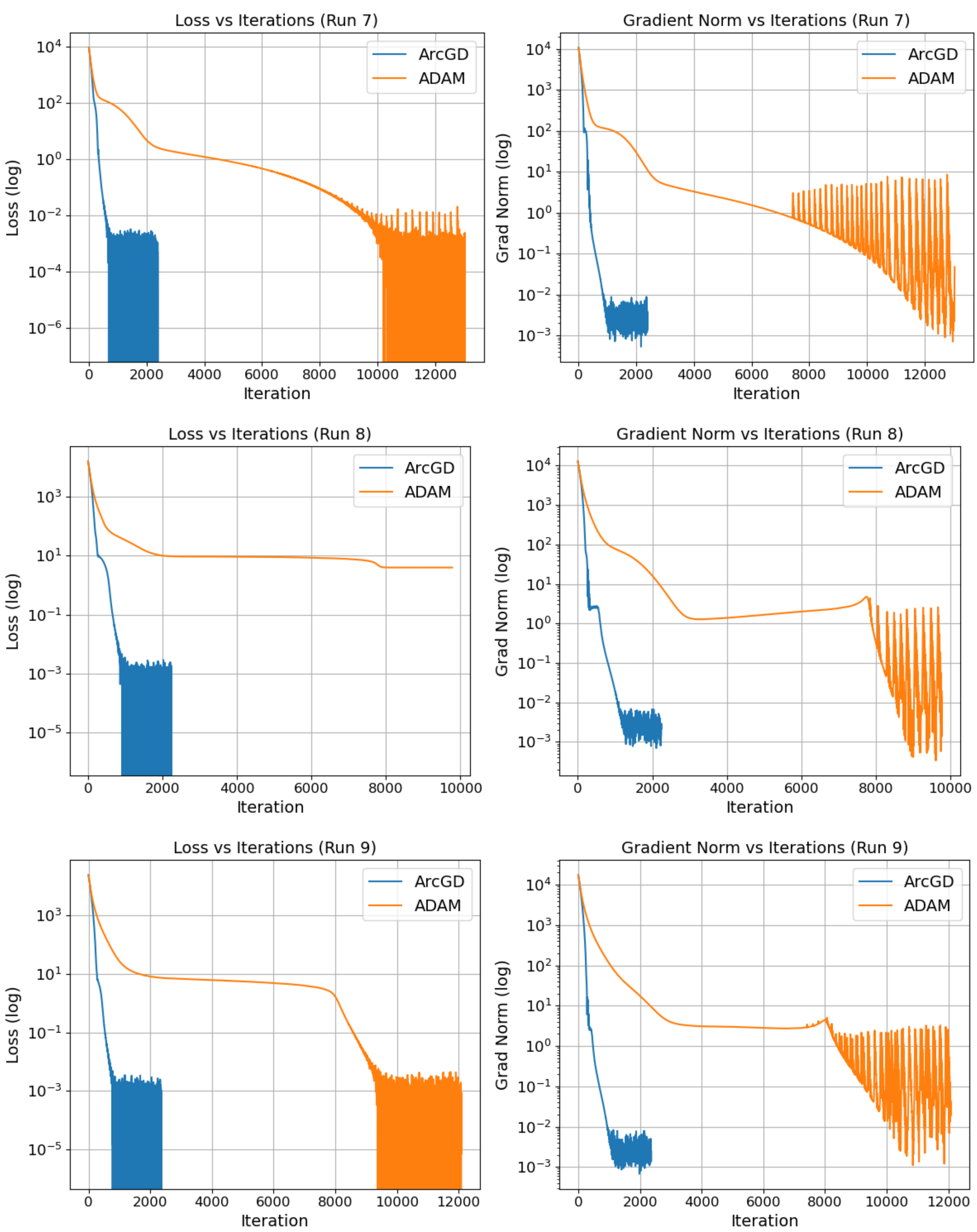

Figure A10.3: Loss and gradient norm for Run 7 to Run 9

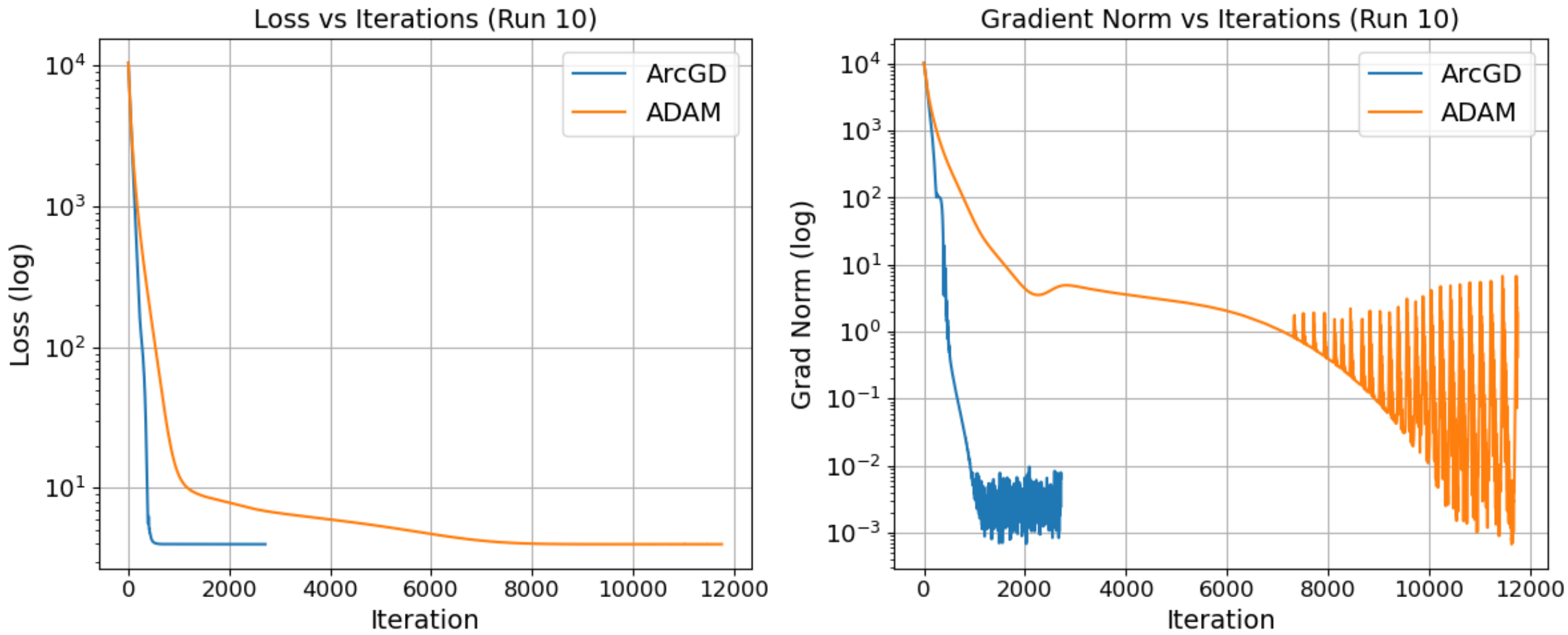

Figure A10.4: Loss and gradient norm for Run 10

## 8.3 Test: A100

Table A100: Detailed results of individual runs for the hundred-dimensional test case (A100).

| Run 100D | Optimizer | EMA_Patience_Stopping & Low_Loss | Iterations | Final_Loss | Final_Gradient_Norm | Distance_to_Minima | Time(s) |
|---|---|---|---|---|---|---|---|
| 1 | ArcGD | TRUE | 5201 | 2.79E-04 | 2.05E-02 | 1.35E-06 | 0.59 |
| 1 | ADAM | TRUE | 13836 | 3.40E-03 | 3.87E+00 | 2.24E-04 | 0.96 |
| 2 | ArcGD | TRUE | 3357 | -1.07E-03 | 1.25E-02 | 1.75E-06 | 0.25 |
| 2 | ADAM | TRUE | 12748 | 7.42E-04 | 1.22E+00 | 7.38E-05 | 1.01 |
| 3 | ArcGD | TRUE | 2808 | -7.05E-04 | 1.27E-02 | 1.60E-06 | 0.24 |
| 3 | ADAM | TRUE | 12874 | -2.21E-04 | 4.35E-01 | 2.72E-05 | 1.74 |
| 4 | ArcGD | TRUE | 4917 | -4.55E-04 | 1.46E-02 | 1.48E-06 | 0.69 |
| 4 | ADAM | TRUE | 13204 | 1.57E-02 | 6.82E+00 | 4.15E-04 | 2.15 |
| 5 | ArcGD | TRUE | 4714 | 5.25E-04 | 1.24E-02 | 8.79E-07 | 0.57 |
| 5 | ADAM | TRUE | 15017 | 4.27E-04 | 1.59E+00 | 9.53E-05 | 1.65 |
| 6 | ArcGD | TRUE | 4202 | -3.62E-04 | 1.62E-02 | 1.18E-06 | 0.31 |
| 6 | ADAM | TRUE | 12395 | 7.36E-04 | 2.96E-01 | 7.64E-05 | 1.43 |
| 7 | ArcGD | FALSE | 4373 | 3.99E+00 | 1.80E-02 | 1.99E-01 | 0.76 |
| 7 | ADAM | FALSE | 15167 | 3.99E+00 | 1.23E+00 | 1.99E-01 | 1.43 |
| 8 | ArcGD | TRUE | 4692 | 1.06E-04 | 1.39E-02 | 1.17E-06 | 0.31 |
| 8 | ADAM | TRUE | 13183 | 5.40E-03 | 4.43E+00 | 2.70E-04 | 0.68 |
| 9 | ArcGD | TRUE | 3908 | 5.92E-04 | 2.18E-02 | 1.58E-06 | 0.24 |
| 9 | ADAM | TRUE | 12888 | 4.31E-03 | 3.78E+00 | 2.37E-04 | 0.83 |
| 10 | ArcGD | TRUE | 5604 | -1.48E-03 | 1.48E-02 | 1.00E-06 | 0.36 |
| 10 | ADAM | TRUE | 14746 | 9.19E-03 | 6.34E+00 | 3.69E-04 | 1.18 |

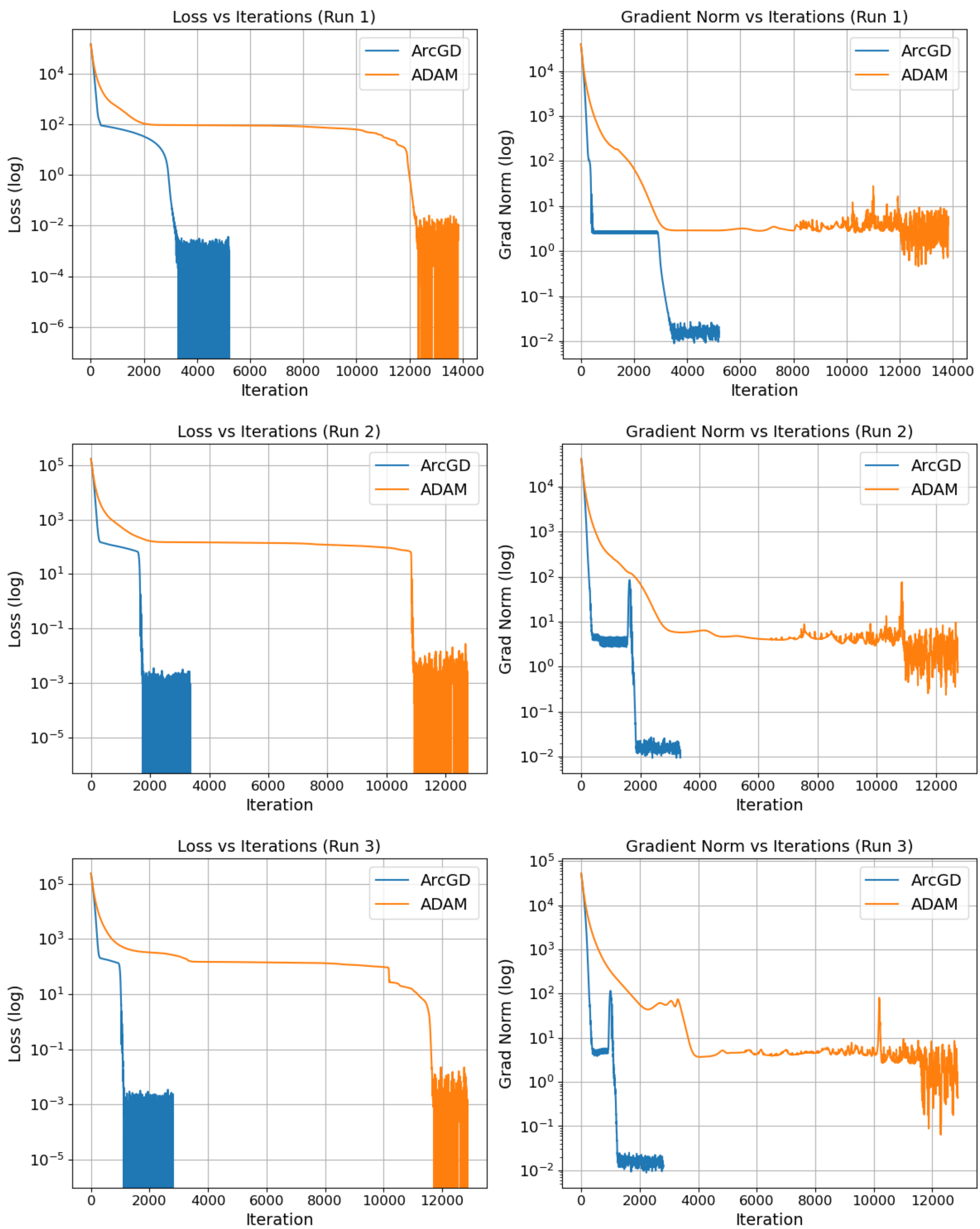

Figure A100.1: Loss and gradient norm for Run 1 to Run 3

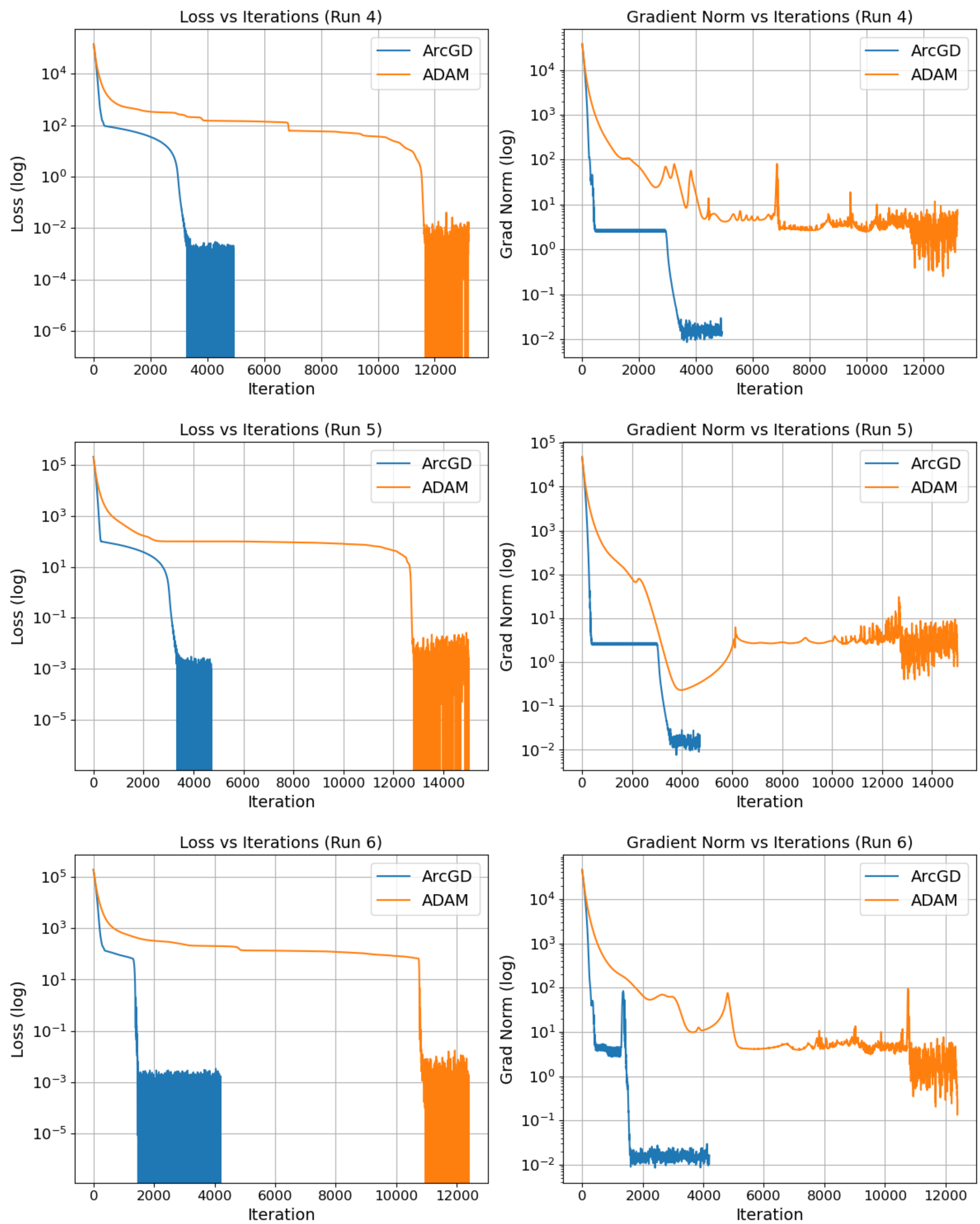

Figure A100.2: Loss and gradient norm for Run 4 to Run 6

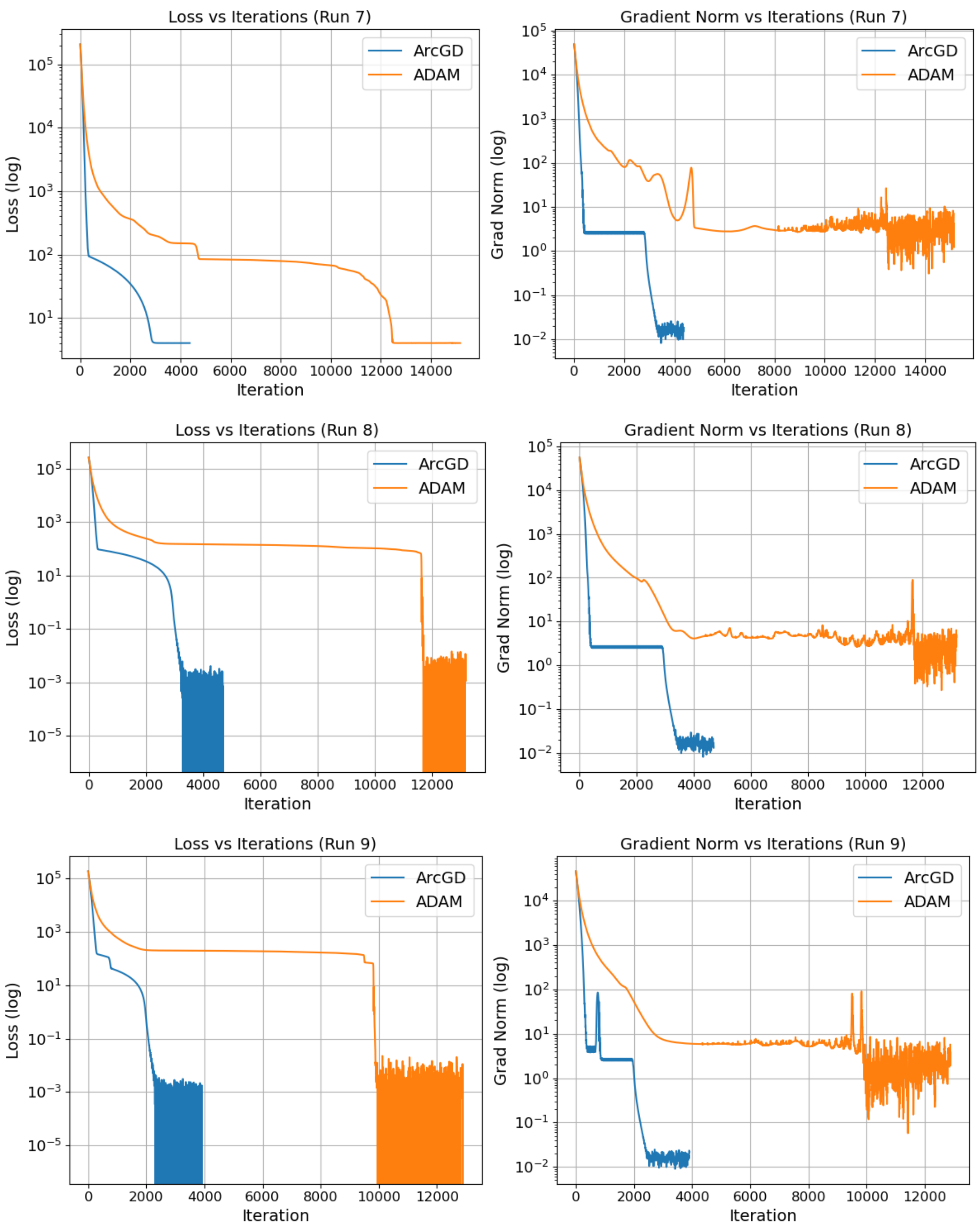

Figure A100.3: Loss and gradient norm for Run 7 to Run 9

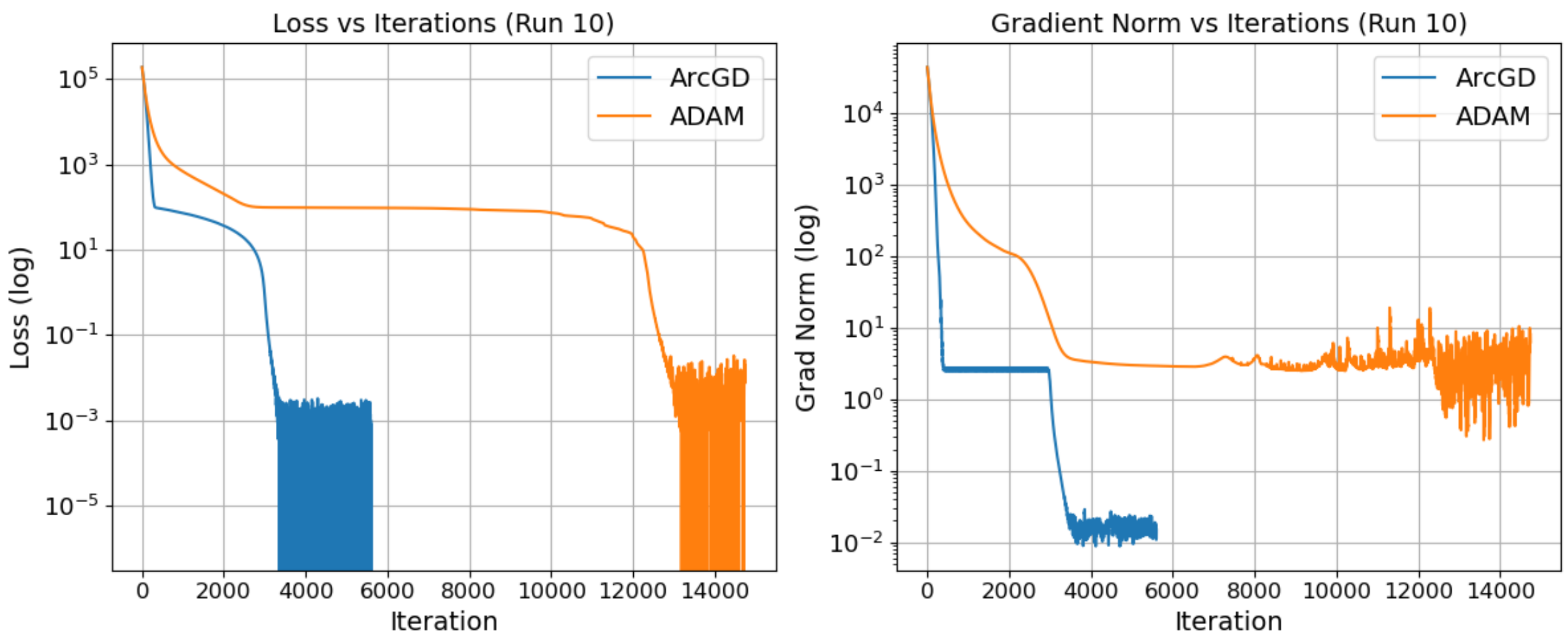

Figure A100.4: Loss and gradient norm for Run 10

## 8.4 Test: A1000

Table A1000: Detailed results of individual runs for the thousand-dimensional test case (A1000).

| Run 1000D | Optimizer | EMA_Patience_Stopping & Low_Loss | Iterations | Final_Loss | Final_Gradient_Norm | Distance_to_Minima | Time(s) |
|---|---|---|---|---|---|---|---|
| 1 | ArcGD | TRUE | 7457 | -2.85E-04 | 5.13E-02 | 1.01E-06 | 0.87 |
| 1 | ADAM | TRUE | 15640 | 5.21E-02 | 1.34E+01 | 2.44E-04 | 1.34 |
| 2 | ArcGD | TRUE | 8803 | 2.23E-03 | 5.38E-02 | 1.13E-06 | 0.86 |
| 2 | ADAM | TRUE | 15557 | 5.70E-02 | 1.42E+01 | 2.58E-04 | 1.8 |
| 3 | ArcGD | TRUE | 8141 | 9.40E-04 | 5.49E-02 | 1.15E-06 | 0.91 |
| 3 | ADAM | TRUE | 14968 | 7.48E-02 | 1.64E+01 | 2.99E-04 | 1.54 |
| 4 | ArcGD | TRUE | 13578 | 1.10E-03 | 5.46E-02 | 1.09E-06 | 1.79 |
| 4 | ADAM | TRUE | 15410 | 1.34E-01 | 2.18E+01 | 3.92E-04 | 2.08 |
| 5 | ArcGD | TRUE | 8483 | -1.05E-03 | 5.01E-02 | 1.35E-06 | 1.32 |
| 5 | ADAM | TRUE | 15174 | 7.58E-02 | 1.64E+01 | 3.00E-04 | 2.01 |
| 6 | ArcGD | TRUE | 7897 | 1.34E-03 | 5.14E-02 | 1.04E-06 | 1.54 |
| 6 | ADAM | FALSE | 14981 | 4.03E+00 | 1.18E+01 | 6.30E-02 | 4.23 |
| 7 | ArcGD | TRUE | 7453 | 1.61E-03 | 5.26E-02 | 1.08E-06 | 1.18 |
| 7 | ADAM | TRUE | 16320 | 6.40E-02 | 1.50E+01 | 2.70E-04 | 2.51 |
| 8 | ArcGD | TRUE | 12970 | 1.09E-03 | 5.17E-02 | 1.17E-06 | 1.53 |
| 8 | ADAM | TRUE | 15546 | 4.88E-02 | 1.31E+01 | 2.37E-04 | 1.59 |
| 9 | ArcGD | TRUE | 9301 | -4.15E-04 | 4.97E-02 | 1.07E-06 | 1.23 |
| 9 | ADAM | TRUE | 15201 | 4.72E-02 | 1.27E+01 | 2.33E-04 | 2.45 |
| 10 | ArcGD | TRUE | 7890 | 3.40E-04 | 5.32E-02 | 1.06E-06 | 0.97 |
| 10 | ADAM | TRUE | 17109 | 3.84E-02 | 1.18E+01 | 2.08E-04 | 2.55 |

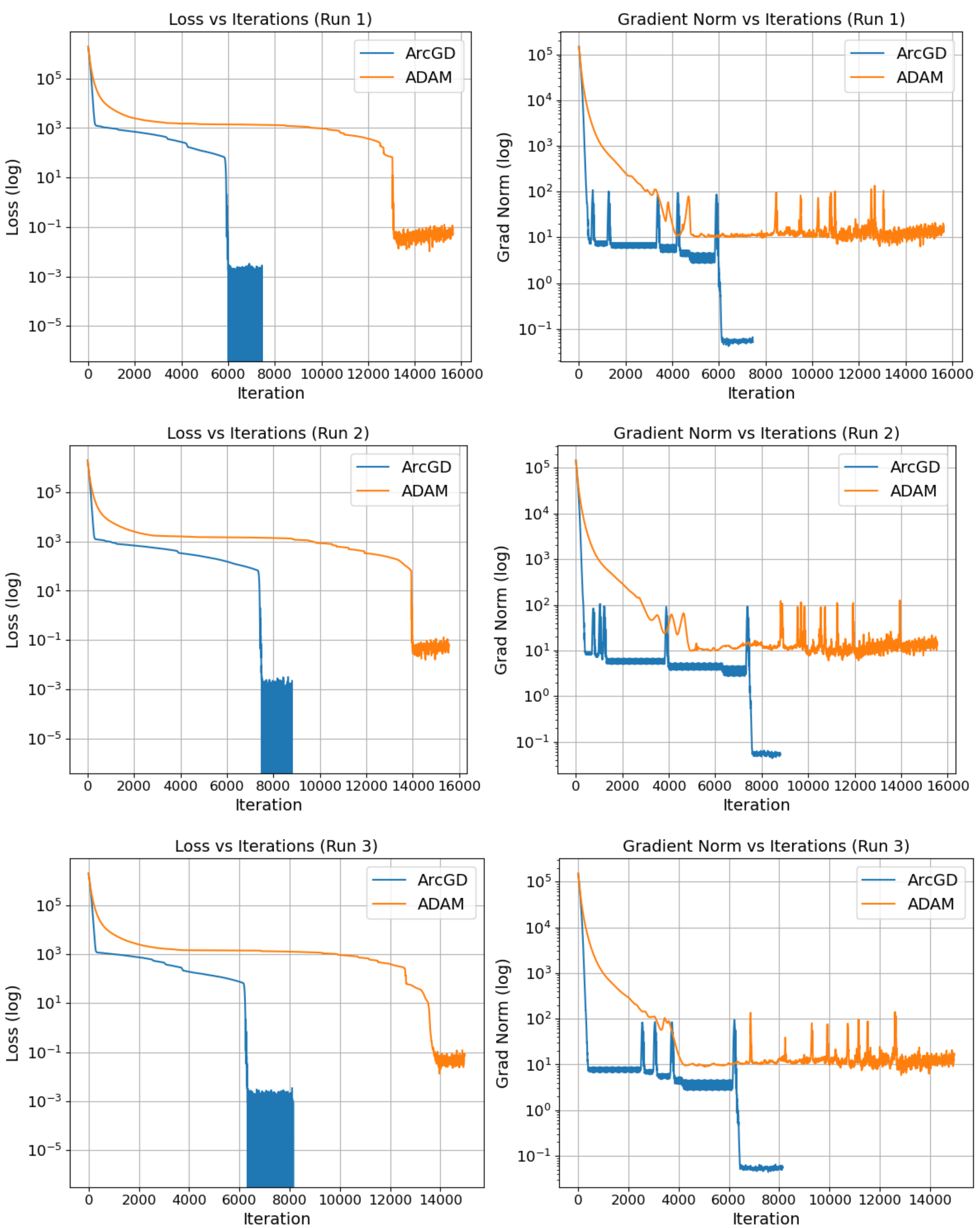

Figure A1000.1: Loss and gradient norm for Run 1 to Run 3

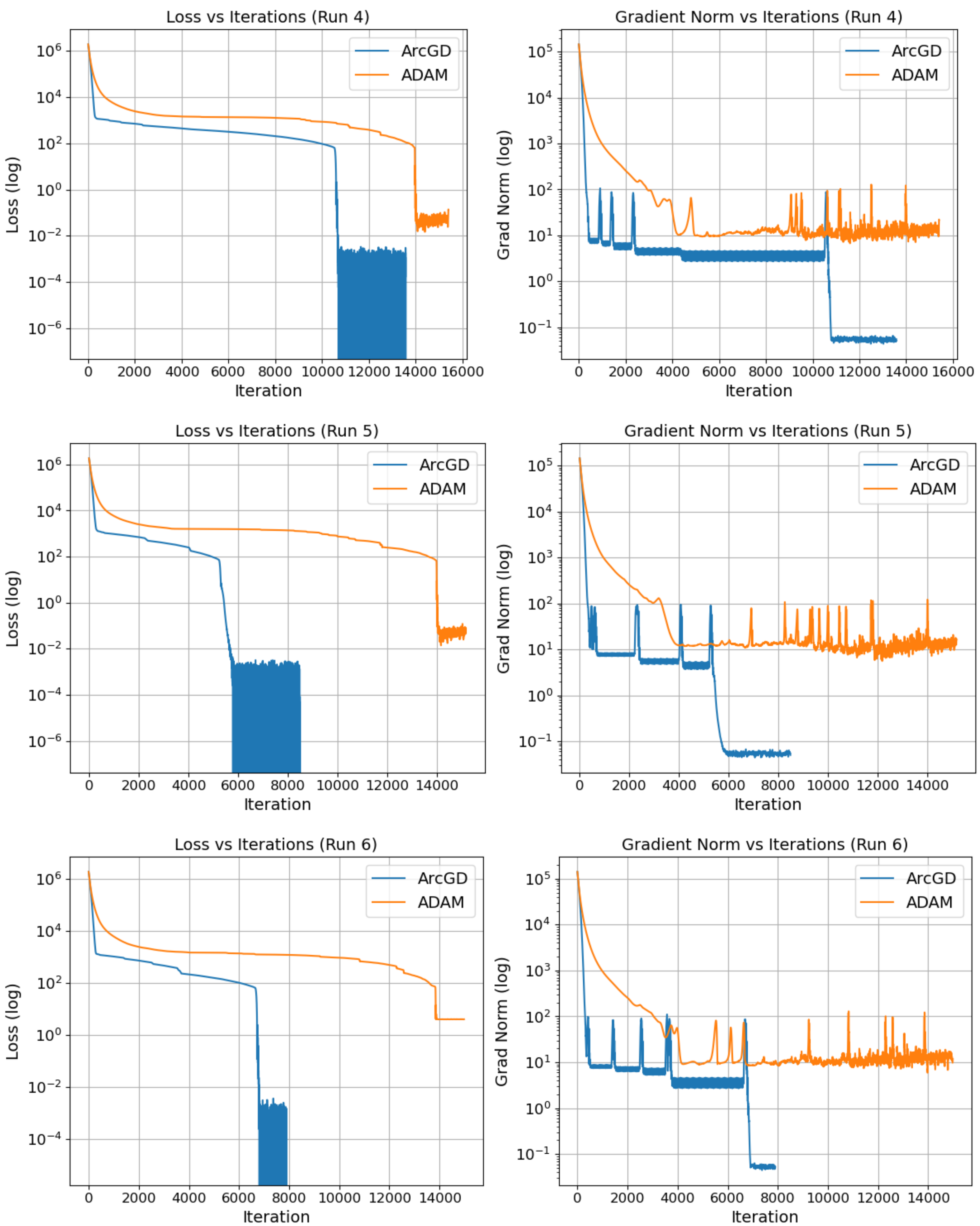

Figure A1000.2: Loss and gradient norm for Run 4 to Run 6

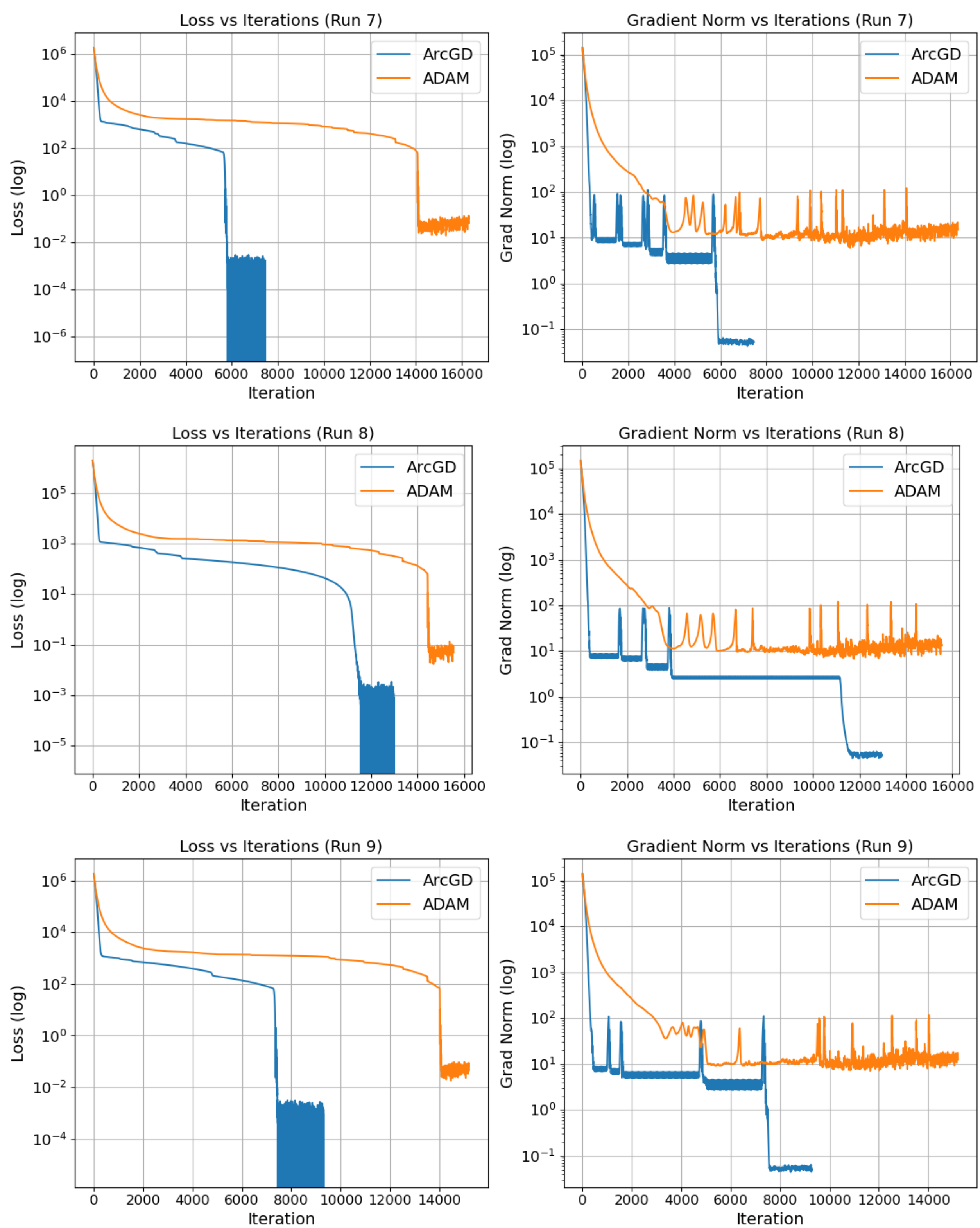

Figure A1000.3: Loss and gradient norm for Run 7 to Run 9

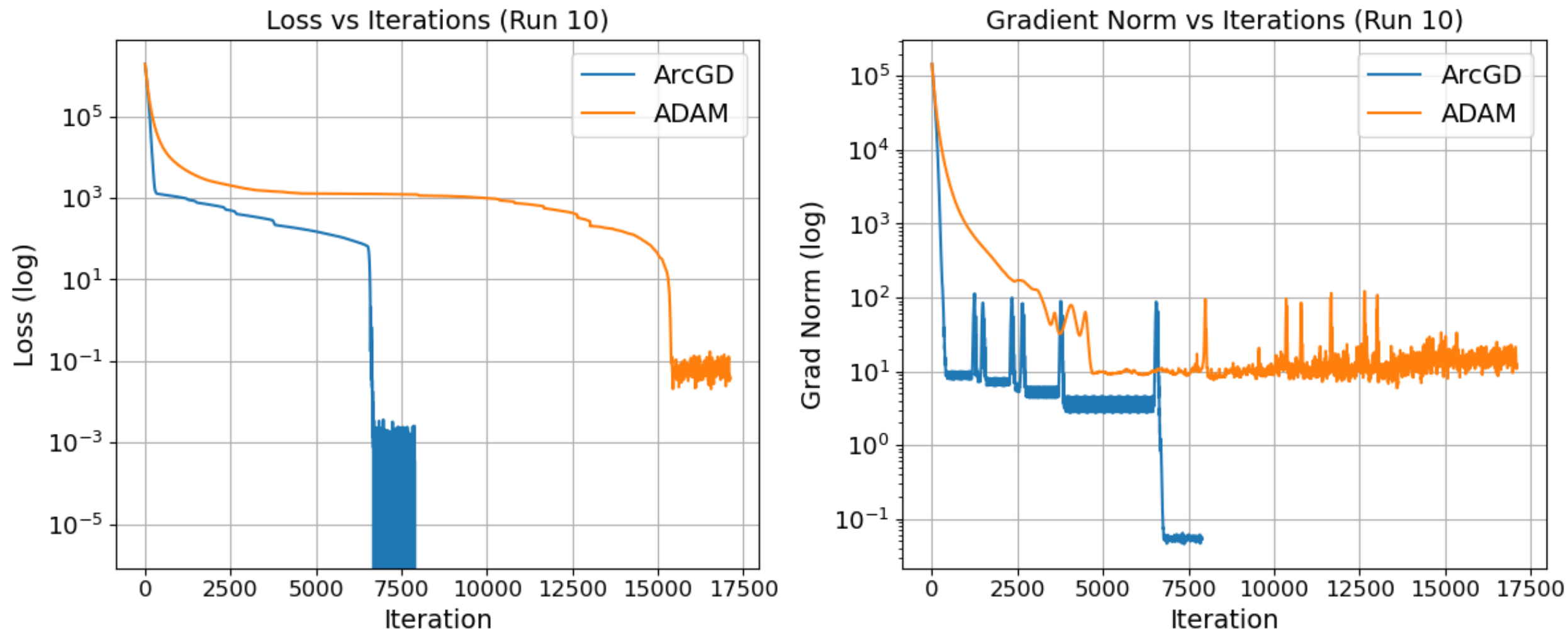

Figure A1000.4: Loss and gradient norm for Run 10

## 8.5 Test: T50000

Table T50000: Detailed results of individual runs for the fifty thousand-dimensional test case (T50000).

| Run 50000D | Optimizer | EMA_Patience_Stopping & Low_Loss | Iterations | Final_Loss | Final_Gradient_Norm | Distance_to_Minima | Time(s) |
|---|---|---|---|---|---|---|---|
| 1 | ArcGD | TRUE | 19605 | -1.53E-03 | 3.89E-01 | 1.08E-06 | 63.57 |
| 1 | ADAM | FALSE | 17728 | 3.75E+00 | 1.16E+02 | 2.90E-04 | 139.59 |
| 2 | ArcGD | TRUE | 20598 | -9.84E-06 | 3.80E-01 | 1.05E-06 | 131.13 |
| 2 | ADAM | FALSE | 17162 | 3.44E+00 | 1.11E+02 | 2.78E-04 | 77.6 |
| 3 | ArcGD | TRUE | 28775 | 1.51E-03 | 3.88E-01 | 1.08E-06 | 118.6 |
| 3 | ADAM | FALSE | 17872 | 3.56E+00 | 1.13E+02 | 2.82E-04 | 96.64 |

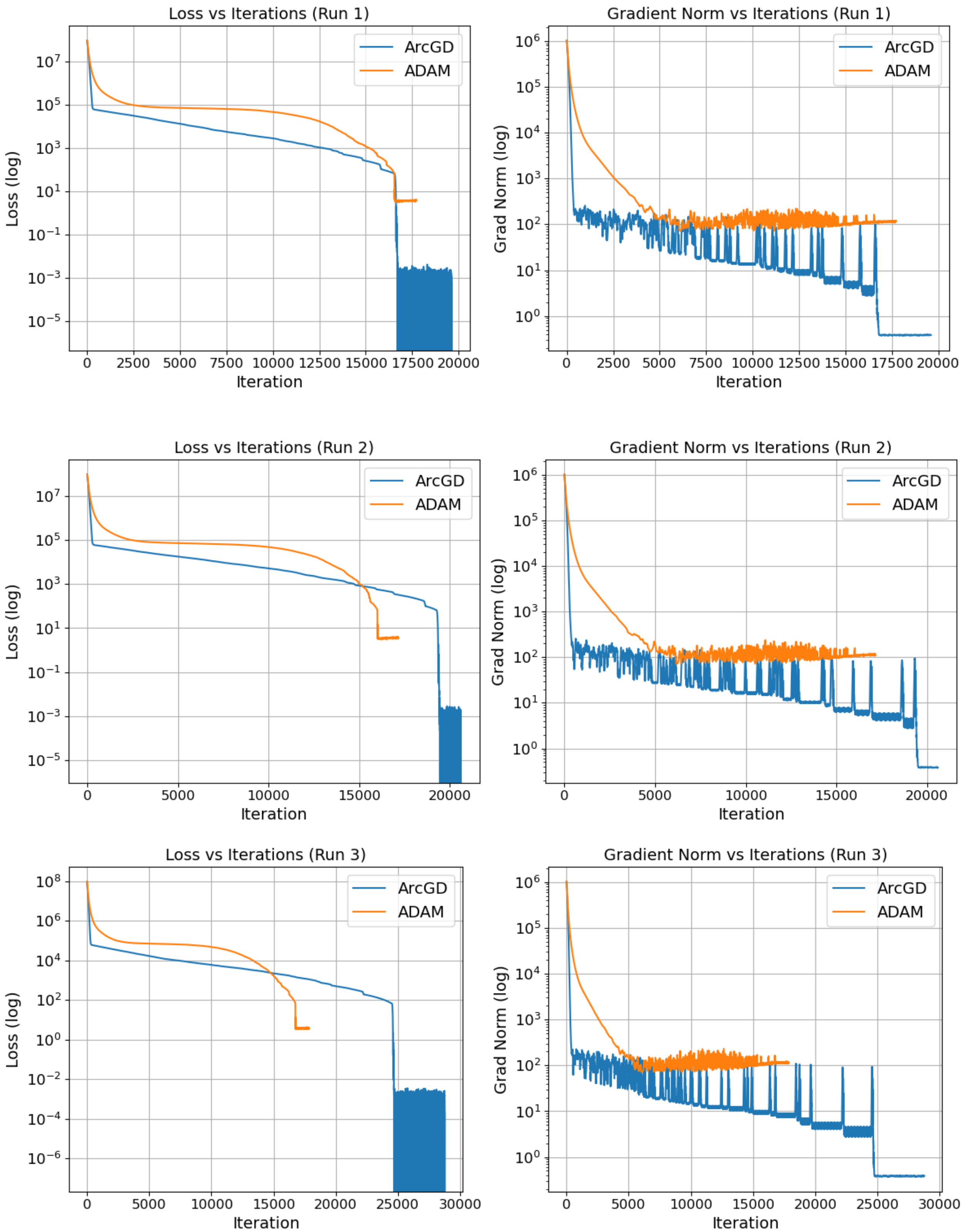

Figure T50000.1: Loss and gradient norm for Run 1 to Run 3

# Appendix C (Evaluation 1: Configuration B)

This appendix presents the detailed results of each run for all test cases under evaluation 1 (configuration B). Individual run data are summarized in tabular form, while the evolution of the loss function and gradient norm is illustrated through graphs. Additionally, for the two-dimensional test case (B2), the iteration-wise changes in the vector components, $(x_1, x_2)$ are also provided to give further insight into the optimization trajectory.

## 8.6 Test: B2

Table B2: Detailed results of individual runs for the two-dimensional test case (B2).

| Run 2D | Optimizer | EMA_Patience_Stopping & Low_Loss | Iterations | Final_Loss | Final_Gradient_Norm | Distance_to_Minima | Time(s) |
|---|---|---|---|---|---|---|---|
| 1 | ArcGD | TRUE | 8808 | 3.44E-04 | 1.64E-04 | 5.67E-06 | 0.38 |
| 1 | ADAM | TRUE | 16572 | 9.82E-04 | 5.70E-04 | 9.92E-04 | 0.7 |
| 2 | ArcGD | TRUE | 10775 | 7.79E-06 | 6.73E-04 | 1.19E-03 | 0.52 |
| 2 | ADAM | TRUE | 19601 | -2.50E-04 | 7.82E-04 | 3.21E-06 | 0.84 |
| 3 | ArcGD | TRUE | 14469 | 6.93E-04 | 2.78E-04 | 1.19E-05 | 0.95 |
| 3 | ADAM | TRUE | 19974 | -5.27E-04 | 1.62E-04 | 1.67E-05 | 0.9 |
| 4 | ArcGD | TRUE | 11477 | 3.38E-04 | 3.68E-04 | 8.55E-06 | 0.6 |
| 4 | ADAM | TRUE | 13491 | -1.49E-03 | 2.22E-04 | 9.22E-06 | 0.73 |
| 5 | ArcGD | TRUE | 10438 | 9.32E-04 | 4.64E-03 | 7.95E-03 | 0.76 |
| 5 | ADAM | TRUE | 19633 | -2.29E-03 | 4.77E-04 | 3.27E-04 | 1.03 |
| 6 | ArcGD | TRUE | 11579 | 1.58E-04 | 1.03E-03 | 1.91E-03 | 0.68 |
| 6 | ADAM | TRUE | 16521 | -1.13E-03 | 2.39E-04 | 6.85E-04 | 0.79 |
| 7 | ArcGD | TRUE | 8358 | 2.01E-04 | 1.33E-03 | 2.46E-03 | 0.62 |
| 7 | ADAM | TRUE | 14988 | -1.04E-03 | 1.54E-04 | 2.82E-06 | 0.67 |
| 8 | ArcGD | TRUE | 10680 | 1.78E-03 | 3.40E-05 | 4.19E-05 | 0.52 |
| 8 | ADAM | TRUE | 16718 | 2.05E-06 | 9.71E-03 | 6.96E-06 | 0.82 |
| 9 | ArcGD | TRUE | 12060 | 7.22E-04 | 1.96E-04 | 1.30E-05 | 0.69 |
| 9 | ADAM | TRUE | 19076 | 1.02E-04 | 2.28E-02 | 1.63E-05 | 0.8 |
| 10 | ArcGD | TRUE | 9003 | 4.26E-04 | 3.59E-04 | 3.30E-04 | 0.45 |
| 10 | ADAM | TRUE | 17860 | 1.95E-04 | 6.02E-04 | 1.77E-05 | 0.86 |

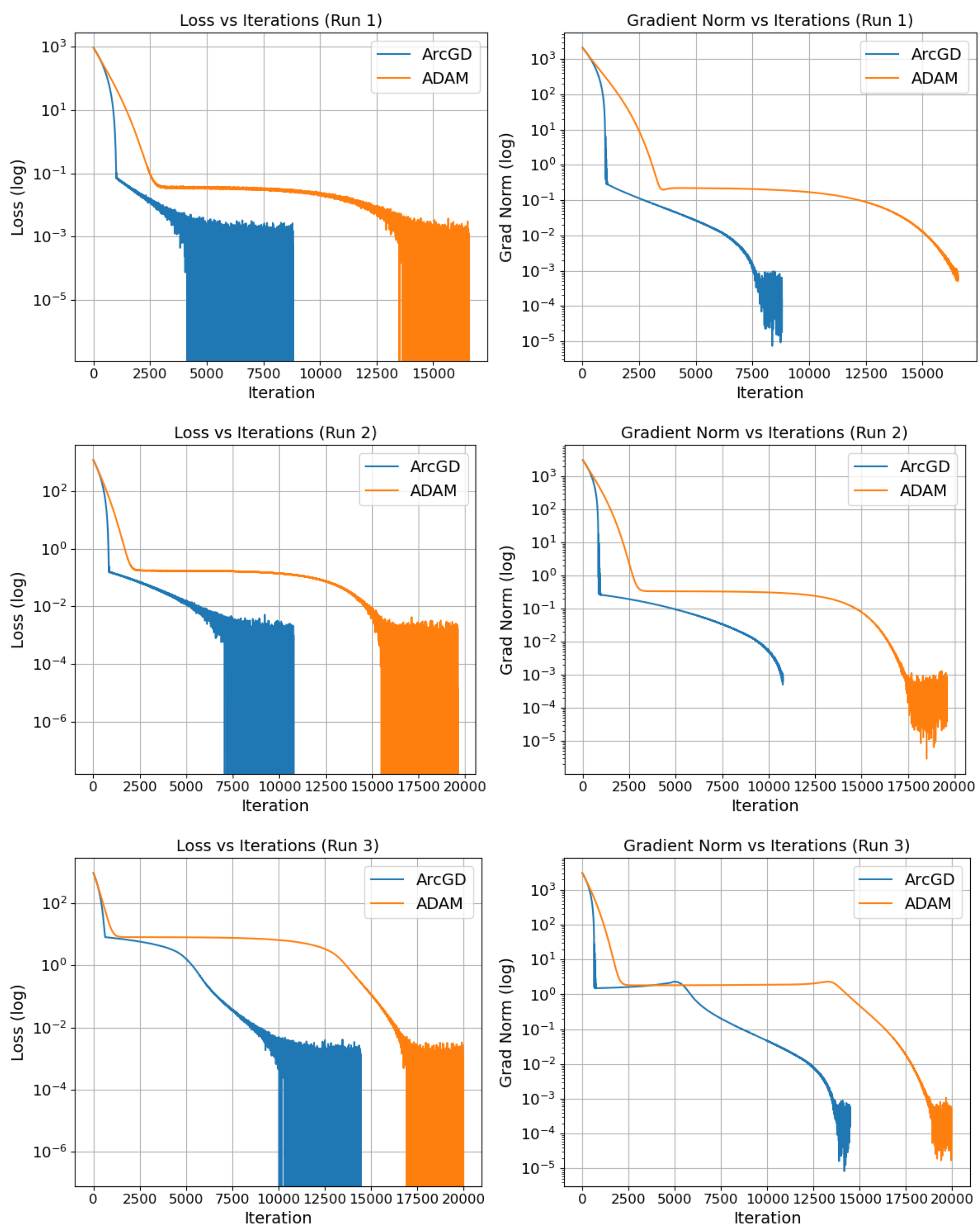

Figure B2.1: Loss and gradient norm for Run 1 to Run 3

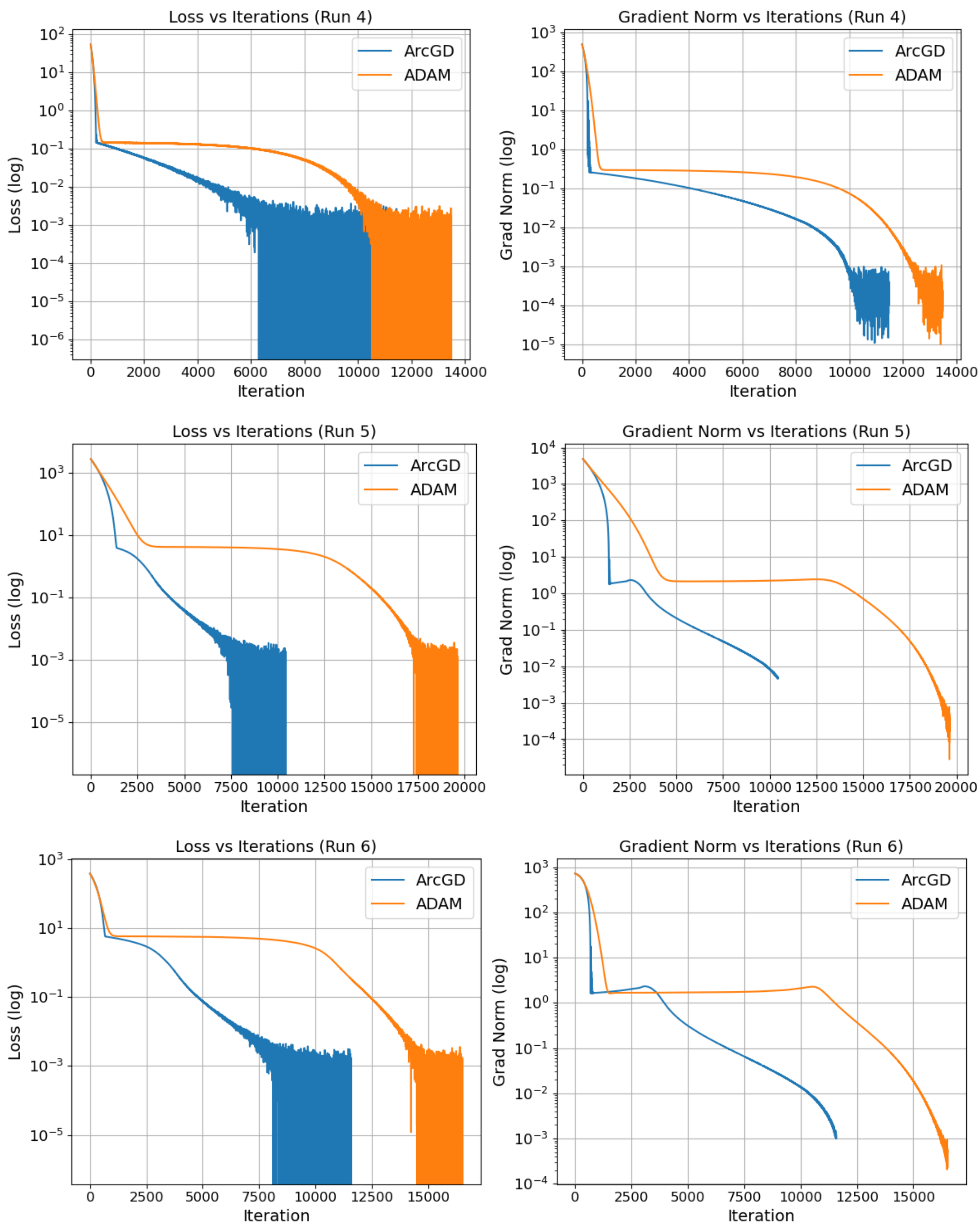

Figure B2.2: Loss and gradient norm for Run 4 to Run 6

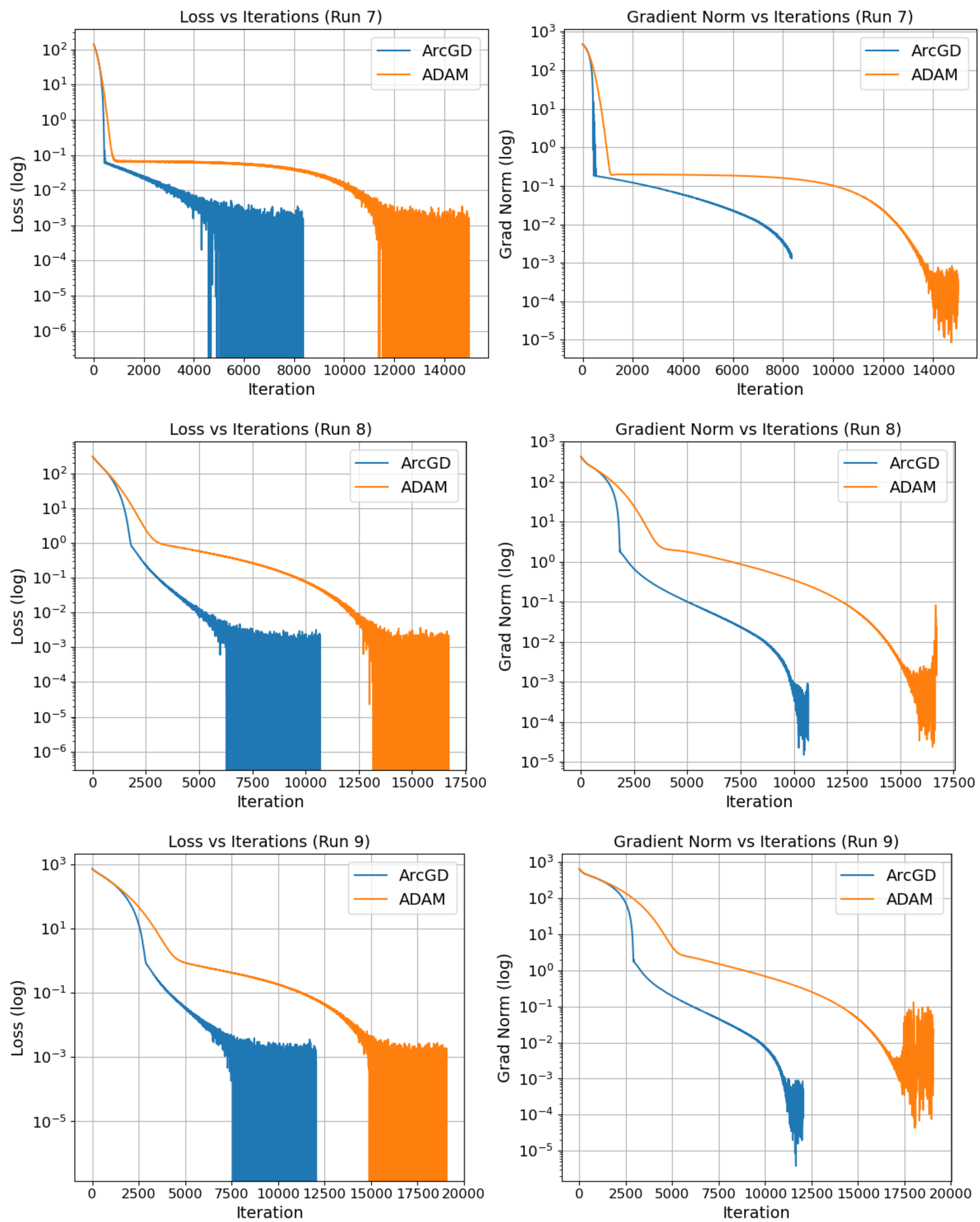

Figure B2.3: Loss and gradient norm for Run 7 to Run 9

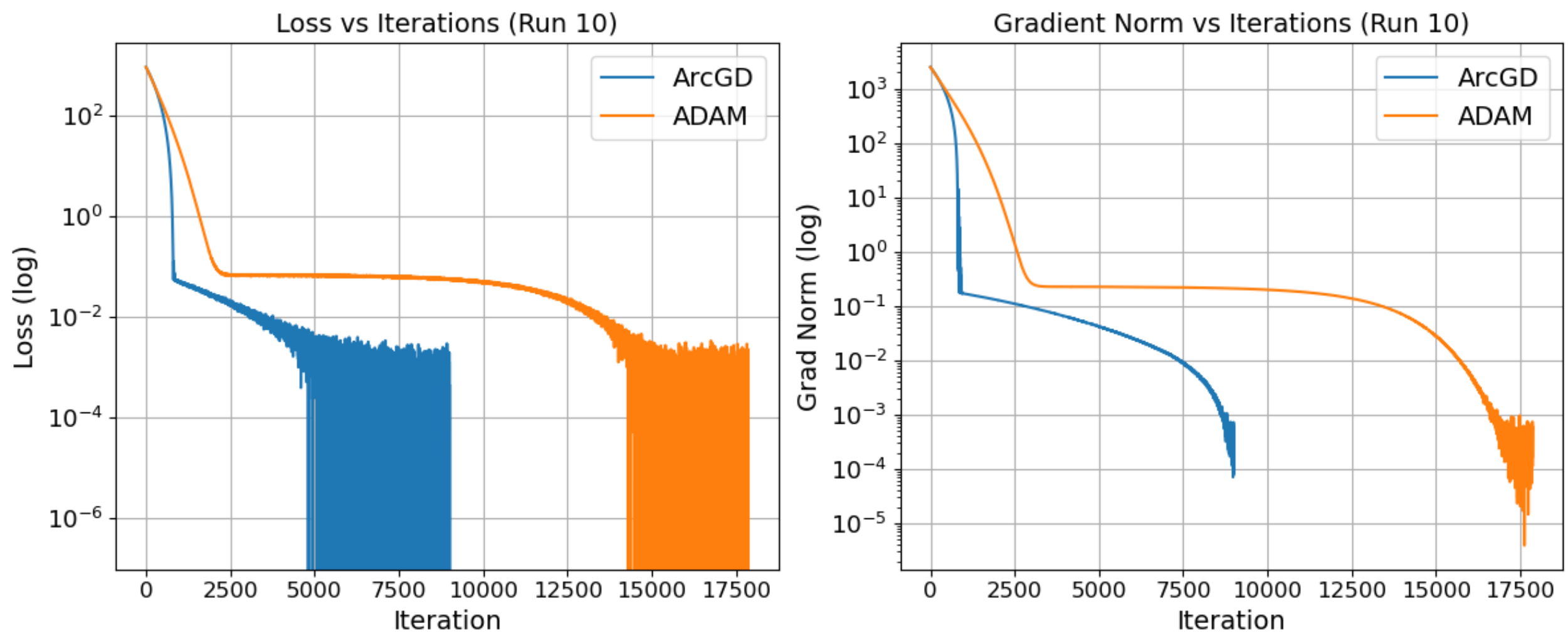

Figure B2.4: Loss and gradient norm for Run 10

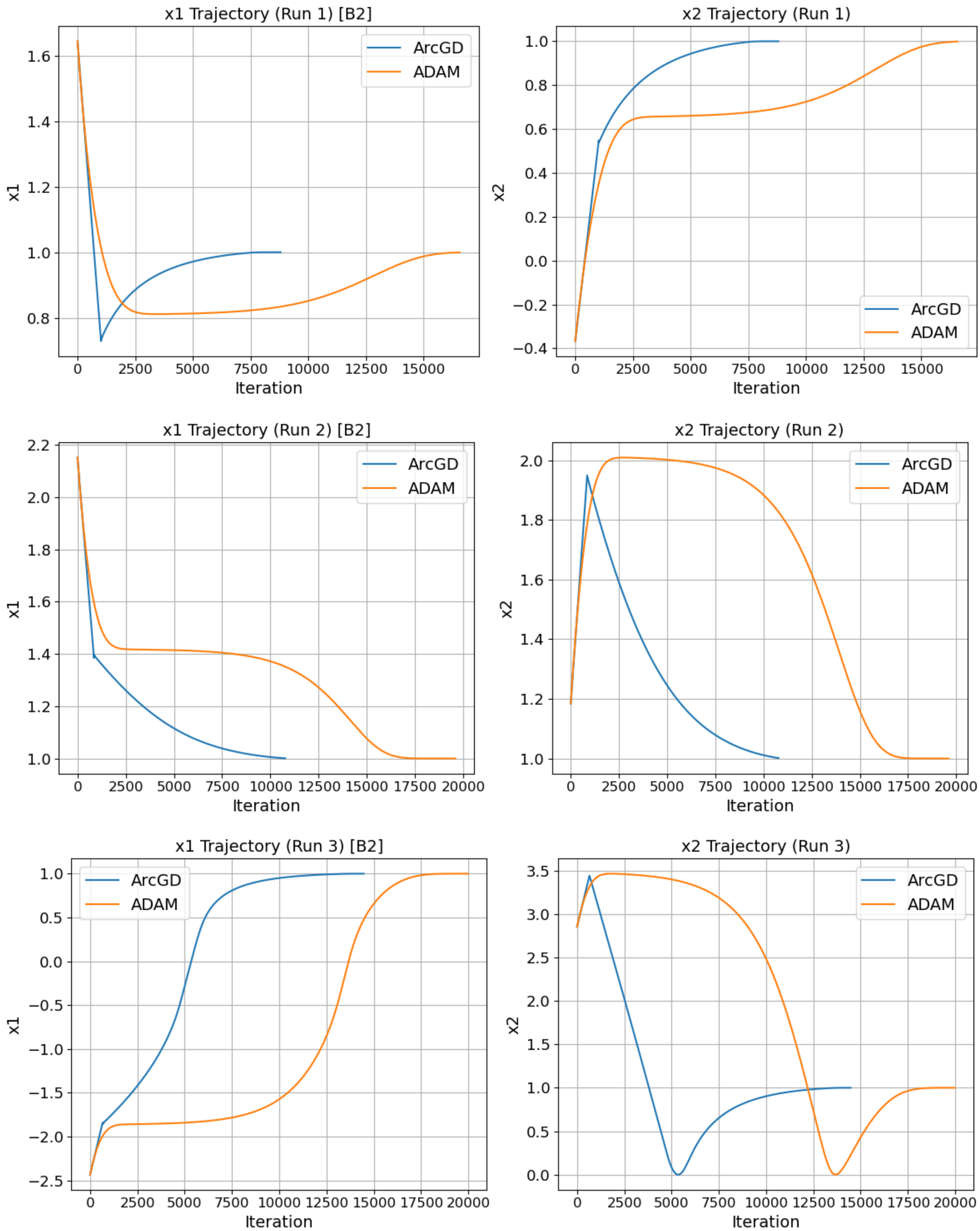

Figure B2.5: Change in $(x_1, x_2)$ for Run 1 to Run 3

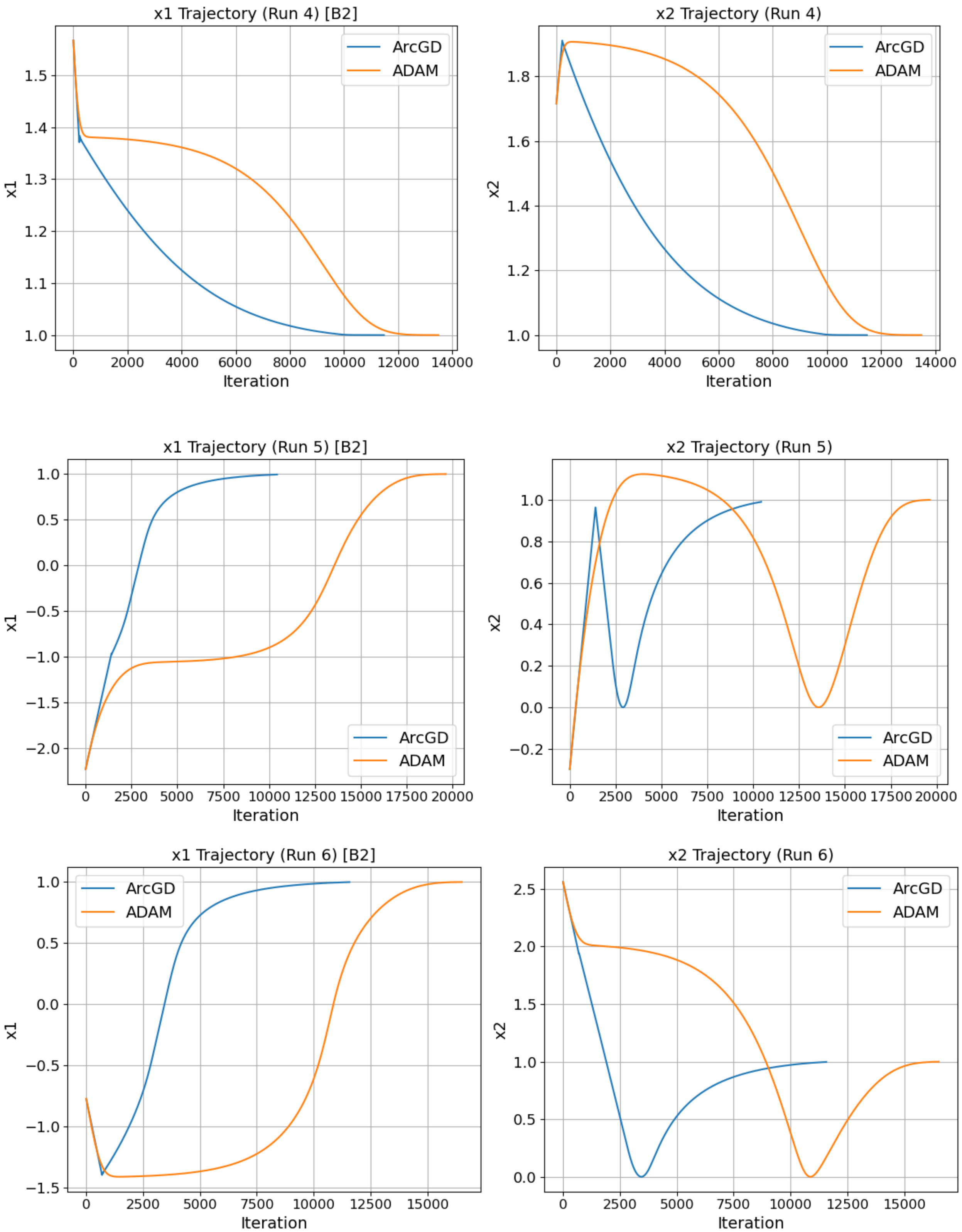

Figure B2.6: Change in $(x_1, x_2)$ for Run 4 to Run 6

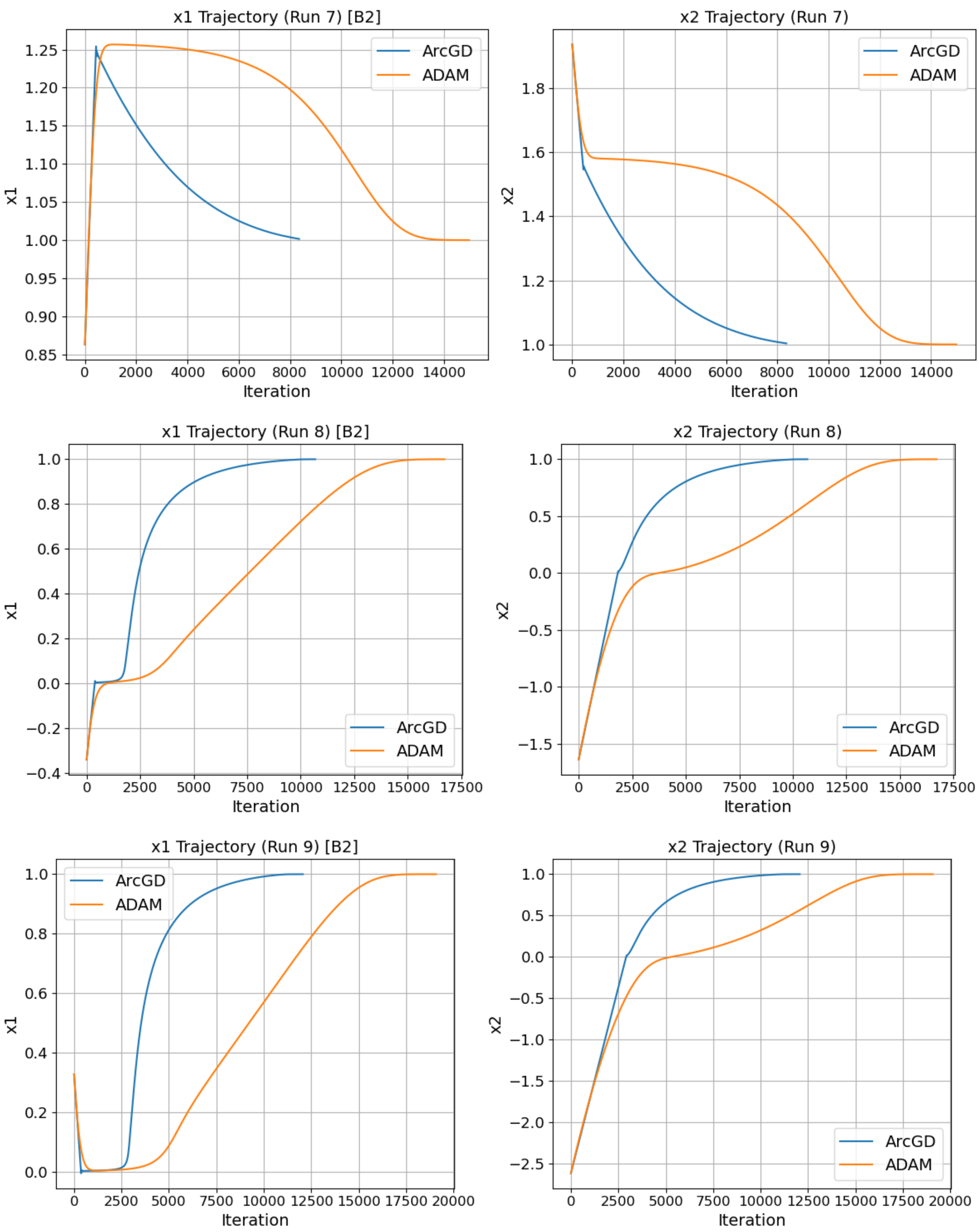

Figure B2.7: Change in $(x_1, x_2)$ for Run 7 to Run 9

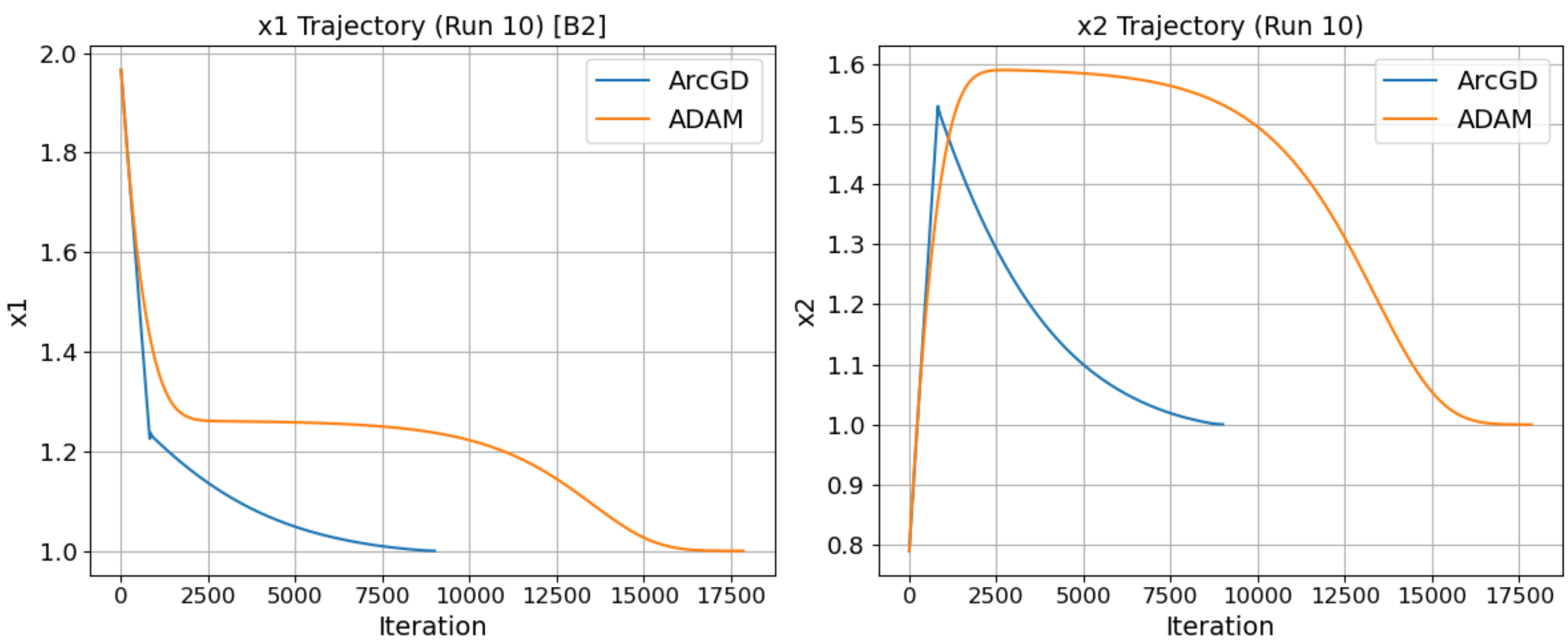

Figure B2.8 Change in $(x_1, x_2)$ for Run 10

## 8.7 Test: B10

Table B10: Detailed results of individual runs for the ten-dimensional test case (B10).

| Run 10D | Optimizer | EMA_Patience_Stopping & Low_Loss | Iterations | Final_Loss | Final_Gradient_Norm | Distance_to_Minima | Time(s) |
|---|---|---|---|---|---|---|---|
| 1 | ArcGD | TRUE | 11801 | 2.24E-04 | 1.01E-03 | 1.70E-05 | 0.50 |
| 1 | ADAM | TRUE | 19662 | -3.92E-04 | 1.17E-02 | 7.75E-05 | 0.84 |
| 2 | ArcGD | FALSE | 11974 | 3.99E+00 | 7.94E-04 | 6.30E-01 | 0.61 |
| 2 | ADAM | FALSE | 18977 | 3.99E+00 | 4.45E-03 | 6.30E-01 | 0.75 |
| 3 | ArcGD | TRUE | 12348 | 9.07E-04 | 5.14E-04 | 1.80E-04 | 0.63 |
| 3 | ADAM | TRUE | 21595 | 2.18E-03 | 1.20E-01 | 2.65E-05 | 0.91 |
| 4 | ArcGD | TRUE | 10878 | 2.02E-04 | 5.39E-04 | 8.03E-06 | 0.59 |
| 4 | ADAM | TRUE | 17044 | -7.87E-05 | 1.38E-02 | 1.09E-04 | 0.93 |
| 5 | ArcGD | FALSE | 11265 | 3.99E+00 | 1.03E-03 | 6.30E-01 | 0.88 |
| 5 | ADAM | FALSE | 17956 | 3.98E+00 | 3.66E-02 | 6.30E-01 | 0.88 |
| 6 | ArcGD | TRUE | 14655 | 2.79E-04 | 1.20E-03 | 1.85E-06 | 0.98 |
| 6 | ADAM | TRUE | 20810 | 1.52E-03 | 2.32E-02 | 1.56E-05 | 1.16 |
| 7 | ArcGD | TRUE | 11998 | -4.90E-04 | 1.14E-03 | 3.17E-06 | 0.69 |
| 7 | ADAM | TRUE | 21581 | 3.05E-03 | 5.69E-03 | 2.19E-06 | 1.13 |
| 8 | ArcGD | TRUE | 14474 | 4.72E-04 | 7.04E-04 | 2.35E-05 | 0.75 |
| 8 | ADAM | FALSE | 17826 | 3.99E+00 | 2.31E-03 | 6.30E-01 | 0.79 |
| 9 | ArcGD | TRUE | 11148 | 2.14E-04 | 4.11E-03 | 2.25E-03 | 0.56 |
| 9 | ADAM | TRUE | 20065 | -1.41E-03 | 7.87E-03 | 3.42E-06 | 0.94 |
| 10 | ArcGD | FALSE | 11551 | 3.99E+00 | 8.63E-04 | 6.30E-01 | 0.69 |
| 10 | ADAM | FALSE | 20011 | 3.99E+00 | 1.82E-02 | 6.30E-01 | 1.35 |

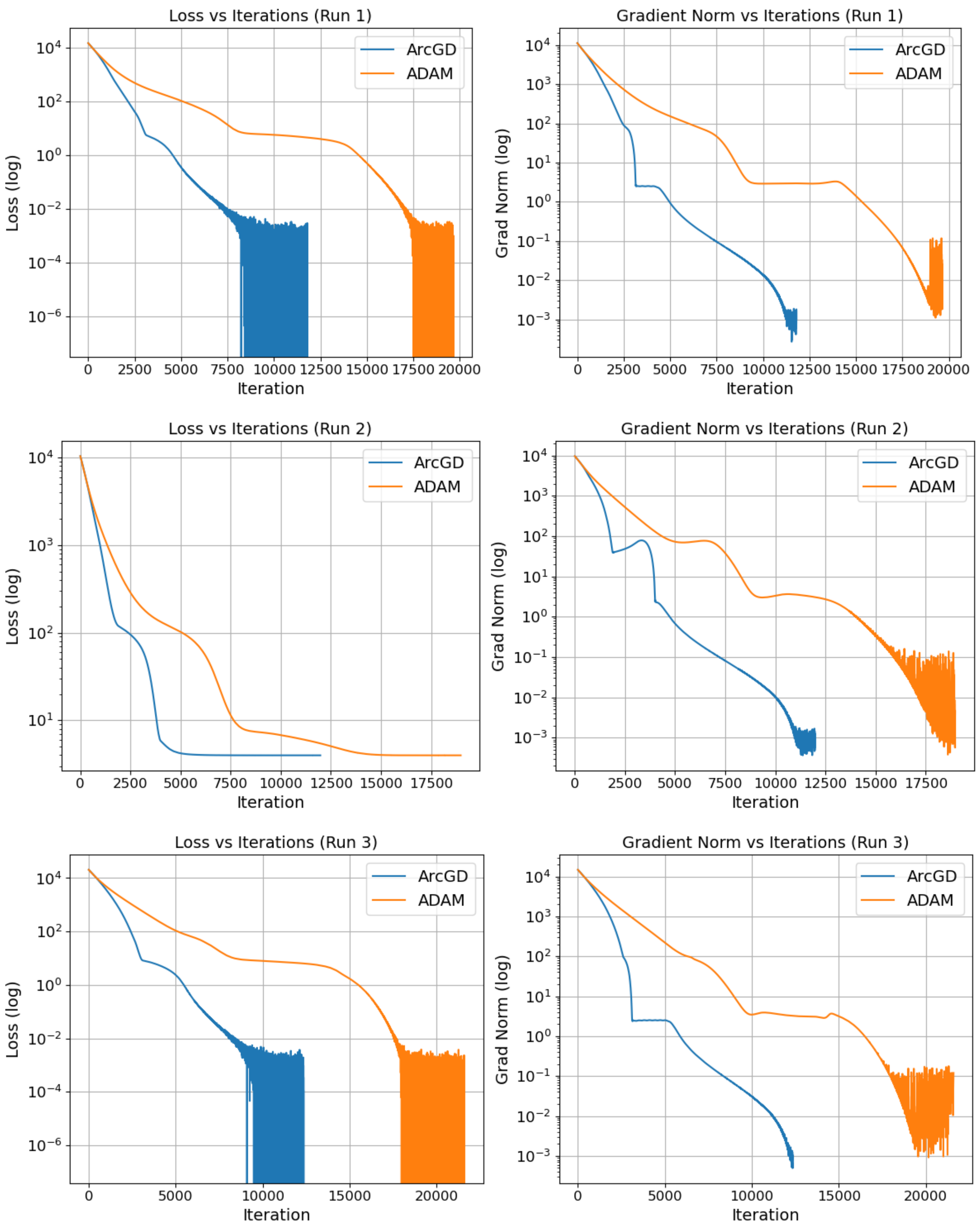

Figure B10.1: Loss and gradient norm for Run 1 to Run 3

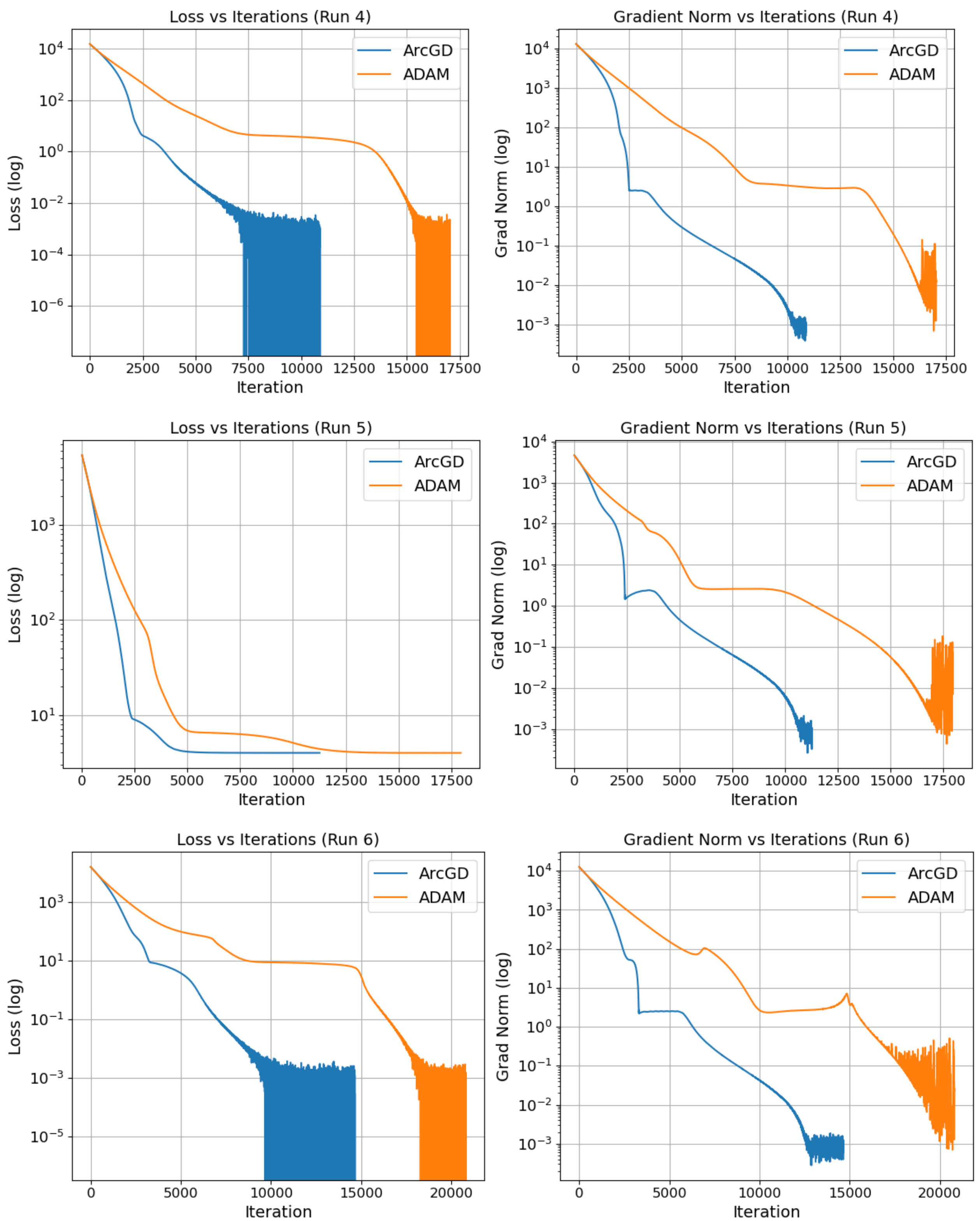

Figure B10.2: Loss and gradient norm for Run 4 to Run 6

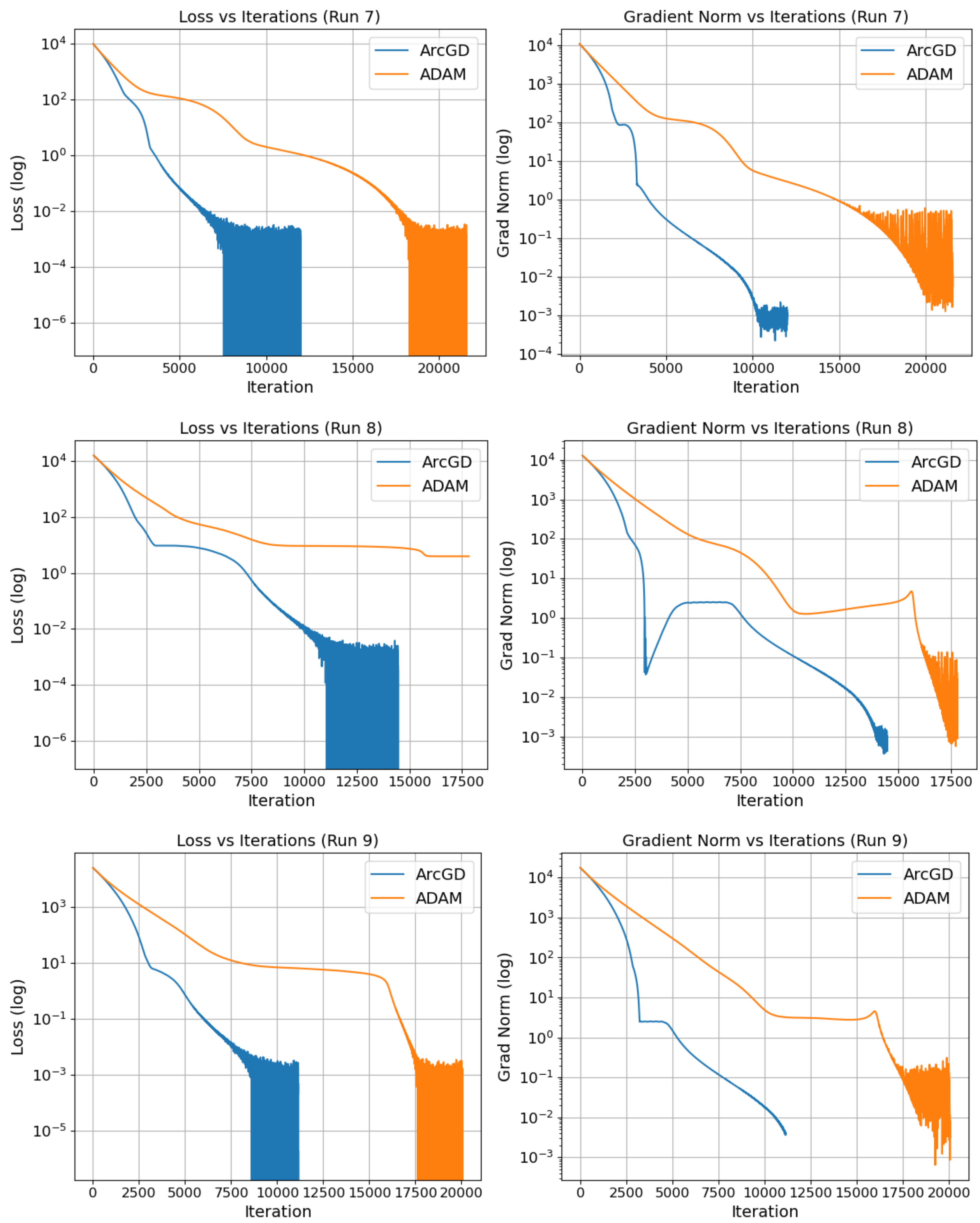

Figure B10.3: Loss and gradient norm for Run 7 to Run 9

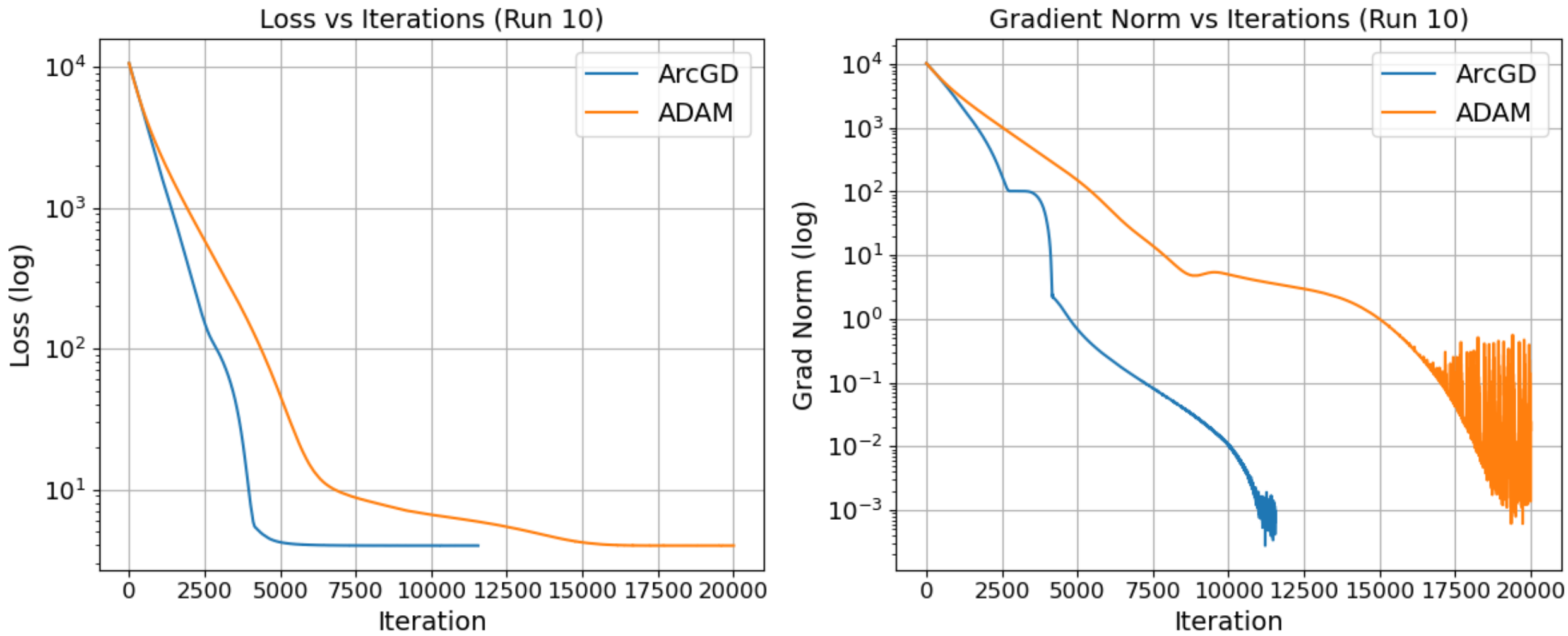

Figure B10.4: Loss and gradient norm for Run 10

## 8.8 Test: B100

Table B100: Detailed results of individual runs for the hundred-dimensional test case (B100).

| Run 100D | Optimizer | EMA_Patience_Stopping & Low_Loss | Iterations | Final_Loss | Final_Gradient_Norm | Distance_to_Minima | Time(s) |
|---|---|---|---|---|---|---|---|
| 1 | ArcGD | TRUE | 40169 | 7.25E-04 | 3.62E-03 | 1.11E-06 | 2.08 |
| 1 | ADAM | TRUE | 24418 | -2.36E-04 | 3.54E-01 | 2.11E-05 | 1 |
| 2 | ArcGD | TRUE | 19292 | 8.64E-04 | 2.88E-03 | 7.07E-07 | 0.95 |
| 2 | ADAM | TRUE | 20945 | -4.79E-04 | 1.43E-01 | 9.28E-06 | 1.02 |
| 3 | ArcGD | TRUE | 23163 | 8.29E-04 | 3.64E-03 | 4.12E-06 | 1.42 |
| 3 | ADAM | TRUE | 22481 | 1.55E-03 | 5.41E-01 | 3.17E-05 | 0.92 |
| 4 | ArcGD | TRUE | 22092 | -4.50E-04 | 3.75E-03 | 4.06E-07 | 1.53 |
| 4 | ADAM | TRUE | 22641 | 1.47E-03 | 1.60E-01 | 1.11E-05 | 1.31 |
| 5 | ArcGD | TRUE | 40218 | -1.25E-03 | 3.26E-03 | 5.33E-05 | 2.2 |
| 5 | ADAM | TRUE | 25158 | -9.64E-04 | 6.59E-02 | 4.50E-06 | 1.22 |
| 6 | ArcGD | TRUE | 19095 | 1.70E-04 | 3.31E-03 | 1.51E-06 | 0.98 |
| 6 | ADAM | TRUE | 22634 | -2.49E-04 | 5.53E-02 | 3.91E-06 | 1.08 |
| 7 | ArcGD | FALSE | 38099 | 3.99E+00 | 2.70E-03 | 1.99E-01 | 2.1 |
| 7 | ADAM | FALSE | 23952 | 3.99E+00 | 2.98E-01 | 1.99E-01 | 1.1 |
| 8 | ArcGD | TRUE | 19192 | -2.77E-04 | 3.44E-03 | 3.39E-07 | 0.94 |
| 8 | ADAM | TRUE | 22836 | 2.62E-04 | 4.61E-01 | 2.68E-05 | 1.03 |
| 9 | ArcGD | TRUE | 28709 | 6.69E-04 | 2.91E-03 | 3.79E-06 | 1.98 |
| 9 | ADAM | TRUE | 20152 | -7.07E-04 | 1.23E-01 | 9.03E-06 | 1.07 |
| 10 | ArcGD | TRUE | 39766 | 1.96E-03 | 2.93E-03 | 2.97E-05 | 2.31 |
| 10 | ADAM | TRUE | 25684 | -4.73E-04 | 1.49E-01 | 8.98E-06 | 1.38 |

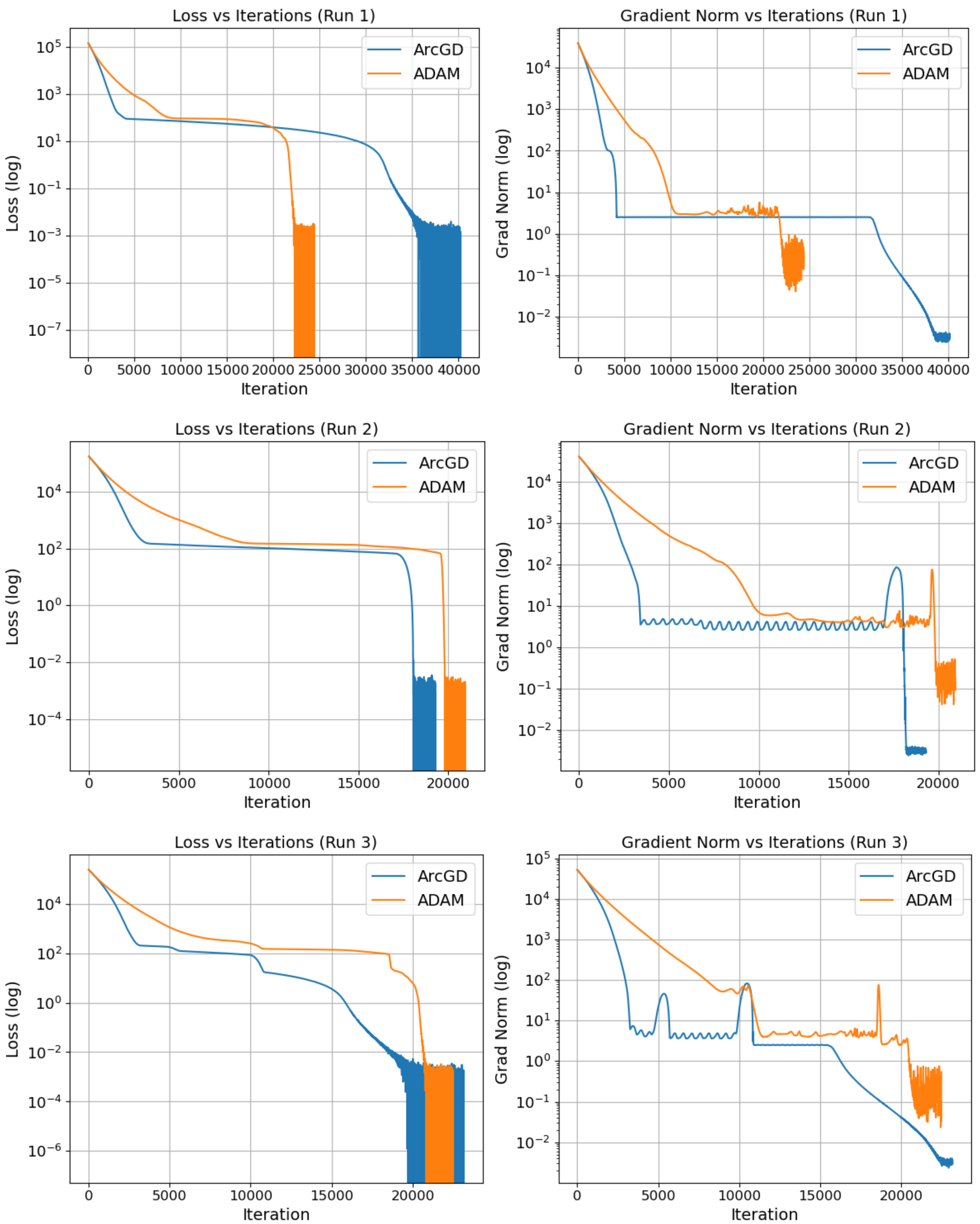

Figure B100.1: Loss and gradient norm for Run 1 to Run 3

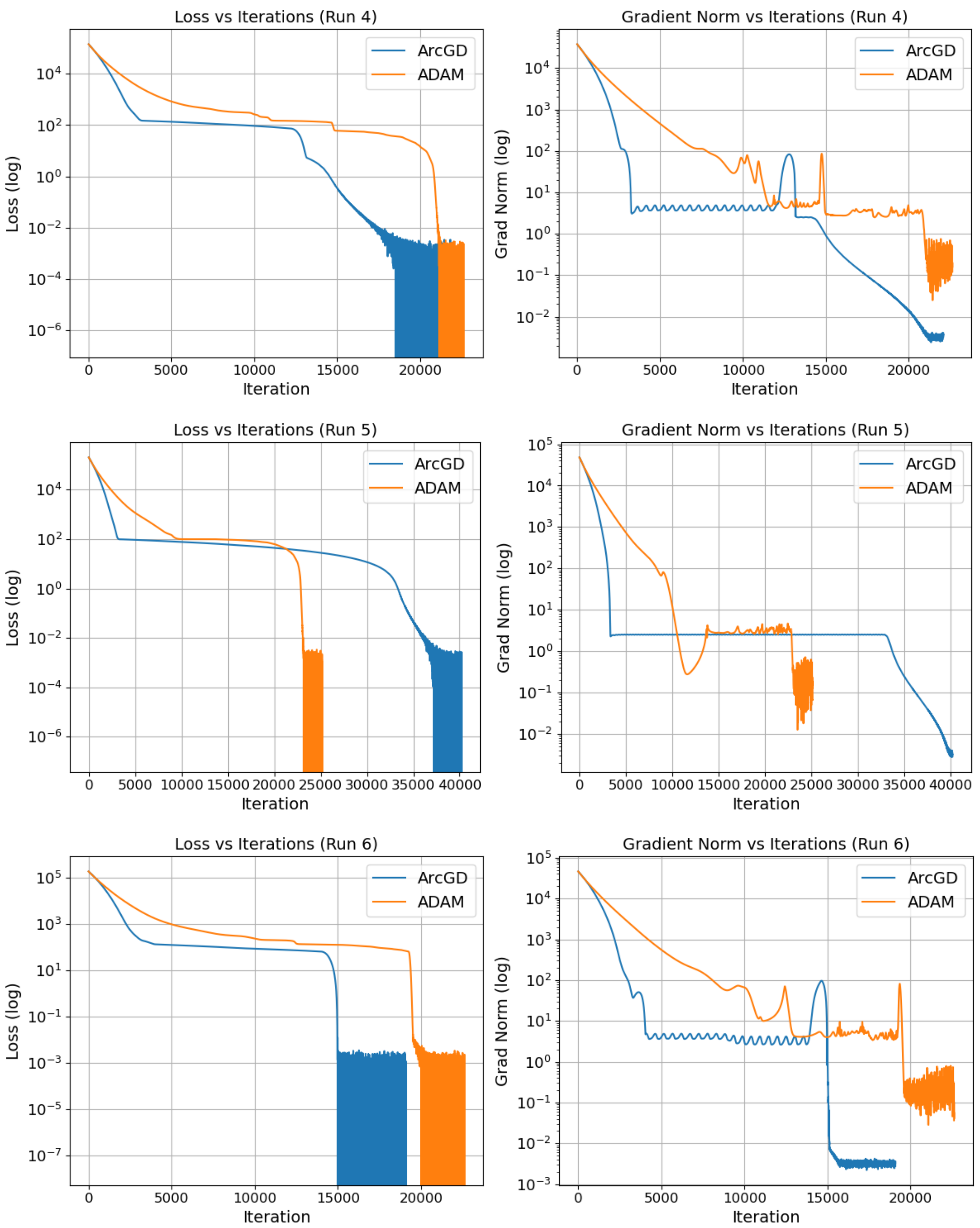

Figure B100.2: Loss and gradient norm for Run 4 to Run 6

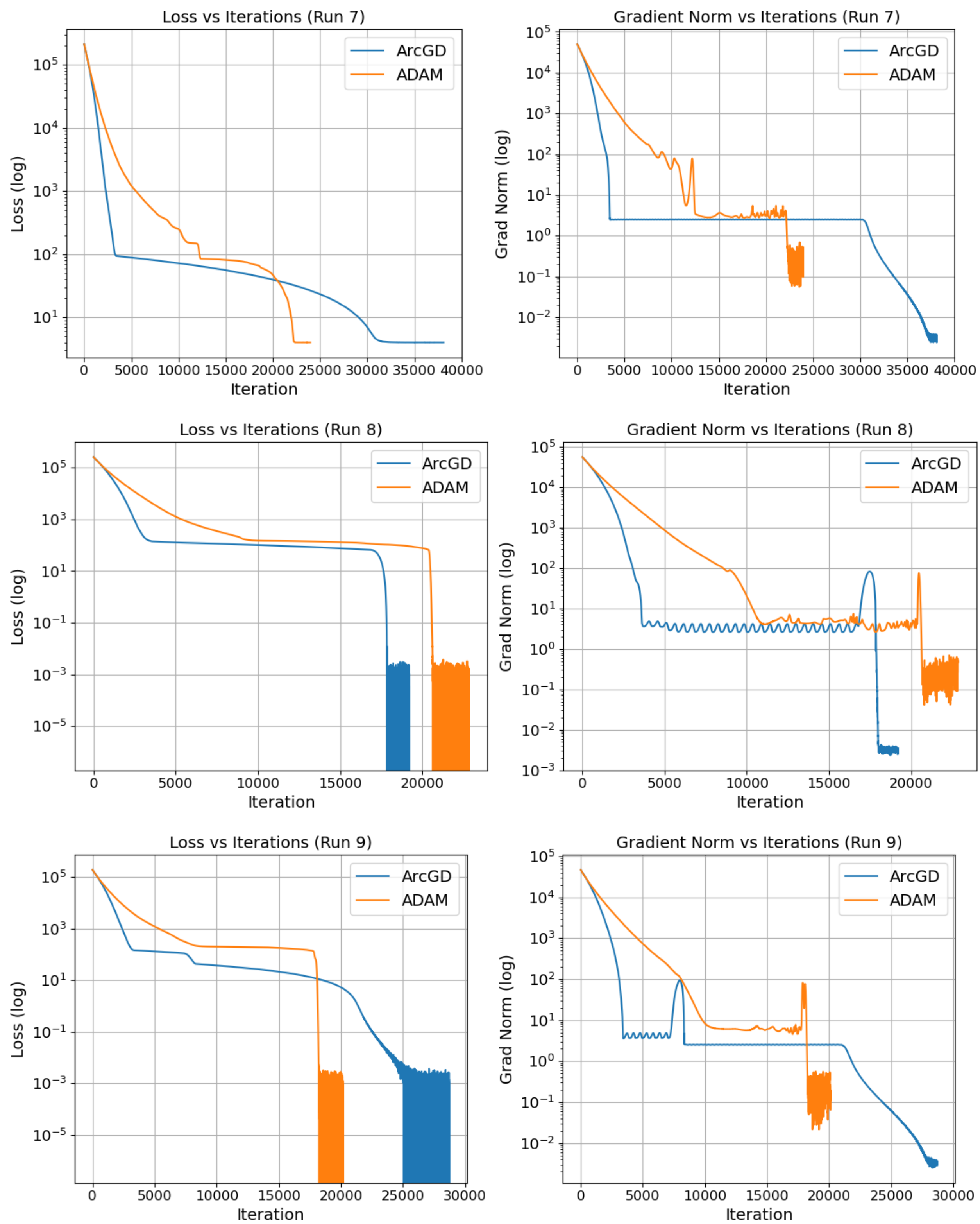

Figure B100.3: Loss and gradient norm for Run 7 to Run 9

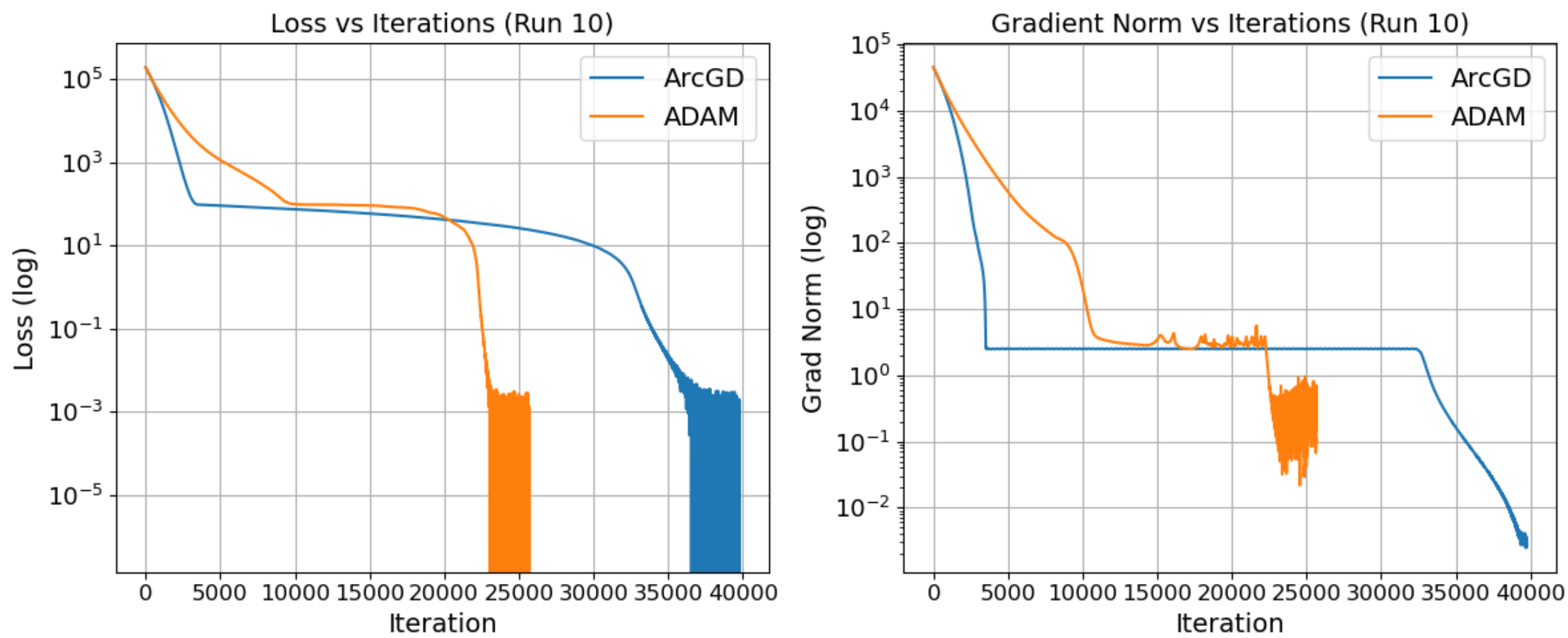

Figure B100.4: Loss and gradient norm for Run 10

## 8.9 Test: B1000

Table B1000: Detailed results of individual runs for the thousand-dimensional test case (B1000).

| Run 1000D | Optimizer | EMA_Patience_Stopping & Low_Loss | Iterations | Final_Loss | Final_Gradient_Norm | Distance_to_Minima | Time(s) |
|---|---|---|---|---|---|---|---|
| 1 | ArcGD | TRUE | 58923 | 1.07E-03 | 9.86E-03 | 7.09E-06 | 5.86 |
| 1 | ADAM | TRUE | 26056 | 1.26E-03 | 1.48E+00 | 2.69E-05 | 2.48 |
| 2 | ArcGD | TRUE | 76383 | 5.08E-04 | 1.05E-02 | 4.91E-06 | 7.24 |
| 2 | ADAM | TRUE | 29328 | 1.61E-03 | 9.74E-01 | 1.74E-05 | 2.47 |
| 3 | ArcGD | TRUE | 68790 | 3.11E-04 | 1.00E-02 | 3.52E-07 | 6.5 |
| 3 | ADAM | TRUE | 26482 | 4.29E-04 | 1.36E+00 | 2.46E-05 | 1.88 |
| 4 | ArcGD | TRUE | 62349 | -5.85E-04 | 1.03E-02 | 6.74E-07 | 5.64 |
| 4 | ADAM | TRUE | 26739 | 1.25E-03 | 1.35E+00 | 2.44E-05 | 1.91 |
| 5 | ArcGD | TRUE | 66251 | 1.04E-03 | 1.01E-02 | 5.30E-05 | 6.3 |
| 5 | ADAM | TRUE | 28228 | -1.39E-03 | 1.27E+00 | 2.24E-05 | 2.04 |
| 6 | ArcGD | TRUE | 74374 | 5.83E-04 | 9.75E-03 | 3.93E-07 | 6.57 |
| 6 | ADAM | FALSE | 27047 | 3.98E+00 | 1.38E+00 | 6.30E-02 | 2.06 |
| 7 | ArcGD | TRUE | 65331 | 4.74E-04 | 9.98E-03 | 3.73E-07 | 6.45 |
| 7 | ADAM | TRUE | 27295 | -7.04E-05 | 1.30E+00 | 2.33E-05 | 2.08 |
| 8 | ArcGD | TRUE | 130182 | -2.80E-04 | 1.03E-02 | 7.41E-07 | 14.32 |
| 8 | ADAM | TRUE | 29204 | -1.27E-04 | 1.33E+00 | 2.36E-05 | 2.44 |
| 9 | ArcGD | TRUE | 66888 | 7.61E-04 | 1.01E-02 | 5.13E-07 | 6.46 |
| 9 | ADAM | TRUE | 28074 | 4.14E-04 | 1.22E+00 | 2.18E-05 | 2.1 |
| 10 | ArcGD | TRUE | 82192 | 9.28E-04 | 9.79E-03 | 3.52E-07 | 7.35 |
| 10 | ADAM | TRUE | 33208 | -1.57E-03 | 1.46E+00 | 2.58E-05 | 2.58 |

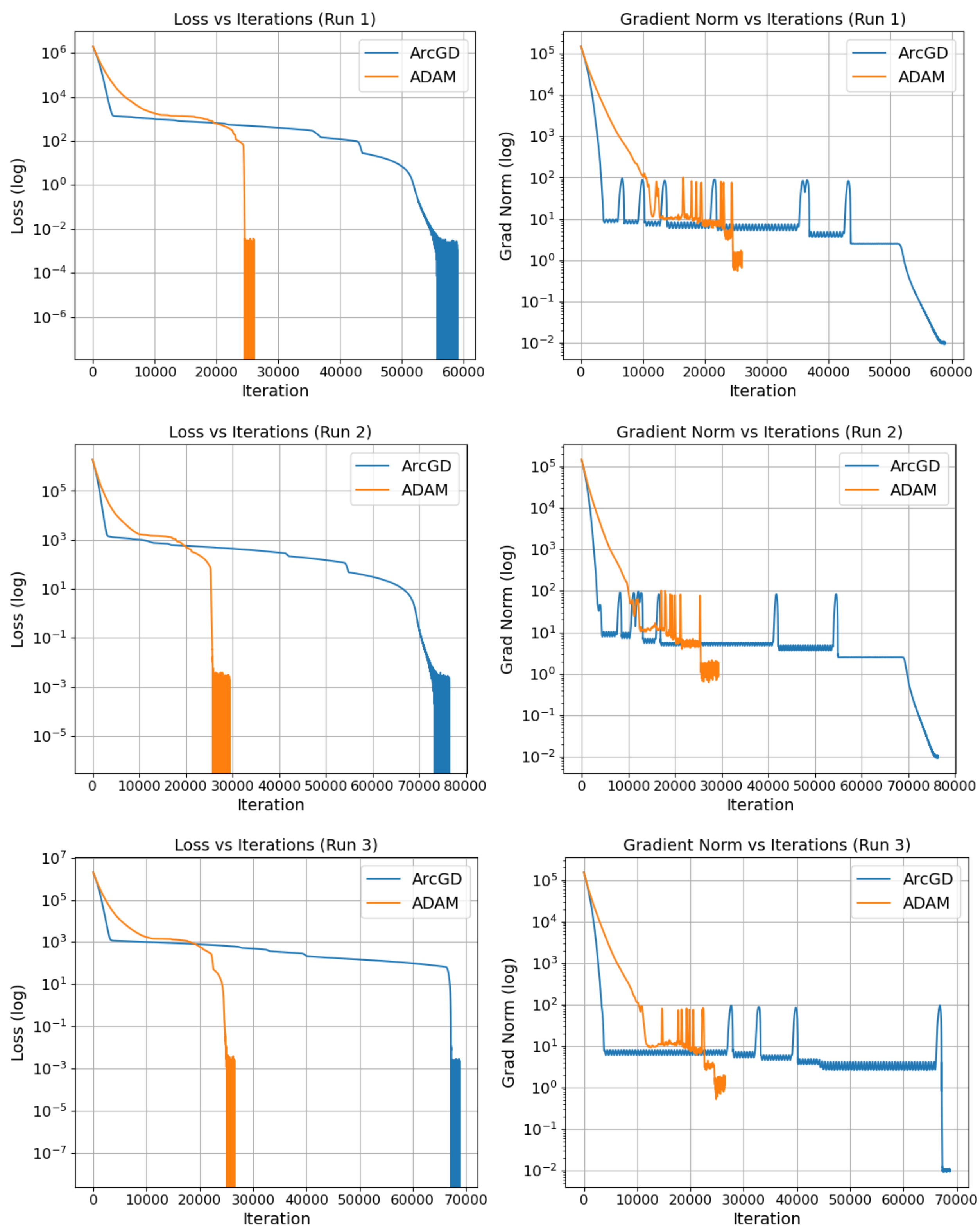

Figure B1000.1: Loss and gradient norm for Run 1 to Run 3

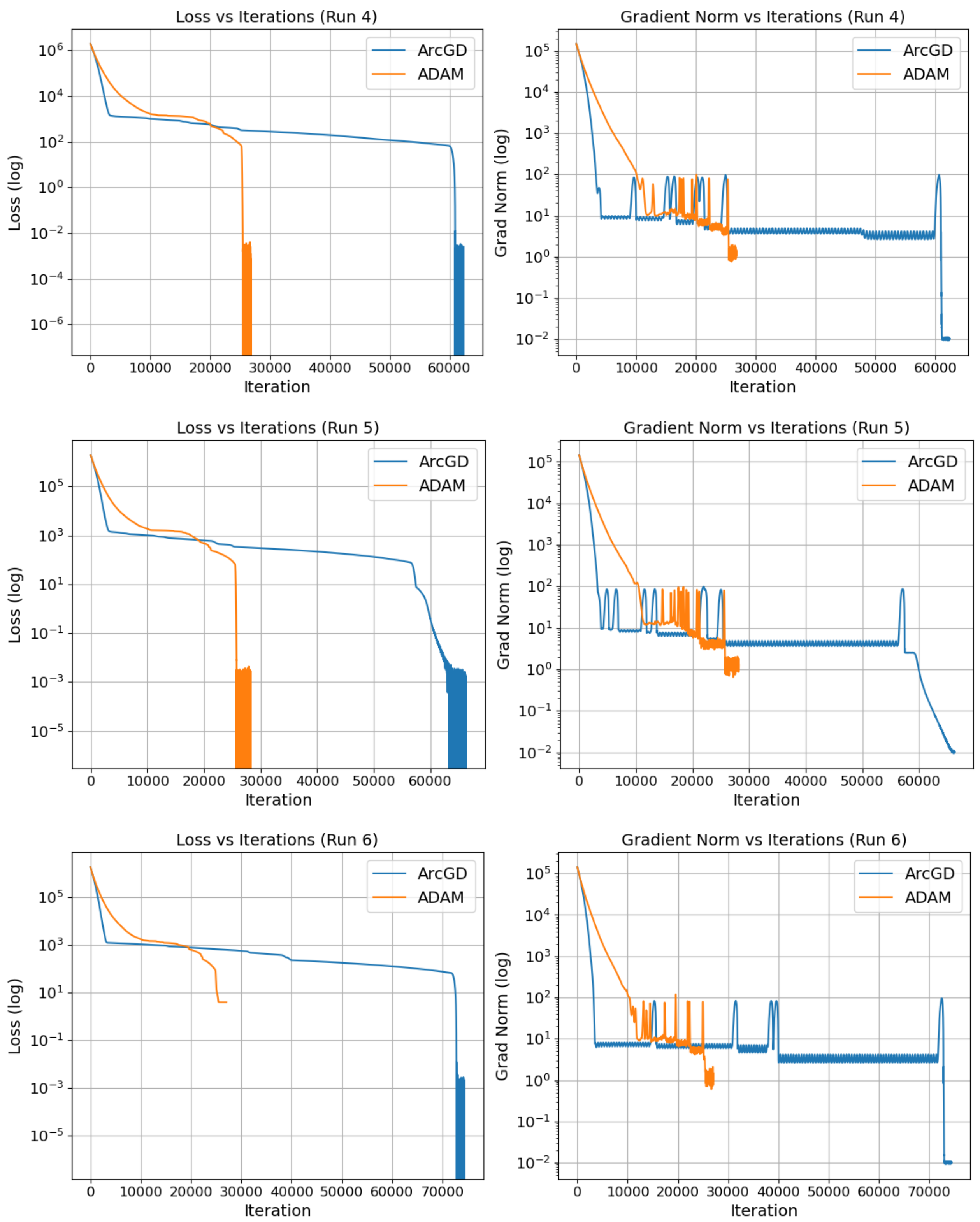

Figure B1000.2: Loss and gradient norm for Run 4 to Run 6

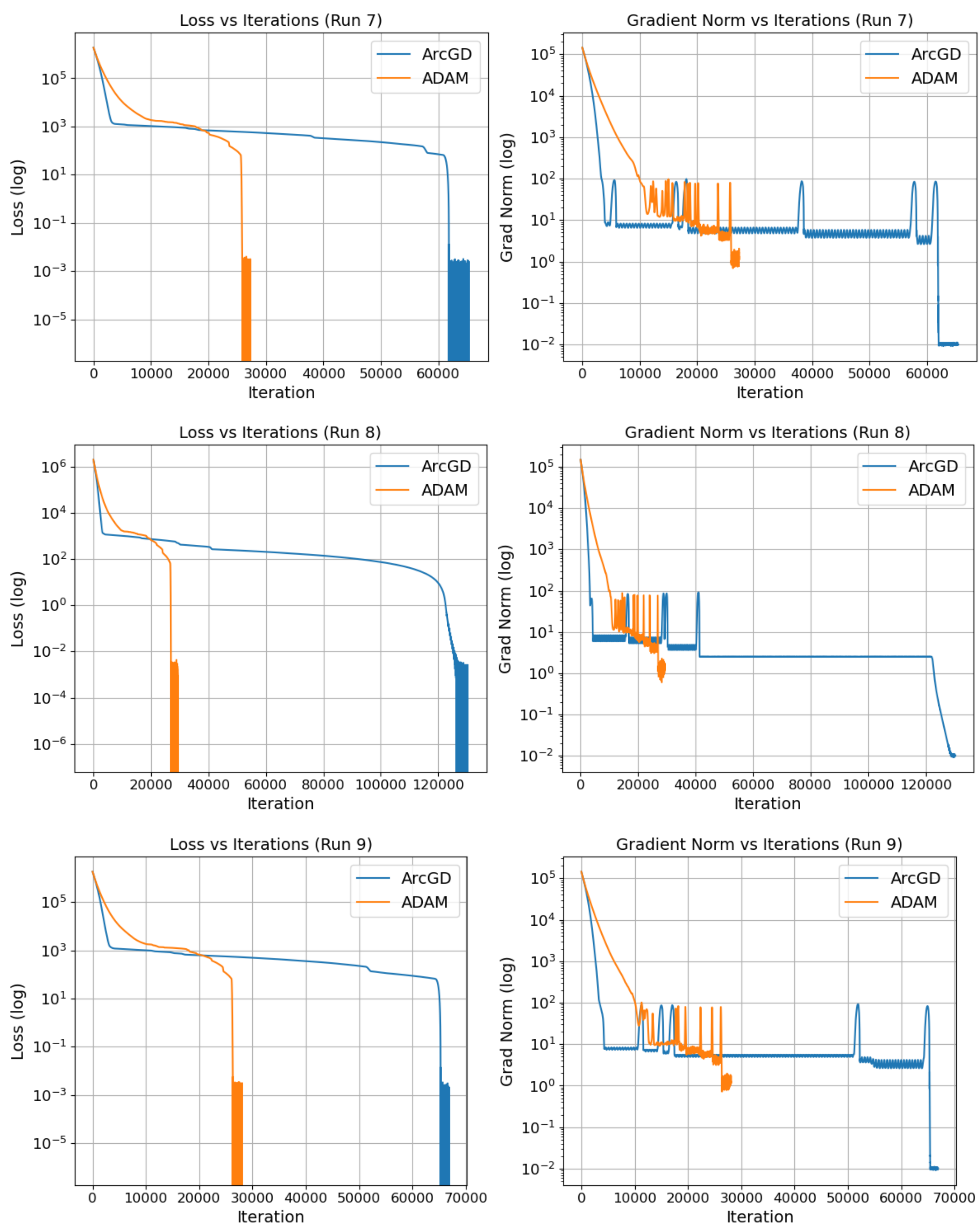

Figure B1000.3: Loss and gradient norm for Run 7 to Run 9

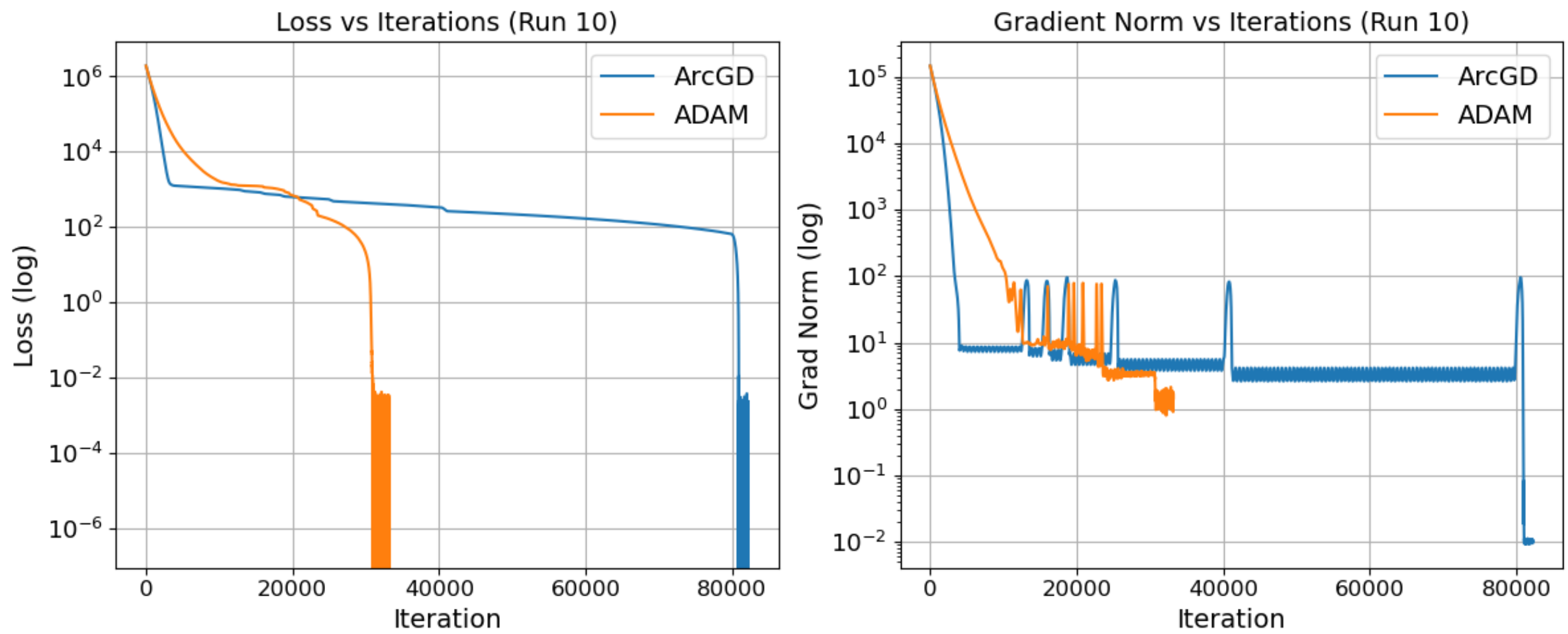

Figure B1000.4: Loss and gradient norm for Run 10

## 8.10 Test: B50000

Table B50000: Detailed results of individual runs for the fifty thousand-dimensional test case (B50000).

| Run 50000D | Optimizer | EMA_Patience_Stopping & Low_Loss | Iterations | Final_Loss | Final_Gradient_Norm | Distance_to_Minima | Time(s) |
|---|---|---|---|---|---|---|---|
| 1 | ArcGD | TRUE | 181713 | -7.12E-04 | 7.13E-02 | 3.52E-07 | 528.43 |
| 1 | ADAM | TRUE | 37224 | 3.07E-02 | 1.06E+01 | 2.66E-05 | 93.11 |
| 2 | ArcGD | TRUE | 210511 | 1.08E-03 | 7.16E-02 | 3.40E-07 | 597.31 |
| 2 | ADAM | TRUE | 34625 | 3.29E-02 | 1.06E+01 | 2.64E-05 | 86.96 |
| 3 | ArcGD | TRUE | 242144 | 1.32E-04 | 7.25E-02 | 3.38E-07 | 682.3 |
| 3 | ADAM | TRUE | 38343 | 3.29E-02 | 1.08E+01 | 2.68E-05 | 97.53 |

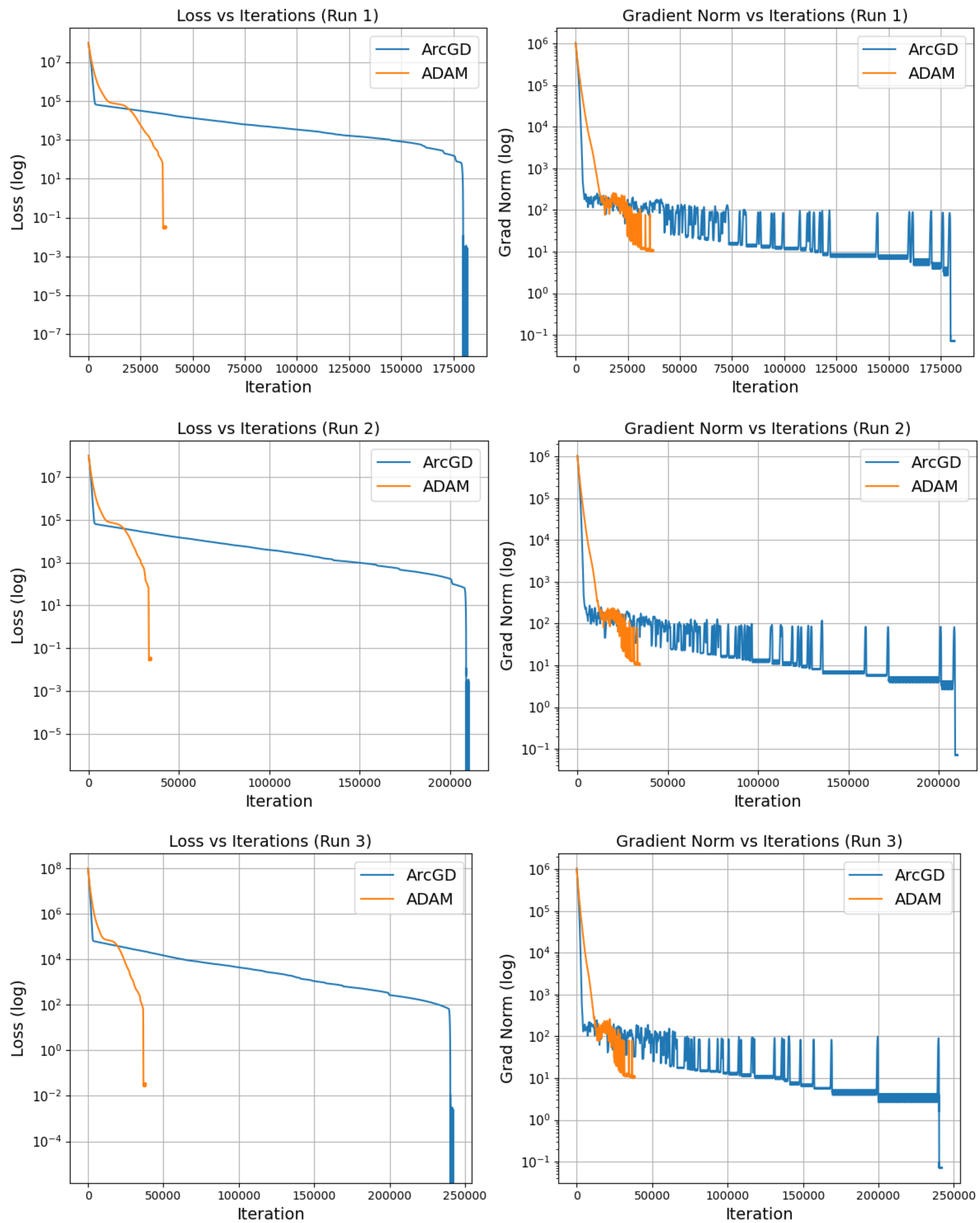

Figure B50000.1: Loss and gradient norm for Run 1 to Run 3

## Appendix D ($eta_{low}$ effect in CIFAR10)

An ablation study was conducted to evaluate the sensitivity of ArcGD's adaptive sign term coefficient ($eta_{low}$) by comparing the default value (0.01) against a 10× increase (0.1) across all 8 MLP architectures on CIFAR-10 over 20,000 iterations. The results demonstrate that $eta_{low} = 0.01$ is overall better, achieving 50.81% average test accuracy compared to 49.93% for $eta_{low} = 0.1$ ($-0.88\%$ degradation). While both variants show similar training dynamics with $eta_{low} = 0.1$ exhibiting faster loss convergence early in training, such aggressive reduction in the adaptive sign term leads to consistent performance degradation across all architectures, with the largest impact on medium-width networks (medium: $-2.07\%$, const_medium: $-1.06\%$) and minimal effect on the deepest architecture (const_deep: $-0.24\%$). The training curves reveal that higher $eta_{low}$ values produce smoother but inferior convergence, suggesting that the default $eta_{low} = 0.01$ provides an optimal balance between the $T_i$ and the adaptive sign correction, preventing premature loss of the fine-grained directional adjustments that contribute to ArcGD's superior generalization.

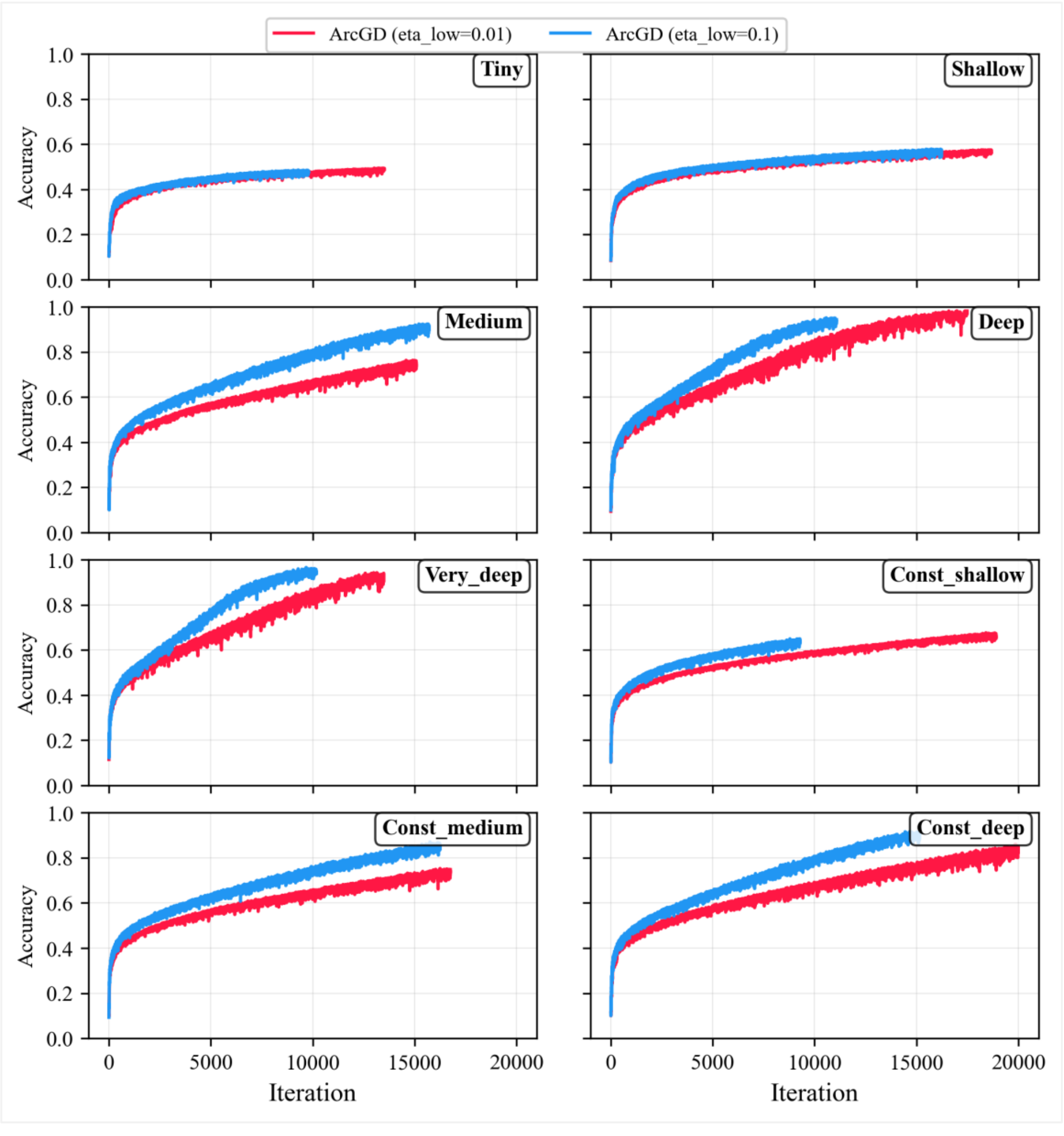

Figure D1. Test accuracy between $eta_{low} = 0.01$ (default) and $eta_{low} = 0.1$.

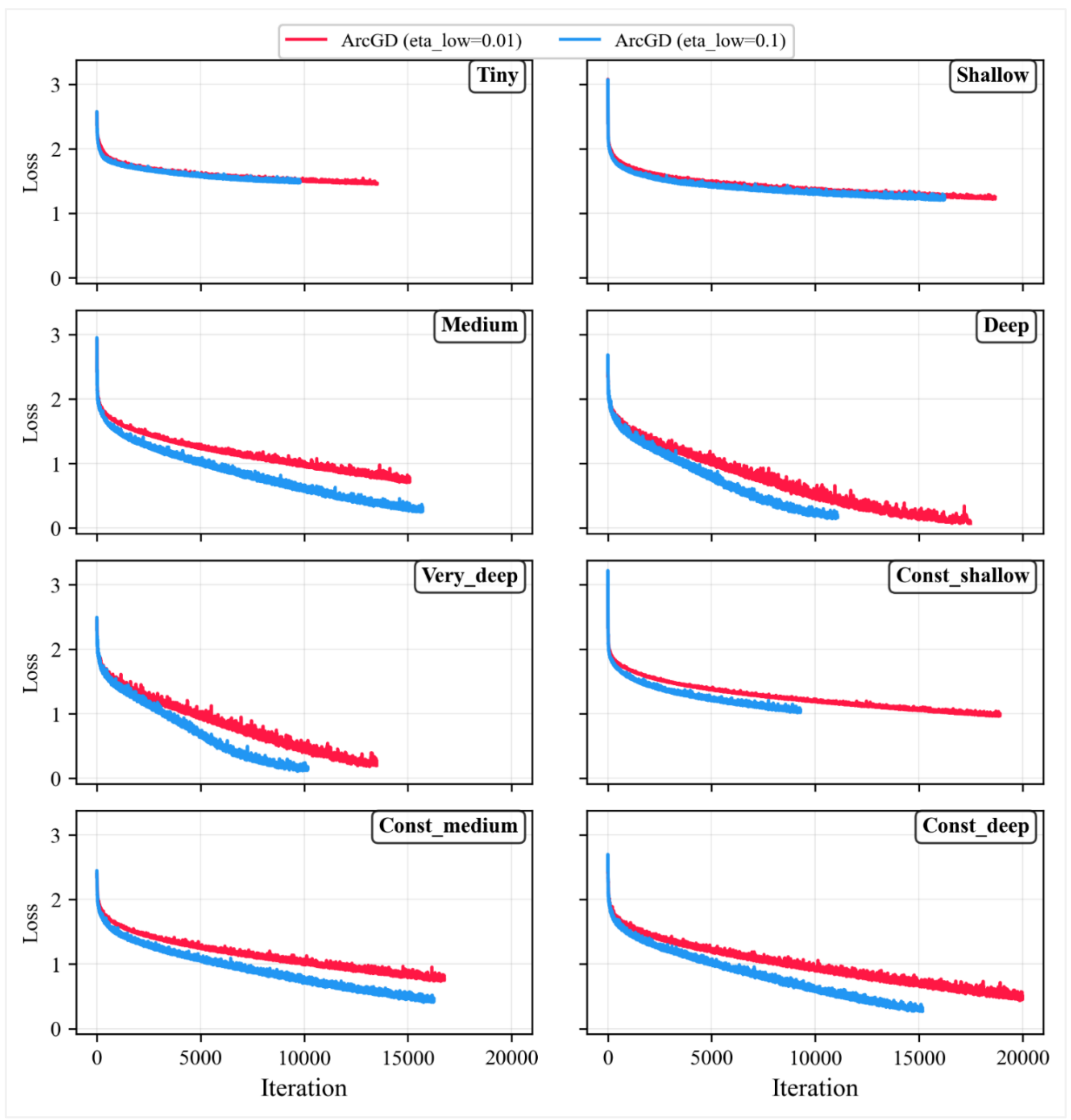

Figure D2. Training loss between $eta_{low} = 0.01$ (default) and $eta_{low} = 0.1$.

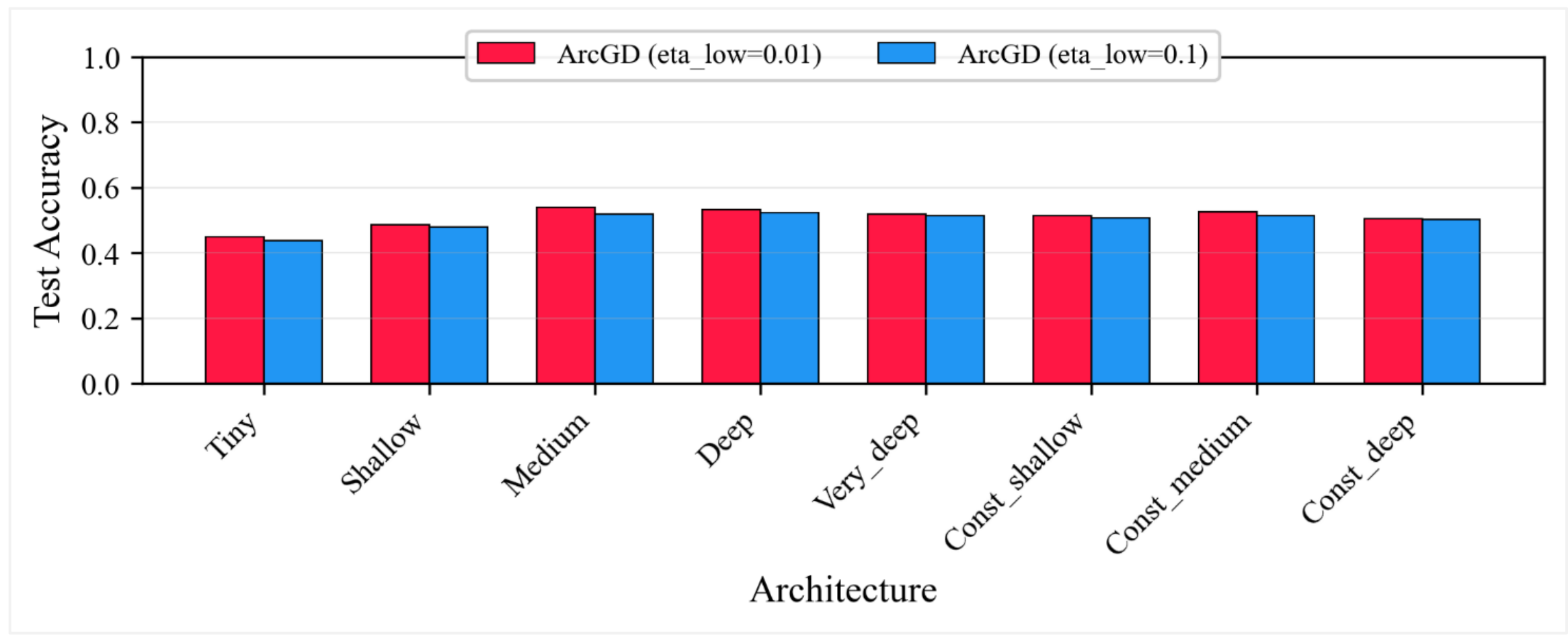

Figure D3. Test accuracy between $eta_{low} = 0.01$ (default) and $eta_{low} = 0.1$.

# Appendix E (Extended Discussion and Additional Insights)

## Geometric Interpretation of ArcGD

In its core form, ArcGD applies the coordinate-wise update

$$\Delta x \;=\; -\,\alpha\,\frac{g_x}{\sqrt{1+g_x^2}} \qquad (F1)$$

where $g_x = \partial f/\,\partial x$ is the partial derivative along coordinate $x$. Consider the 2D coordinate slice given by the graph of the loss restricted to the coordinate, $\gamma(x) = (x, f(x))$. The (non-unit) normal to that curve is $(-g_x,\, 1)$, and the corresponding unit normal is

$$\mathbf{n}(x) = \frac{(-g_x,\, 1)}{\sqrt{1+g_x^2}} \qquad (F2)$$

The $x$-component of the unit normal equals

$$\mathbf{n}(x)\cdot \mathrm{e}_x \;=\; -\frac{g_x}{\sqrt{1+g_x^2}} \qquad (F3)$$

So, **the ArcGD core step can be interpreted as scaling the projection of the unit normal onto the parameter axis**. Equivalently, letting $\theta_x$ denote the inclination angle of the tangent to the curve, with $\tan(\theta_x) = g_x$, we have

$$\sin(\theta_x) = \frac{g_x}{\sqrt{1+g_x^2}} \qquad (F4)$$

and therefore

$$\Delta x = -\alpha\sin(\theta_x) \qquad (F5)$$

**Such interpretation makes the key distinction explicit: standard gradient descent scales step magnitude with slope magnitude $|g_x|$, whereas ArcGD scales the step with the orientation of the coordinate-slice curve (encoded by $\theta_x$), yielding a smooth, bounded map $\sin(\theta_x) \in (-1, 1)$.** In particular, as $|g_x| \to \infty$, $\sin(\theta_x) \to \mathrm{sign}(g_x)$, so the update remains bounded by $|\Delta x| \le \alpha$ even under arbitrarily steep slopes.

## Quadratic Bézier (Bernstein) Interpretation of the Update Profile

The ArcGD's update rule as function of $T$ with constant $c$ is reproduced here

$$\Delta x = -a.T - b.T(1-|T|) - c.sign(T)(1-|T|)$$

For $T \in [0,1)$ and $sign(T) = 1$, the update rule can be simplified as

$$\Delta x = c + (a+b-c)T - bT^2 \qquad (F6)$$

A Quadratic Bézier (Bernstein) scalar curve [Wyss, 2021] for $t \in [0,1]$ is

$$B(t) = (1-t)^2P_0 + 2(1-t)tP_1 + t^2P_2 \qquad (F7)$$

Expanding,

$$B(t) = P_0 + 2(P_1 - P_0)t + (P_0 - 2P_1 + P_2)t^2 \qquad (F8)$$

On comparing Equation (F6) and Equation (F8), following can be noted:

- Constant term: $P_0 = c$
- Linear term: $2(P_1 - P_0) = a + b - c \Rightarrow P_1 = c + \frac{a+b-c}{2} = \frac{a+b+c}{2}$
- Quadratic term: $P_0 - 2P_1 + P_2 = -b \Rightarrow P_2 = a$

Thus,

$$-\Delta x = (1 - T)^2 c + 2(1 - T)T\left(\frac{a + b + c}{2}\right) + T^2 a \qquad (F9)$$

For $T < 0$, the same algebra applies with $|T| = -T$ and $\text{sign}(T) = -1$, yielding a symmetric branch in $|T|$ with the appropriate sign. Thus, Equation (F9) expressing $-\Delta x$ happens to be a quadratic Bézier curve whose control values $P_0, P_1$, and $P_2$ are given explicitly in terms of $a$, $b$, and $c$; equivalently, the ArcGD update is simply the negative of that Bézier profile. As a result, the hyperparameters admit a direct geometric "curve-shaping" interpretation, providing a compact and interpretable way to regulate the update profile across different gradient regimes, while making clear that above describes the update-response curve (designed in $T$-space), not the parameter trajectory in the original space.

Moreover, a cubic Bézier curve for $t \in [0,1]$ with control points $P_0, P_1, P_2, P_3$ defined as:

$$B(t) = (1 - t)^3 P_0 + 3(1 - t)^2 t\, P_1 + 3(1 - t)t^2\, P_2 + t^3 P_3 \qquad (F10)$$

when factored will lead to

$$B(t) = P_0 + 3(P_1 - P_0)t + 3(P_2 - 2P_1 + P_0)t^2 + (P_3 - 3P_2 + 3P_1 - P_0)t^3 \qquad (F11)$$

Equation (F11) suggests a natural cubic generalization of ArcGD, where the coefficients of $t^0, t^1, t^2$,and $t^3$correspond to independently tunable hyperparameters.

Thus, ArcGD's update profile in the $T$-space admits an exact Bernstein polynomial representation, where the quadratic form corresponds to a Bézier curve on each sign branch. Under such interpretation, the optimizer can be viewed as designing a curve in $T$-space, with its hyperparameters directly controlling the shape of the update response. The perspective naturally extends to higher-order Bernstein polynomials, suggesting a principled pathway to construct more expressive optimizer variants. Importantly, the interpretation applies to the update-response mapping rather than the full parameter trajectory, providing both interpretability and a systematic framework for optimizer design.

## Gradient Statistics Driven Control of the ArcGD Update Profile

A key advantage of the ArcGD formulation is that the location of its update-profile peak admits an explicit analytical expression. In particular, the quadratic response governing the update magnitude (Equation (F6)) achieves its maximum for $T \geq 0$ at

$$T^* = \frac{a + b - c}{2b} \qquad (F12)$$

where $a$, $b$, and $c$ denote the ceiling, transition, and floor parameters, respectively. Such closed-form relation enables a principled mechanism for adapting the effective operating regime of the optimizer by aligning the peak update location with gradient statistics. Unlike fixed hyperparameter schedules, it allows ArcGD to reposition its strongest updates toward the most informative gradient magnitudes encountered at different stages of optimization.

*Estimating $T_{max}$ (maximum possible value of T) from Gradient Statistics*

At iteration $t$, ArcGD calculates $T_t$ for a given dimension as

$$T_t = \frac{g_t}{\sqrt{1 + g_t^2}} \qquad (F13)$$

which can be computed elementwise (or aggregated per layer) using exponential moving average. Since $T_t \in (-1,1)$, it provides a stable and bounded representation of gradient magnitude, well suited for statistical tracking. A representative estimate of the current gradient regime can be obtained by maintaining a running statistic of $|T_t|$, for example via an exponential moving average or a robust estimator:

$$\hat{T}_t = \mathrm{Stat}(|T_t|) \qquad (\mathrm{F14})$$

Conceptually:

$$\hat{T}_t \approx \text{“where most gradients live right now”}$$

In practice, $T_{\max}$ can be determined in two ways:

- **Fixed estimate:** obtained from gradient statistics collected during an initial warm-up phase or from historical training runs and kept constant thereafter.
- **Adaptive estimate:** updated online using current gradient statistics.

The fixed variant provides simplicity and stability, while the adaptive variant enables responsiveness to changing training dynamics.

*Slow Online Adaptation of $T_{max}$*

In the adaptive setting, $T_{\max}$ can be possibly updated using a slow update rule:

$$T_{\max}^{(t)} = (1 - \lambda)\, T_{\max}^{(t-1)} + \lambda\, \hat{T}_t, \qquad \lambda \ll 1 \qquad (F15)$$

A small adaptation rate $\lambda \in [1\mathrm{e}-4, 1\mathrm{e}-2]$ can ensure smooth evolution of the update profile, robustness to stochastic gradient noise, and preservation of ArcGD's bounded-update behavior. For the first run, set $T_{max}$ either to a short warm-up average of the first few $\hat{T}_t$ values or simply to $\hat{T}_0$, where $\hat{T}_0 = | T_0 | \in [0,1)$ (or, heuristically, $\hat{T}_0 = 0.60$, which is $> 0.5$ and therefore safe as discussed below).

Such slow update can allow the optimizer to track long-term shifts in gradient distributions, such as the transition from sharp early-stage dynamics to flatter late-stage regimes. Given an estimate $T_{\max}$ (either fixed or adaptively updated), the quadratic peak condition can be enforced statically or dynamically. Keeping $a$ and $c$ fixed, the transition parameter $b$ is adjusted so that the peak aligns with the chosen operating point.

Solving

$$T_{\max} = \frac{a + b - c}{2b} \qquad (\mathrm{F16})$$

for $b$ yields

$$b = \frac{a - c}{2T_{\max} - 1} \tag{F17}$$

Equation (F17) preserves the global ceiling and floor of the update magnitude while shifting the location of maximal response. Provided $a > c$ and $T_{\max} > 0.5$, the resulting $b$ remains positive. When $T_{\max}$ evolves slowly, the induced variation in $b$ is smooth and avoids discontinuities or explicit phase transitions, requiring no additional optimizer state beyond the scalar $T_{\max}$. However, to ensure that peak never drops into mathematically invalid zone ($< 0.5$) resulting in negative $b$, instead of using $T_{max}$, $T_{safe}$ should be used as

$$T_{safe} = \max\left(T^{(t)}, T_{min}\right) \tag{F18}$$

which will lead to

$$b = \frac{a - c}{2T_{\text{safe}} - 1} \tag{F19}$$

$T_{min}$ should be greater than 0.5 and can be fixed as 0.55 or 0.6 or any other suitable value for practical purpose. Moreover, $b$ can also be capped for practical use.

In summary, such formulation elevates ArcGD from a fixed geometric optimizer to a self-regulating system, where gradient statistics statically or dynamically steer the location of maximal update response. By leveraging a closed-form relation, it enables smooth, principled adaptation without sacrificing boundedness or structural integrity. Such advantage results in an optimizer that not only reacts to gradients but adapts its own update geometry to the evolving landscape of optimization.

## Generalisation of ArcGD

ArcGD admits a natural generalisation by introducing exponents on the modulation terms. For $p, q > 0$, the general update rule as a function of $T$ can be written as:

$$\Delta x = -(aT + bT(1 - |T|)^p + c\, sign(T)(1 - |T|)^q \tag{F20}$$

Such formulation provides additional flexibility in shaping the update profile. In particular, the term $(1-|T|)$, which governs the attenuation of updates, can now be modulated nonlinearly. When the exponent is less than 1, the corresponding term is amplified in the mid-region ($|T| \approx 0.5$) relative to the base case ($p = q = 1$). Conversely, when the exponent is greater than 1, the contribution is suppressed in that region. As a result, the generalised form allows controlled redistribution of update strength across the gradient spectrum. However, such flexibility comes at the cost of losing the original quadratic (Bezier-like) structure of the update profile.

**Illustrative Cases**

**Case 1: $p = 1$**

In this case, only the floor term is reshaped:

- For $0 < q < 1$, the floor contribution is strengthened in the mid-region relative to the base case.
- For $q > 1$, the floor contribution is weakened in the mid-region relative to the base case.For example, at $T = 0.5$, the contribution of the floor term becomes:
- $0.5c$ for $q = 1$ (base case)

- $0.707c$ for $q = 0.5$
- $0.353c$ for $q = 1.5$

**Case 2: $p = 1$, $q = |\mathrm{T}|$**

$$\Delta x = -(a\,\mathrm{T} \;+\; b\,\mathrm{T}(1 - |\mathrm{T}|) \;+\; c\,\mathrm{sign}(\mathrm{T})\,(1 - |\mathrm{T}|)^{|\mathrm{T}|}) \tag{F21}$$

Here, the exponent becomes self-adaptive. Since $|\mathrm{T}| < 1$, the exponent remains below unity, resulting in a stronger mid-region contribution compared to the base case. As $|\mathrm{T}| \to 0$, the exponent vanishes and:

$$\lim_{|\mathrm{T}|\to 0} (1 - |\mathrm{T}|)^{|\mathrm{T}|} = 1$$

ensuring full recovery of the floor magnitude $c$ in the near-zero regime.

**Case 3: $p = 1$, $q = \frac{1}{|\mathrm{T}|}$**

$$\Delta x = -(a\,\mathrm{T} \;+\; b\,\mathrm{T}(1 - |\mathrm{T}|) \;+\; c\,\mathrm{sign}(\mathrm{T})\,(1 - |\mathrm{T}|)^{\frac{1}{|T|}}) \tag{F22}$$

In this case, the exponent exceeds unity, leading to a reduced contribution in the mid-region. Notably, the limiting behavior as $|\mathrm{T}| \to 0$ is:

$$\lim_{|\mathrm{T}|\to 0} (1 - |\mathrm{T}|)^{\frac{1}{|T|}} = \boldsymbol{e^{-1}}$$

Thus, the floor term asymptotically approaches $ce^{-1}$, rather than $c$. To preserve the intended floor magnitude, the coefficient may be rescaled as $c \to ce$.

The generalisation as shown by Case 2 and Case 3 extends ArcGD from a fixed quadratic update to a family of self-adjusting update shapes, where the exponents depend directly on the $|\mathrm{T}|$. Unlike fixed exponents, the self-adjusting design lets the optimizer change its floor and transition contributions automatically based on the current gradient, without extra hyperparameters or tuning.

- For $q = |\mathrm{T}|$: the floor term is stronger for mid-sized gradients and fully recovered near zero, so the optimizer naturally focuses on the most informative gradients during training.
- For $q = \frac{1}{|\mathrm{T}|}$: the floor term is weaker in the mid-region while still being bounded.

Such mechanism keeps (generalised) ArcGD bounded, and smooth while adding dynamic, gradient-driven shaping. Overall, it shows that (generalised) ArcGD is not a static update rule, but a self-adjusting optimizer that can change its update emphasis in real time which can potentially improve both stability and efficiency across different optimization tasks.

## Appendix F (Heuristic suggestions)

The term $T_i$ in the following suggestions stands for:

$$T_i = \frac{g_{t,i}}{\sqrt{1 + g_{t,i}^2}}, i = 1,2, \ldots, n,$$

where $g_t = \nabla f(x_t)$ and $i$ indexes the model parameters.

**1. Bounded Adaptive Floor Update ($c_{\text{adapt}}$)**

The floor constant $c$ can be adapted per parameter as:

$$c_{\text{adapt},i} = \max\left(c_{\text{low}}, \min\left(c_{\text{high}}, \frac{\eta_{\text{low}} \mid T_i \mid}{1 - \mid T_i \mid}\right)\right),$$

where $c_{\text{low}}$ and $c_{\text{high}}$ define the allowed bounds.

- Example: $c_{\text{high}} = 10^{-4},\ c_{low} = 10^{-8}$.
- It ensures that the floor update is bounded and can be referred to as **Bounded Adaptive C**.

**2. Overshooting Control Heuristic for Constant $c$**

To reduce overshooting when $T_i$ is very small ($T_i < 0.1$ or $T_i < 0.01$):

1. **Halve the floor constant**:

$$c \leftarrow c/2$$

2. **Smooth the update**:

$$x_{\text{new}} = 0.5(x_{\text{new}} + x_{\text{old}})$$

It prevents abrupt changes in flat regions.

**3. Alternative Formulation for $eta_{low}$**

Another approach is:

$$eta_{\text{low, i}} = \frac{c(1 - \mid T_i \mid)}{\mid T_i \mid + \epsilon}, \epsilon = 10^{-8},$$

which leads to:

$$eta_{\text{low, i}} T_i = \frac{c(1-|T_i|)}{|T_i| + \epsilon} T_i$$

- $\epsilon$ ensures numerical stability.
- The chosen $c$ controls the effective magnitude of the floor update.